\documentclass[twoside,11pt]{article}

\usepackage[disable]{todonotes} 
\usepackage{amsmath, mathtools} 
\usepackage{amssymb} 
\usepackage{graphicx, array, adjustbox, booktabs} 
\usepackage{subfigure}
\usepackage[inline]{enumitem} 
\usepackage{longtable}
\usepackage{multirow}
\usepackage{tikz}
\usepackage{diagbox} 
\usepackage[]{bm}
\usepackage{upgreek} 
\usepackage[flushleft]{threeparttable} 

\newcommand{\ie}{i.e.}

\usepackage{caption}

\DeclareMathOperator*{\sigmoid}{sigmoid}
\DeclareMathOperator*{\softmax}{softmax}
\DeclareMathOperator*{\grSigmoid}{sigmoid_{gr}}
\DeclareMathOperator*{\grTanh}{tanh_{gr}}

\usepackage{jmlr2e} 

\usepackage{lastpage}
\jmlrheading{23}{2022}{1-\pageref{LastPage}}{2/21; Revised
3/22}{6/22}{21-0211}{Bhumika Mistry, Katayoun Farrahi and Jonathon Hare} 

\ShortHeadings{A Primer for Neural Arithmetic Logic Modules}{Mistry, Farrahi and Hare} 
\firstpageno{1}

\begin{document}

\title{A Primer for Neural Arithmetic Logic Modules}

\author{\name Bhumika Mistry \email bm4g15@soton.ac.uk \\
        \name Katayoun Farrahi \email k.farrahi@soton.ac.uk \\
        \name Jonathon Hare \email jsh2@ecs.soton.ac.uk \\
       \addr Department of Vision Learning, and Control\\ 
       Electronics and Computer Science\\
       University of Southampton\\
       Southampton, SO17 1BJ, United Kingdom 
       }

\editor{Stefan Harmeling}

\maketitle

\begin{abstract}
Neural Arithmetic Logic Modules have become a growing area of interest, though remain a niche field. 
These modules are neural networks which aim to achieve systematic generalisation in learning arithmetic and/or logic operations such as $\{+, -, \times, \div, \leq, \textrm{AND}\}$ while also being interpretable. 
This paper is the first in discussing the current state of progress of this field, explaining key works, starting with the Neural Arithmetic Logic Unit (NALU). 
Focusing on the shortcomings of the NALU, we provide an in-depth analysis to reason about design choices of recent modules. 
A cross-comparison between modules is made on experiment setups and findings, where we highlight inconsistencies in a fundamental experiment causing the inability to directly compare across papers. 
To alleviate the existing inconsistencies, we create a benchmark which compares all existing arithmetic NALMs. 
We finish by providing a novel discussion of existing applications for NALU and research directions requiring further exploration. 
\end{abstract}


\section{Introduction}
The ability to learn by composition of already known knowledge is a form of \textit{systematic generalisation}~\citep{fodor1988connectionism}, also termed as \textit{compositional generalisation}~\citep{lake2019compositional}.
Humans can apply such generalisations for arithmetic after learning the relevant underlying rules. 
For example, combining primitive operations such as addition ($a+b$) and multiplication ($a\times b$) on already observed inputs to produce more complex expressions (such as $(a+b)\times(c+d)$). 
Humans can also transfer their skills in applying operations on limited set of numbers (for example between 1-100) to the range of unobserved numbers. 
This ability to \textit{extrapolate}, that is generalise to out-of-distribution (OOD) data, is a desirable property for neural networks. 
Research suggests neural networks struggle to extrapolate even for the simplest of tasks such as learning the identity function~\citep{trask2018neural}. 
Rather than generalising, networks lean towards memorisation in which the model memorises the training labels~\citep{Zhang2020Identity}. 

To address this issue, \citet{trask2018neural} introduce the first of a new class of neural modules which we term \textbf{Neural Arithmetic Logic Modules (NALMs)}. 
Their module, the NALU, aims to learn systematic generalisation for arithmetic computations. 
For example, learning the relation between input [$x_1, x_2, x_3, x_4$] and output $o_1$ where the input elements are real numbers and output is expression $x_1 + x_3 - x_2$. 
To achieve this they assume an inductive bias such that particular weight values can be intuitively interpreted as different primitive arithmetic operations. For example, using a weight of 1 to represent addition and weight of -1 to represent subtraction. 
This form of interpretability is comparable to the definition of \textit{decomposable transparency} by \citet{Lipton2016}. 
Though NALU shows promising improvements over networks such as Multilayer Perceptrons (MLPs) for extrapolation, the unit still presents various shortcomings in architecture, convergence, and transparency. 
These areas for improvement inspired the design of other modules~\citep{heim2020neural, madsen2020neural, schlor2020inalu, rana2019exploring}. 
Due to the growing interest of NALMs, we believe it is important to have a resource, this paper, to explain current motivations, strengths, weaknesses and gaps in this line of research. 
\subsection{Contributions} 
\begin{enumerate}
    \item We provide the first definition to describe this research field by defining a NALM---a neural network with the ability to model logic and/or arithmetic in a generalisable manner to OOD data whilst expressing an interpretable solution. 
    \item We explain how recent modules are designed to overcome various shortcomings of NALU including: inability to process negative inputs and outputs, lack of convergence and adhering to its inductive bias, weak modelling of the division operation, and lack of compositionality. 
    \item We highlight how a popular experiment for testing modules arithmetic capabilities is inconsistent between different papers with regards to hyperparameters and experiment setup. This leads to providing a new benchmark for comparing existing (and future) arithmetic NALMs. 
    \item Using the first NALM, the NALU, as a focal point we show the usefulness of NALUs in larger differentiable applications which require arithmetic and extrapolation capabilities, while also making aware situations in which NALU is sub-optimal.
    \item We outline possible research directions regarding modelling division, robustness across different training ranges, compositionality of modelled expressions, and analysing the impact of NALMs when integrated with other networks. 
\end{enumerate}

\subsection{Outline} 
In this paper we begin by defining a NALM, motivating their aim and uses in Section~\ref{sec:NALM}. 
Section~\ref{sec:nalu-arch} and \ref{sec:other-units} explains the definitions of key NALMs: NALU, iNALU, NAU, NMU, and NPU to build understanding. 
Using the first NALM, the NALU, as a focal point, Section~\ref{sec:shortcomings-and-solutions} provides an in-depth analysis of the shortcomings of NALU to understand the motivation behind design choices for more recent NALMs. 
Section~\ref{sec:exp-and-findings} highlights inconsistencies in experiment setup and compares findings across existing modules. 
Additionally, we provide our own findings comparing arithmetic NALMs using a consistent experiment setup. 
Section~\ref{sec:exp-and-findings-logic} discusses the findings of NALMs which specialise in logic operations. 
Section~\ref{sec:applications} shows the diversity in NALU's use in applications, while also indicating situations in which NALU is sub-optimal. 
Section~\ref{sec:remaining-gaps} considers all discussed issues and outlines remaining gaps, suggesting possible research directions to take as a result. 
We end by providing related work in Section~\ref{sec:rel-work} which takes a step back and considers the wider research around areas relevant to NALMs including extrapolative mathematics, inductive biases and specialist modules. 

\subsubsection{Mathematical Notation}
When the individual NALM modules are defined, the mathematical notation will be in element-wise form which provides how to calculate an output element $y_o$ indexed at $o$ given a single data sample (input vector $\mathbf{x}$). 
For completeness, we also provide illustrations for each module using the matrix/vector with symbols and colouring following the key in Appendix~\ref{app:arch-key}. 

\section{What are NALMs and Why use them?}\label{sec:NALM}
We begin by defining NALMs. More specifically, before we detail instances of NALMs, we first answer three questions: \begin{enumerate*}
  \item What is a NALM?
  \item What is the aim of a NALM?
  \item Why is a NALM useful?
\end{enumerate*}

From answering these questions, we shall arrive at the following \textbf{definition}: \textit{A NALM is a neural network that performs arithmetic and/or logic based expressions which can extrapolate to out-of-distribution (OOD) data when parameters are appropriately learnt whilst expressing an interpretable solution.}

\subsection{What is a NALM?}
NALM stands for Neural Arithmetic Logic Module. \textit{Neural} refers to neural networks. 
\textit{Arithmetic} refers to the ability to learn arithmetic operations such as addition. 
\textit{Logic} refers to the ability to learn operations such as selection, comparison based logic (e.g., greater than) and boolean based logic (e.g., conjunction). 
\textit{Module} refers to the architecture's of the neural units which learns the arithmetic and/or logic operations. 
The term module encompasses both a single (sub-)unit and multiple (sub-)units combined together. 

\textbf{What kind of operations can be learnt?} Existing work has tried to model arithmetic operations including addition, subtraction, multiplication, division, square, and square-root. 
Logic based operations include logic rules (e.g., conjunction)~\citep{reimann2019neural}, control logic (e.g., $<=$)~\citep{faber2020neural} and selection of relevant inputs. 

\textbf{How are operations learnt?} Because a NALM is a neural network, a module can model the relation between input and output vectors via supervised learning which trains weights through backpropagation. 
To learn the relation between input and output, requires learning to select relevant elements of the input and apply the relevant arithmetic operation/s to the selected input to create the output. 

\begin{figure*}[t]
\vskip 0.2in
\centering
\includegraphics[width=\textwidth]{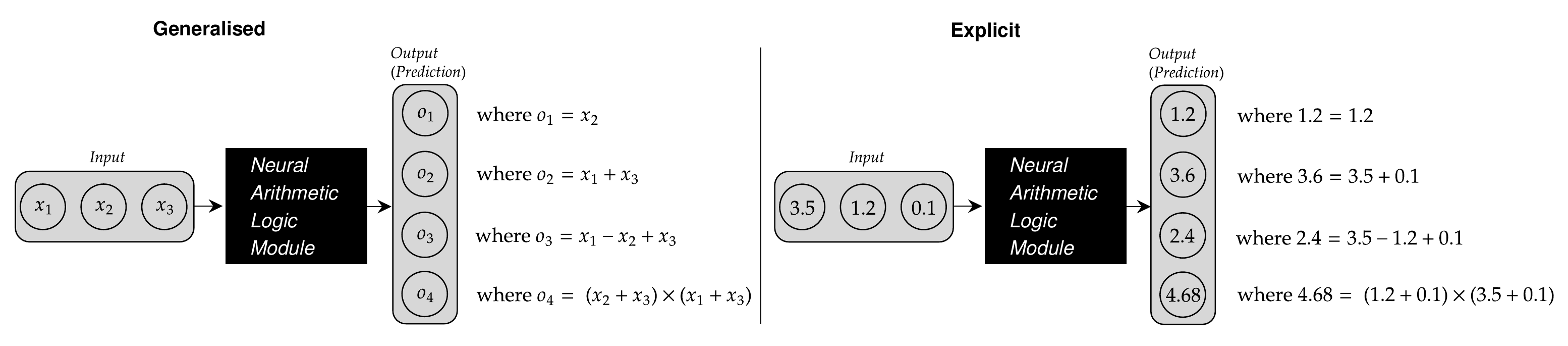}
\caption{High-level example of the input output structure into a NALM. Both networks are the same. The generalised network defines the notation of each element in the input and output. The explicit network is an example of valid input and output values.}
\label{fig:in-nalm-out}
\vskip -0.2in
\end{figure*}

\textbf{How is data represented?} The input and outputs are both vectors. 
Each vector element is a real-valued number which is implemented as a floating point number. 
Each output element can learn a different arithmetic expression. 
For a single data sample, this can be illustrated in Figure~\ref{fig:in-nalm-out} where we assume that the NALM used (made from two stacked sub-units) can learn addition, subtraction and multiplication. In practice data would be given in batch form.

\subsection{What is the Aim of a NALM?}
The main aim of NALMs is to provide systematic generalisation in learning arithmetic and/or logic expressions. 
Once the learning state (training) has ended, if the correct weights have been learned, the NALM is able to also extrapolate to unseen data (i.e., OOD data). 

\textbf{What does interpretability mean for NALMs?} 
Imagine modelling the relation between input $\mathbf{x}$ and output $\mathbf{y}$ with a module $f$ parameterised by $\theta$, i.e., $\mathbf{y} = f_\theta(\mathbf{x})$. 
We say a NALM is interpretable if you can set the module's parameters ($f_\theta$) to express the underlying relation between $\mathbf{x}$ and $\mathbf{y}$ in a provable way. 
Simply put, the weights of a NALM can be set such that, if the expression which NALM calculates is written out, we get an expression which holds for all valid inputs. 
For example, for modelling the addition of two inputs ($x_1$ and $x_2$), having a model which takes in the two inputs and applies a dot product with a weight vector set to ones results in  $\mathbf{y} = f_\theta(\mathbf{x}) = \begin{bmatrix}w_1 & w_2\end{bmatrix}\cdot \begin{bmatrix}x_1\\x_2\end{bmatrix} = \begin{bmatrix}1 & 1\end{bmatrix} \cdot \begin{bmatrix}x_1\\x_2\end{bmatrix}$ which  will always result in the output being $x_1 + x_2$ no matter the values of $x_1$ and $x_2$. 

More broadly speaking, the type of interpretability we want from a NALM is \textit{decomposable transparency} \citep{Lipton2016}. Transparency means to understand how the model works. Decomposability is transparency at component level defined by \citet{Lipton2016} as `each part of the model - each input, parameter, and calculation - admits an intuitive explanation'. 
For example, for modelling $\textrm{force} = \textrm{mass} \times \textrm{acceleration}$, there are: the two inputs into the NALM representing mass and acceleration, the parameters representing the operation (multiplication) which are set to 1, and the calculation that multiplies the two inputs resulting in the value for the force. 

\textbf{What does extrapolation on OOD data mean for NALMs?} \textit{OOD data} refers to data which is sampled outside the training distribution. For example, if trained on a range [0,10] a valid OOD range could be [11,20]. \textit{Extrapolation} is the ability to correctly predict the output when given OOD data. 
In the context of NALMs, extrapolation means that the model successfully learns the underlying principles for modelling the (arithmetic/logic) operations it is designed for.  
From a practical viewpoint, a NALM with successful extrapolative capabilities can be considered as a module where loss in predictive accuracy occurs due to numerical imprecisions of hardware limitations. 

\subsection{Why is a NALM useful?}
The ability to learn arithmetic seems trivial in comparison to other architectures such as LSTMs, CNNs or Transformers which can be used as standalone networks which learn tasks such as arithmetic, object recognition and language modelling. So, why care about NALMs? 

Learning arithmetic, though it may seem a simple task, remains unsolved for neural networks. To solve this problem requires precisely learning the underlying rules of arithmetic such that failure cases will not occur on cases of OOD data. Therefore, before considering more complex tasks, solving the simple tasks seems reasonable. 

Furthermore, even though NALMs specialise in arithmetic there is no restriction in using them as part of larger end-to-end neural networks. 
For example, attaching a NALM to a CNN as residual connections \citep{rana2020systematically} to improve counting in images. 
Utilising a NALM as a specialist module biased towards arithmetic operations provides more focused learning. 
In Section~\ref{sec:applications}, we show a vast array of applications in which NALMs are being utilised. 
Being used as a sub-component in a larger network implies that the sub-component has the ability to learn regardless of the data distribution. 
Therefore, the ability to extrapolate is essential. 

\section{Overview of the NALU Architecture}\label{sec:nalu-arch}
\begin{figure*}[!h]
\centering
\includegraphics[width=\textwidth]{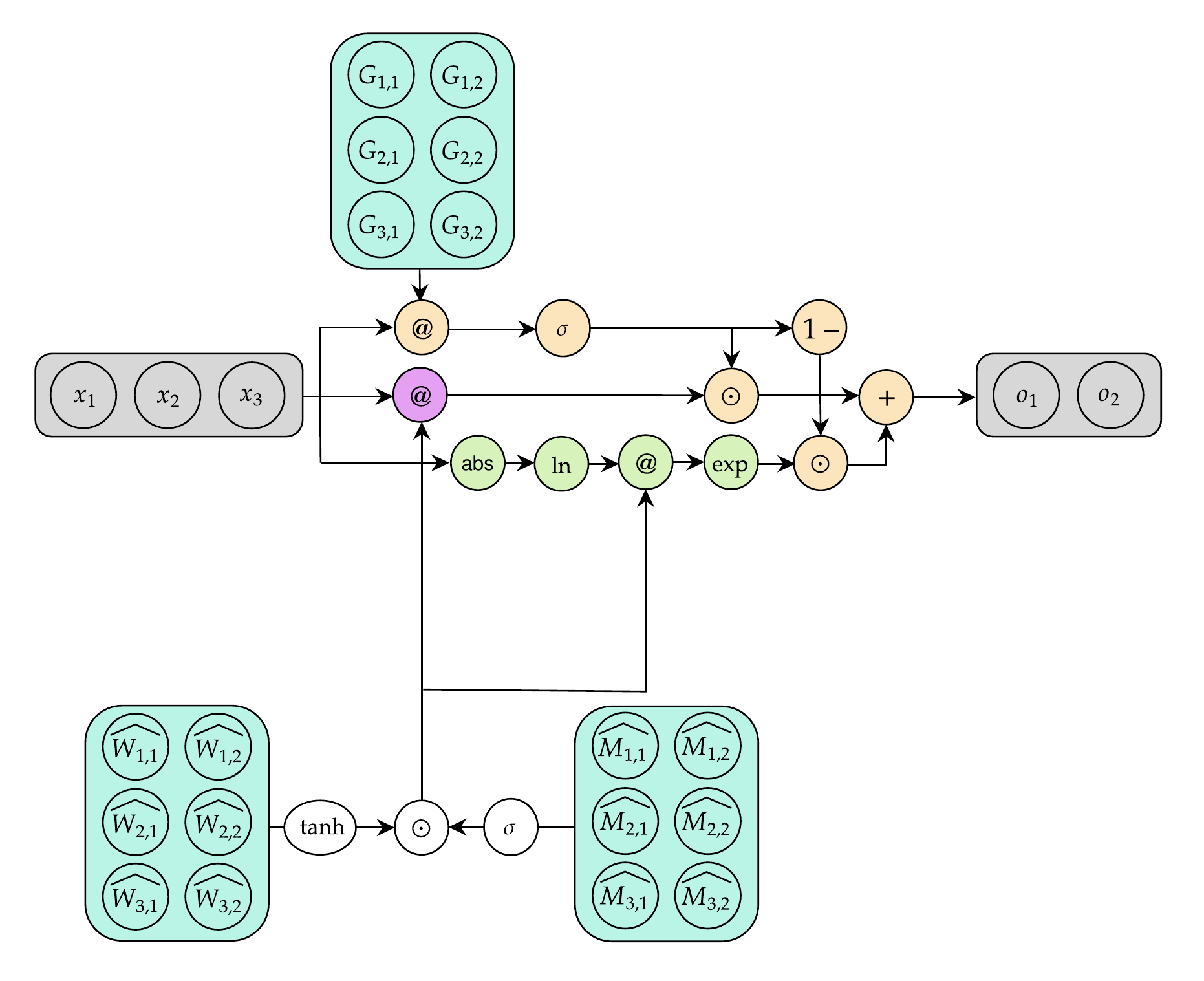}
\caption{NALU architecture. Example of a 3-feature input and 2-feature output model.}
\label{fig:arch-nalu}
\end{figure*}

The NALU, illustrated in Figure~\ref{fig:arch-nalu}, provides the ability to model basic arithmetic operations, specifically: addition, subtraction, multiplication, division. 
NALU requires no indication of which operation to apply and aims to learn weights that provide extrapolation capabilities if correctly converged. 
NALU comprises of two sub-units, a summative unit which models $\{+, -\}$ and a multiplicative unit which models $\{\times, \div\}$. 
Following the notation of~\citet{madsen2020neural} we denote the sub-units as $\mathrm{NAC}_{+}$ and $\mathrm{NAC}_{\bullet}$ respectively. 
Formally, for calculating a specific output value, the NALU is expressed as:
\begin{align}
W_{i,o} &= \tanh(\widehat{W_{i,o}}) \odot \sigmoid(\widehat{M_{i,o}}) \label{eq:W}\\
\mathrm{NAC}_{+}: a_o &= \sum_{i=1}^{I} \left(W_{i,o}\cdot \mathrm{x}_{i} \right) \label{eq:NAC+}\\
\mathrm{NAC}_{\bullet}: m_o &= \exp\left(\sum_{i=1}^{I} (W_{i,o} \cdot \ln(|x_i| + \epsilon))\right) \label{eq:NAC*}\\
g_o &= \sigmoid\left(\sum_{i=1}^{I} (G_{i,o}\cdot x_i)\right) \label{eq:g}\\
\mathrm{NALU}: \hat{y_o} &= g_o \cdot a_o + (1 - g_o) \cdot m_o \label{eq:NALU}
\end{align}

where $\widehat{\bm{W}}, \widehat{\bm{M}} \in \mathbb{R}^{I \times O}$ are learnt matrices ($I$ and $O$ represent input and output dimension sizes). 
A non-linear transformation is applied to each matrix and then both are combined via element-wise multiplication to form $\bm{W}$ (Equation~\ref{eq:W}). 
Due to the range values of $\mathrm{tanh}$ and $\sigmoid$, $\bm{W}$ aims to have a inductive bias towards values $\{-1,0,1\}$ which can be interpreted as selecting a particular operation \textit{within} a sub-unit (\ie, intra-sub-unit selection). 
For example, in $\mathrm{NAC}_{+}$ +1 is addition and -1 is subtraction, and in $\mathrm{NAC}_{\bullet}$ +1 is multiplication and -1 is division. 
In both sub-units, 0 represents not selecting/ignoring an input element.
A sigmoidal gating mechanism (Equation~\ref{eq:g}) enables selection \textit{between} the sub-units (inter-sub-unit), where an open gate, 1, selects the $\mathrm{NAC}_{+}$ and closed gate, 0, selects the $\mathrm{NAC}_{\bullet}$. 
Once trained the gating should ideally select a single sub-unit. 
${\bm{G}}$ is learnt, and the gating vector $\bm{g}$ represents which sub-unit to use for each element in the output vector. 
Finally, Equation~\ref{eq:NALU} gates the sub-units and sums the result to give the output. 
NALU's gating only allows for each output element to have a mixture of operations from the same sub-unit.
Therefore, each output element is an expression of a combination of operations from either $\{+, -\}$ or $\{\times, \div\}$ but not $\{+, -, \times, \div\}$. This issue is fixed by stacking NALUs such that the output of one NALU is the input of another. 
A step-by-step example for the NALU is given in Appendix~\ref{app:step-by-step-nalu}. 
Next, we overview architectures of some recent modules. 

\section{NALU Influenced Modules}\label{sec:other-units}
NALU has inspired the creation of other modules including: Improved NALU (iNALU)~\citep{schlor2020inalu}, Neural Addition Unit (NAU)/ Neural Multiplication Unit (NMU)~\citep{madsen2020neural}, Neural Power Unit (NPU)~\citep{heim2020neural}, Golden Ratio NALU (G-NALU)~\citep{rana2019exploring}, Neural Logic Rule Layer (NLRL)~\citep{reimann2019neural} and Neural Status Register (NSR)~\citep{faber2020neural}. 
This section will go through each of the modules definitions, providing a consistent notation for each module along with an illustrations of the module's architecture.\footnote{Module illustrations from the original papers are provided in Appendix~\ref{app:unit-imgs}.}

\subsection{iNALU}
\begin{figure*}[!h]
\centering
\includegraphics[width=\textwidth]{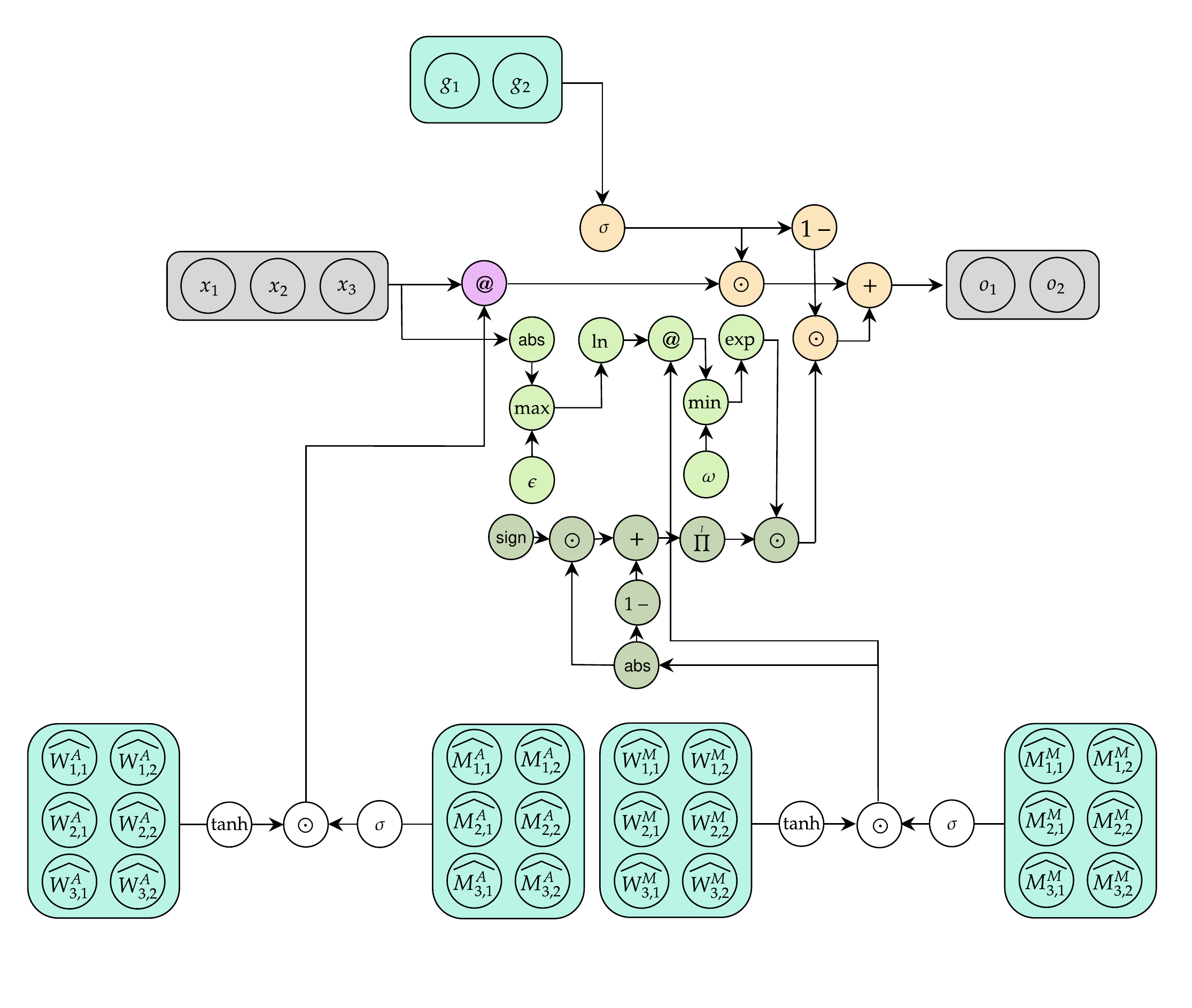}
\caption{iNALU architecture. Example of a 3-feature input and 2-feature output model.}
\label{fig:arch-inalu}
\end{figure*}

The \textbf{iNALU} identifies key issues in NALU and modifies the unit to incorporate solutions (detailed in Section~\ref{sec:shortcomings-and-solutions}). In particular, they use:
\begin{itemize}
    \item \textbf{Independent weight matrices}. To allow the multiplicative and summative paths to learn their own set of $\bm{\hat{W}}$ and $\bm{\hat{M}}$ weights to be used in calculating $\bm{a}$ for the $\mathrm{NAC}_{+}$ and $\bm{m}$ for the $\mathrm{NAC}_{\bullet}$.  
    \begin{align}
        W^{\textrm{A}}_{i,o} &= \tanh(\widehat{W^{\textrm{A}}_{i,o}}) \cdot \sigmoid(\widehat{M^{\textrm{A}}_{i,o}}) \label{eq:inlau-Wa}, \\
        W^{\textrm{M}}_{i,o} &= \tanh(\widehat{W^{\textrm{M}}_{i,o}}) \cdot \sigmoid(\widehat{M^{\textrm{M}}_{i,o}}) \label{eq:inlau-Wm}.
    \end{align}
    \item \textbf{Clipping}. Clipping the multiplicative weights using the equation below (with $\omega=20$) and clipping the gradient of learnable parameters between [-0.1,0.1]. 
    \begin{align}
        m_o &= \exp(\min(\ln(\max(|x_i|,\epsilon))\cdot W^{\textrm{M}}_{i,o}, \;\omega). \label{eq:inalu-mo-clip}
    \end{align}
    \item \textbf{Multiplicative sign correction}. Retrieve the output sign of the multiplicative path, 
    \begin{align}
        msv_o &= \prod_{i=1}^I \left(\mathrm{sign}(x_i) \cdot |W^{\textrm{M}}_{i,o}| + 1 - |W^{\textrm{M}}_{i,o}| \right) \;. \label{eq:inalu-msv}
    \end{align}
    \item \textbf{Regularisation}. Include a regularisation loss term which avoids having near-zero learnable parameters, 
    \begin{align}
        \mathcal{R}_{\mathrm{sparse}} = \frac{1}{t}\sum_{\mathclap{\bm{\theta} \in 
            \{{\bm{\widehat{W^\textrm{A}}}, 
                \bm{\widehat{M^\textrm{A}}},
                \bm{\widehat{W^\textrm{M}}}, 
                \bm{\widehat{M^\textrm{M}}}, 
                \bm{g}}
            \}}}
        \frac{\sum_o^O\sum_i^I\max(\min(-\theta_{i,o}, \theta_{i,o}) + t, 0)}{O\cdot I} \;, \label{eq:inalu-rsparse}
    \end{align}
    where $t=20$. This activates if the loss is under 1 and there have been over 10 iterations of training data.
    \item \textbf{Reinitialisation}. Reinitialise the model weights if the average loss collected over a number of consecutive iterations has not improved. 
    More specifically, reinitialisation occurs for every $10^{th}$ iteration, if over 10,000 iterations have occurred and the average loss of the first half of those iterations of the errors is less than the average loss of the second half plus its standard deviation, and the average loss of the latter half of errors is larger than 1. 
    \item \textbf{Independent gating}. Remove the dependence of the input values when learning the gate parameters,  
        \begin{align}
            g_o =& \sigmoid(g_{o}) \;. \label{eq:inalu-g}
        \end{align}
    
\end{itemize}

The iNALU expression for calculating a single output element indexed at $o$ is
\begin{align}
    \mathrm{iNALU}: \hat{y_o} &= g_o \cdot a_o + (1 - g_o) \cdot m_o \cdot msv_o \;.
\end{align}

\subsection{NAU and NMU}
\begin{figure*}[!h]
\centering
\includegraphics[width=0.6\textwidth]{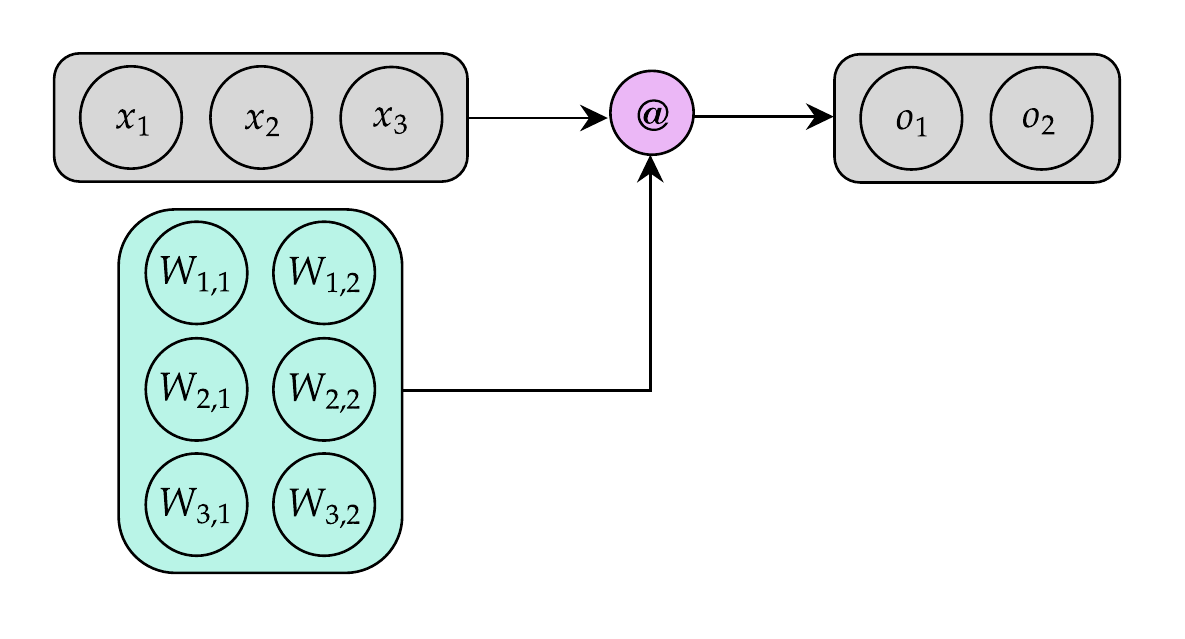}
\caption{NAU architecture. Example of a 3-feature input and 2-feature output model.}
\label{fig:arch-nau}
\end{figure*}

\begin{figure*}[!h]
\centering
\includegraphics[width=\textwidth]{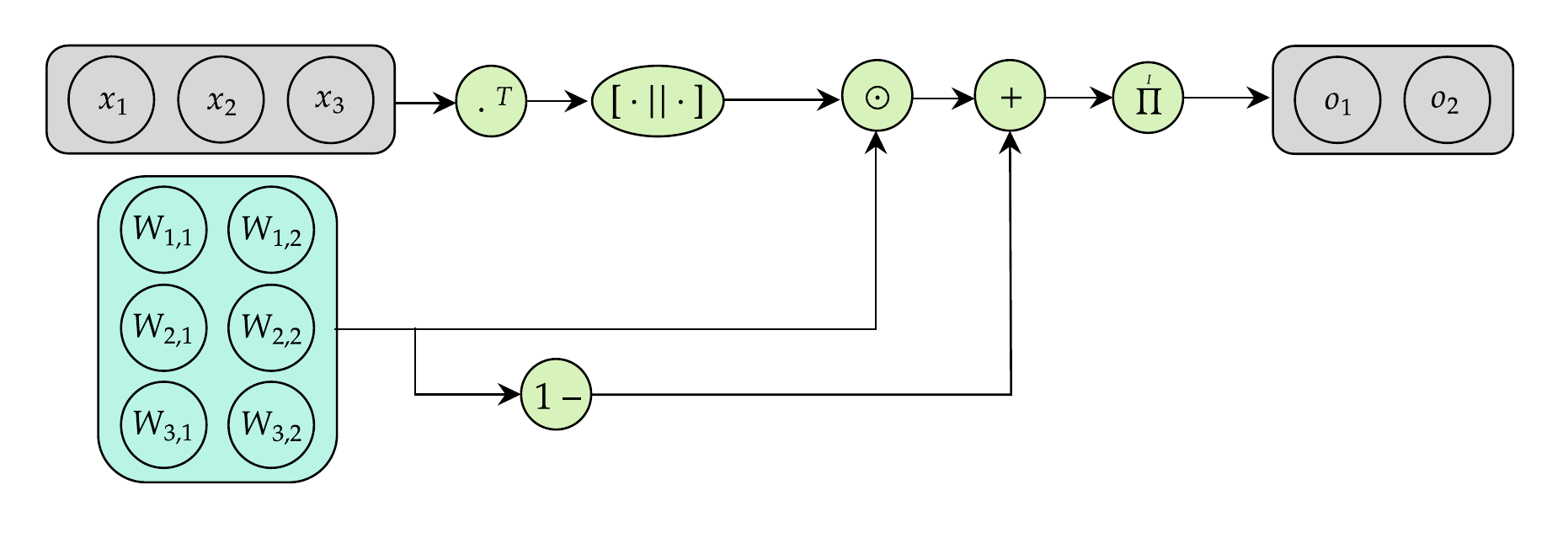}
\caption{NMU architecture. Example of a 3-feature input and 2-feature output model.}
\label{fig:arch-nmu}
\end{figure*}

The \textbf{NAU} and \textbf{NMU} are sub-units for addition/subtraction and multiplication respectively. Architecture and initialisations of the sub-units have strong theoretical justifications and empirical results to validate design choices.
The NAU and NMU definitions for calculating an output element indexed at $o$ is:
\begin{align}
\textrm{NAU}: a_o &= \sum_{i=1}^{I} \left(W_{i,o}\cdot \mathrm{x}_{i} \right), \label{eq:nau}\\
\textrm{NMU}: m_o &= \prod_{i=1}^{I} \left(W_{i,o}\cdot \mathrm{x}_{i} + 1 - W_{i,o} \right) \;, \label{eq:nmu}
\end{align}
where the $\bm{W}$ is unique for each sub-unit.
Prior to applying the weights of a sub-unit to the input vector, each element of $\bm{W}$ is clamped between [-1,1] if using the NAU, or [0,1] if using the NMU. 
Therefore, considering discrete weights $\{-1,0,1\}$, Equation~\ref{eq:nau} will do the summation of the inputs where each input is either added ($W_{i,o}=1$), ignored ($W_{i,o}=0$), or subtracted ($W_{i,o}=-1$). 
When considering the discrete weight values of the NMU $\{0,1\}$, the result is the product of the inputs where each input is either multiplied ($W_{i,o}=1$) or not selected ($W_{i,o}=0$). Rather than allowing the product of the inputs to be multiplied by 0 whenever an irrelevant input (i.e., with weight of 0) is processed, Equation~\ref{eq:nmu} will also convert the input to be 1 (the multiplicative identity value) resulting in the input not having any effect towards the final output. 

To enforce the module weights to become discrete values, the following regularisation loss term is also used, 
\begin{equation}
\mathcal{R}_{sparse} = \frac{1}{I \cdot O} \sum_{o=1}^{O} \sum_{i=1}^{I} \min\left(|W_{i,o}|, 1 - |W_{i,o}|\right) . \label{eq:madsen-rsparse}
\end{equation}

A scaling factor
\begin{equation}
\lambda = \hat{\lambda} \max\left(\min\left(\frac{iteration_i - \lambda_{start}}{\lambda_{end} - \lambda_{start}}, 1\right), 0\right), 
\label{eq:regualizer-scaling}
\end{equation}
is multiplied to $\mathcal{R}_{sparse}$ to get the final value, where regularisation strength is scaled by a predefined $\widehat{\lambda}$.

\subsection{NPU and RealNPU}
\begin{figure*}[!h]
\centering
\includegraphics[width=\textwidth]{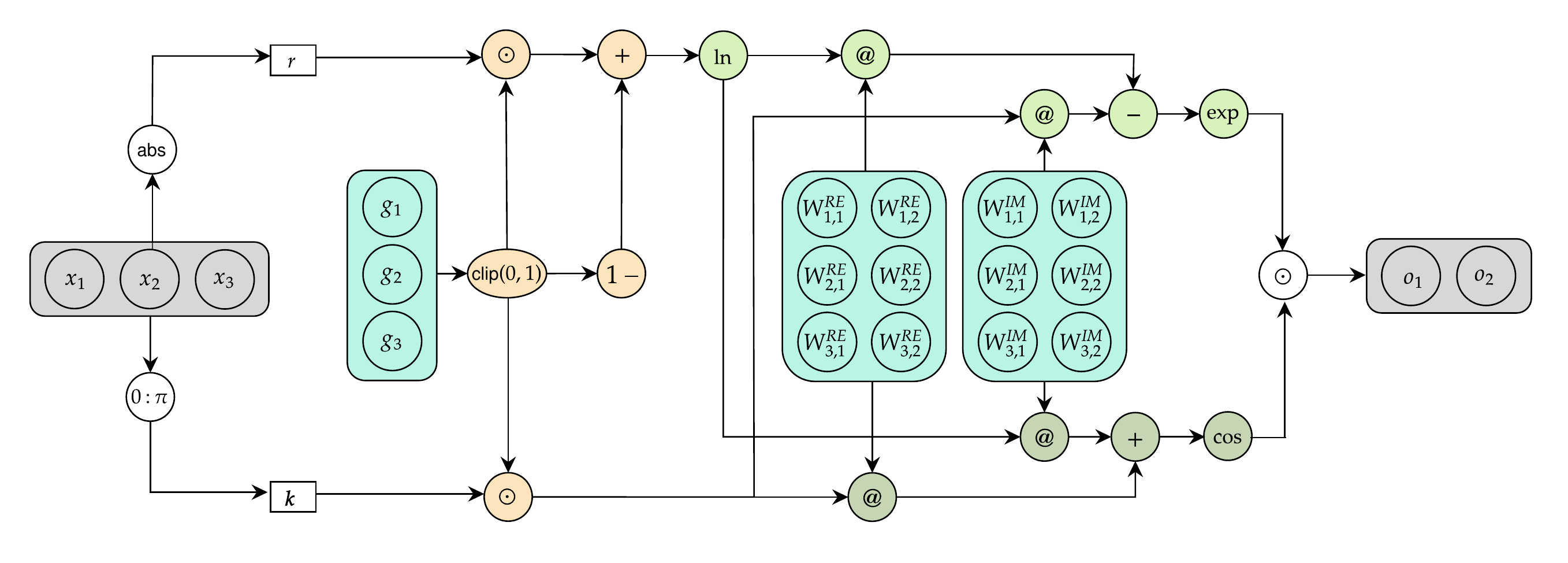}
\caption{NPU architecture. Example of a 3-feature input and 2-feature output model.}
\label{fig:arch-npu}
\end{figure*}

The \textbf{NPU} (Equation~\ref{eq:npu}) focuses on improving the division ability of the $\mathrm{NAC}_{\bullet}$ by applying a complex log transformation and using real and complex weight matrices ($\bm{W^{\textrm{RE}}}$ and $\bm{W^{\textrm{IM}}}$ respectively).  
NPU based modules can model products of arbitrary powers ($\prod x_i^{w_i}$), therefore the learnable weight parameters do not require to be discrete. 
For example, modelling the square-root operation requires $W^{\textrm{RE}}_{i,o}=0.5$. 
The $\bm{r}$ (Equation \ref{eq:npu-r}) converts values close to 0 into 1 to avoid the output multiplication becoming 0. 
To do this, a relevance gate ($\bm{g}$) is learnt representing if an input element is relevant and should be used as part of an output expression ($g_i=1$) or not be selected ($g_i=0$). 
Furthermore, each element of $\bm{g}$ is clipped between the range [0,1] (Equation~\ref{eq:npu-g}).  
\begin{equation}
\begin{aligned}
\mathrm{NPU} : y_o = &\exp \left(\sum_{i=1}^{I}(W^{\textrm{RE}}_{i,o}\cdot \ln(r_i)) - \sum_{i=1}^{I}(W^{\textrm{IM}}_{i,o} \cdot k_i)\right) \\& \cdot  \cos\left(\sum_{i=1}^{I}(W^{\textrm{IM}}_{i,o}\cdot \ln(r_i)) + \sum_{i=1}^{I}(W^{\textrm{RE}}_{i,o}\cdot k_i)\right) \label{eq:npu}
\end{aligned}
\end{equation}
where
\begin{align}
r_i &= g_i \odot (|x_i|+ \epsilon) + (1 - g_i) , \label{eq:npu-r}\\
k_i & = 
    \begin{cases}
       0 & x_i \geq 0 \\
       \pi\mathrm{g_i} & x_i < 0 
    \end{cases} ,\\
\shortintertext{and}
g_i &= \mathrm{min}(\mathrm{max}(g_i,0),1) \;. \label{eq:npu-g}
\end{align}
Additionally a simplified version of the NPU exists, named the \textbf{RealNPU}, considering only real values of Equation~\ref{eq:npu}: 
\begin{equation}
\begin{aligned}
\mathrm{RealNPU} := &\exp \left(\sum_{i=1}^{I}(W^{\textrm{RE}}_{i,o}\cdot \ln(r_i))\right) \cdot \cos\left(\sum_{i=1}^{I}(W^{\textrm{RE}}_{i,o} \cdot k_i)\right) . \label{eq:real-npu}
\end{aligned}
\end{equation}

As the NPU and RealNPU can express arbitrary powers, using a regulariser to enforce discrete parameters like in the iNALU, NAU or NMU would restrict the expressiveness. Therefore, \citet{heim2020neural} use a scaled L1 penalty, where the scaling value $\beta$ grows between predefined values $\beta_{start}$ to $\beta_{end}$ and is increased every $\beta_{step} = 10,000$ iterations by a growth factor $\beta_{growth} = 10$. 

\subsection{G-NALU}
\begin{figure*}[!h]
\centering
\includegraphics[width=0.8\textwidth]{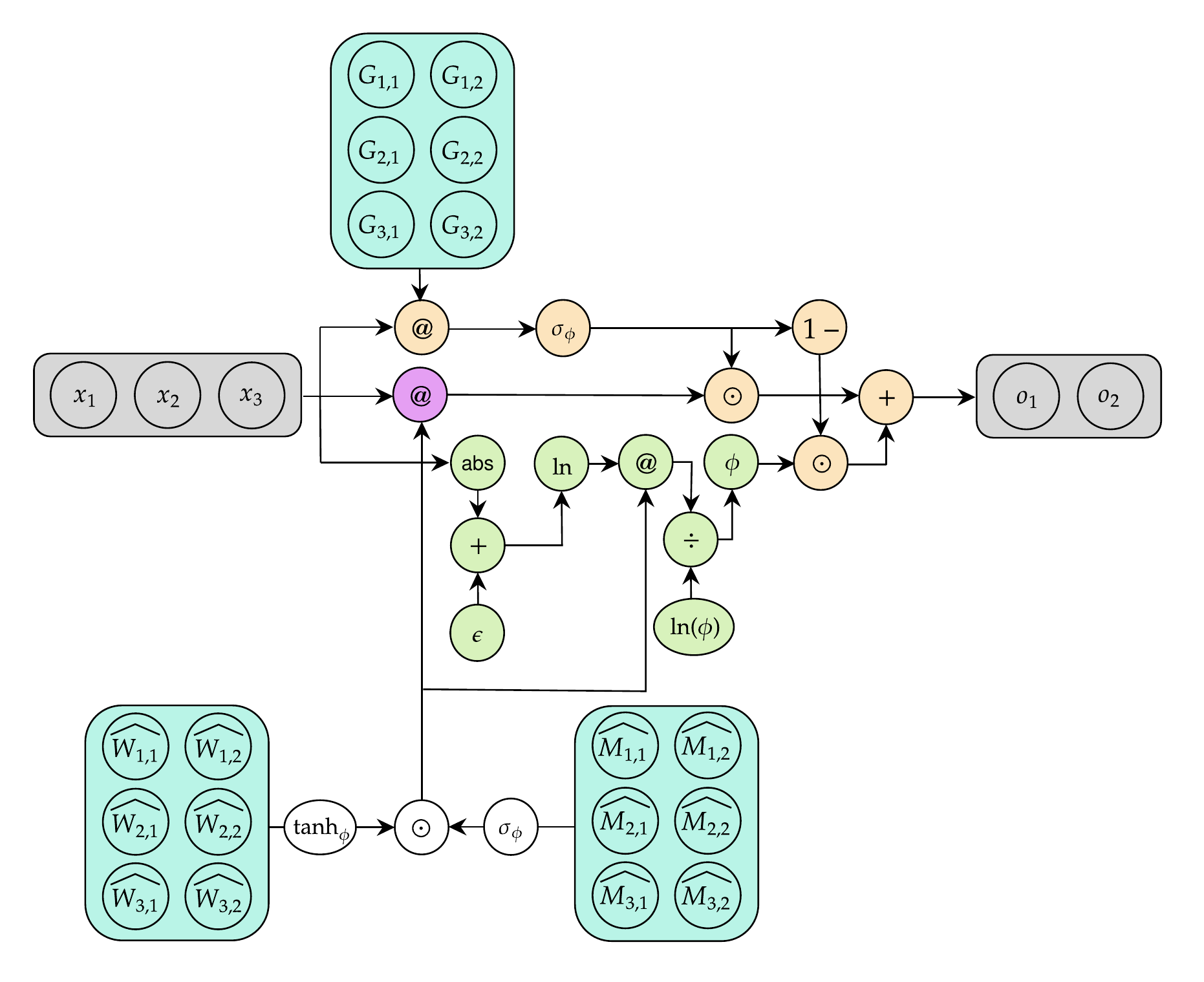}
\caption{G-NALU architecture. Example of a 3-feature input and 2-feature output model.}
\label{fig:arch-gnalu}
\end{figure*}

The \textbf{G-NALU} replaces the exponent base in the $\mathrm{tanh}$ and $\sigmoid$ operations when calculating NALU's weight matrix with a golden ratio base value:
\begin{align}
    \phi &= \frac{1 + \sqrt{5}}{2} \approx 1.618 \\
    \grSigmoid &= \frac{1}{(1 + \phi^{-x})} \\
    \grTanh &= \frac{\phi^{2x}-1}{\phi^{2x}+1} 
\end{align}
The use of a golden ratio base also requires the $\mathrm{NAC}_{\bullet}$ definition (Equation~\ref{eq:NAC*}) to be modified into Equation~\ref{eq:G-NAC*} to allow for the $\ln$-$\exp$ transformation to work. 
\begin{align}
    \mathrm{NAC}_{\bullet}: m_o &= \phi^{\left(\sum_{i=1}^{I} \frac{(W_{i,o} \cdot \ln(|x_i| + \epsilon))}{\ln(\phi)}\right)} \label{eq:G-NAC*}
\end{align}

\subsection{NLRL}
\begin{figure*}[b]
\centering
\includegraphics[width=\textwidth]{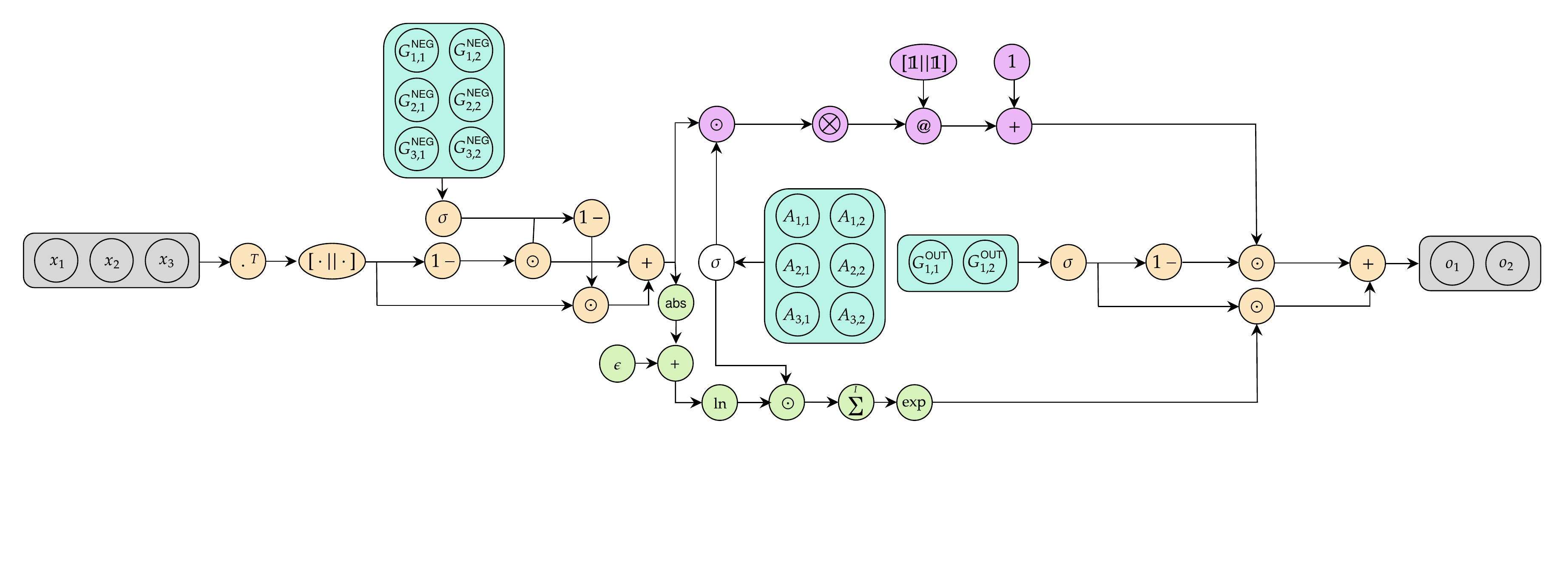}
\caption{NLRL architecture. Example of a 3-feature input and 2-feature output model.}
\label{fig:arch-nlrl}
\end{figure*}

The \textbf{NLRL}, Figure~\ref{fig:arch-nlrl}, is a module to express boolean logic rules and simple arithmetic operations (add, subtract and multiply) via modelling AND (conjunction), OR (disjunction) and NOT (negation). 
By stacking NLRLs together, it is also possible to represent more complex relations including implication, exclusive OR and equivalence. 
The architecture is designed under the assumption of modelling the logic rules on booleans, therefore the input values must be booleans. 
The default NLRL architecture consists of the following four parts in which the three base operators (negation, conjunction and disjunction) are modelled: 
\begin{itemize}
    \item \textbf{Negation gating}, which models the negation operator. The (negation) gate determines if an input element should be negated (gate value of 1) or simply passed along (gate value of 0). 
    \begin{align}
        \hat{x}_{i,o} &= (1 -\sigmoid(G^\textrm{NEG}_{i,o})) \cdot x_{i,o} + \sigmoid(G^\textrm{NEG}_{i,o})\cdot(1-x_{i,o})  \;.
    \end{align}
    \item \textbf{OR calculation}, which applies disjunctions (weight value of 1) for the output of the input gating. This can also be used to model addition and subtraction.
    \begin{align}
        z_o^\textrm{OR} &= \bigotimes_{i=2}^{I}(\begin{bmatrix} 1 & -A_{i,o}\cdot\hat{x}_{i,o} \end{bmatrix} \otimes \begin{bmatrix} -1 & A_{1,o}\cdot\hat{x}_{1,o} \end{bmatrix}) \bm{1} + 1 \;.
    \end{align}
    \item \textbf{AND calculation}, which applies conjunctions (weight value of 1) over the output of the input gating. This can also be used to model multiplication. The definition is the same as the $\mathrm{NAC}_{\bullet}$ (Equation~\ref{eq:NAC*}) used in the NALU. 
    \begin{align}
         z_o^\textrm{AND} &= \exp\left(\sum_{i=1}^I (A_{i,o}\cdot \ln(|\hat{x}_{i,o}| + \epsilon))\right)  \;.
    \end{align}
    \item \textbf{Output gating}, which determines whether an output value should use the OR calculation (gate value 0) or AND calculation (gate value 1). 
    \begin{align}
        \hat{y}_{o} &= (1 -\sigmoid(G^\textrm{OUT}_{i,o})) \cdot  z_o^\textrm{AND} + \sigmoid(G^\textrm{OUT}_{i,o})\cdot  z_o^\textrm{OR})  \;.
    \end{align}
\end{itemize}
 
Three parameter matrices require to be learnt. One for learning the gate values for negation ($\bm{G^{\textrm{NEG}}}$), another for learning the (shared) weight values for the AND and OR calculations ($\bm{A}$) and one for learning the gate values for the output ($\bm{G^{\textrm{OUT}}}$). 

Optionally, the application of De-Morgan laws which enables representing a conjunction using only negation and disjunction and represent disjunction using only negation and conjunction, makes it possible to modify the architecture to only need either the AND or OR calculation block. 
The changes require removing the unwanted calculation block and replacing the output gate with a negation gate. 
Using only the negation and conjunction operators is favoured as the implementation of disjunction requires using the Kronecker product which scales poorly with input size. 

\subsection{NSR}
\begin{figure*}[!h]
\centering
\includegraphics[width=\textwidth]{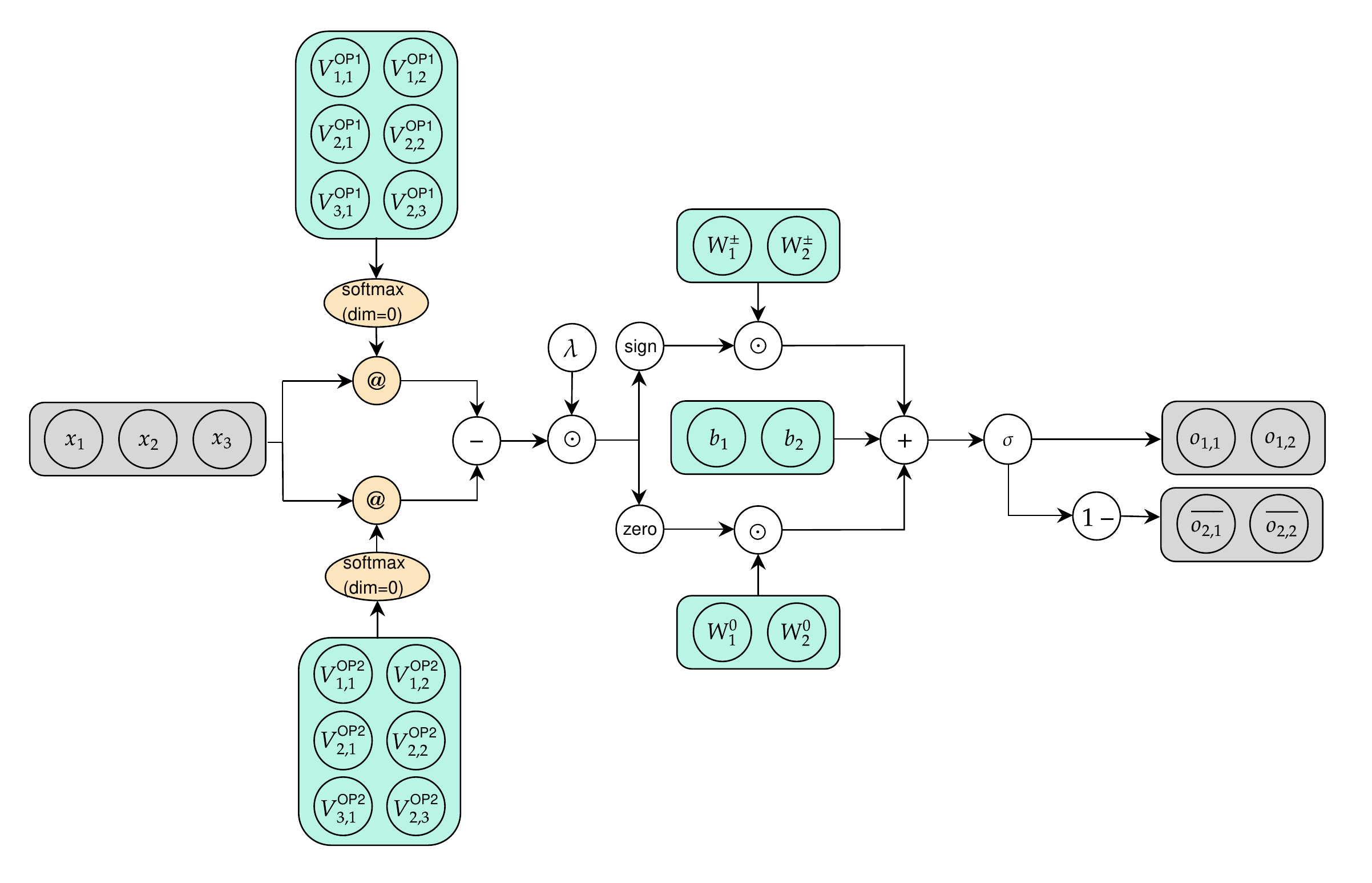}
\caption{NSR architecture. Example of a 3-feature input and 2-feature output model.}
\label{fig:arch-nsr}
\end{figure*}

The \textbf{NSR} (inspired by physical status registers found in the Arithmetic Logic Unit's of computers), models comparison based control logic: $<$, $>$, $!=$, $=$, $>=$, $<=$. 
Simply put, the NSR does quantitative reasoning by learning what input elements to compare and how to compare them. A NSR will output two elements. The first represents if the comparison is true (or false) and the second is the negation of the first output (i.e., $1-o_1$). 
The negation is given such that when the NSR is used in downstream task, the other layers can have access to either branch of the comparison.  
To do this, the NSR architecture does the following: 
\begin{enumerate}
    \item Learns two matrices ($\bm{V^{\textrm{OP1}}}$ and $\bm{V^{\textrm{OP2}}}$) whose purpose is to select two inputs to be operands ($\bm{\widehat{x^{\textrm{OP1}}}}$ and $\bm{\widehat{x^{\textrm{OP2}}}}$) of the comparison function. 
    \begin{align}
        \widehat{x_o^{\textrm{OP1}}} &= \sum_i^I(x_i\cdot \softmax(V_{i,o}^{\textrm{OP1}})) \\
        \widehat{x_o^{\textrm{OP2}}} &= \sum_i^I(x_i\cdot \softmax(V_{i,o}^{\textrm{OP2}}))
    \end{align}
    \item Takes the difference of the two selected operands.
        \begin{align}
            \widehat{x_o} &= \widehat{x_o^{\textrm{OP1}}} - \widehat{x_o^{\textrm{OP2}}}
        \end{align}
    \item Scales the difference with a hyperparameter ($\lambda$) to avoid vanishing gradients. Authors indicate an inverse relation between $\lambda$ and the difference of the input values which can be used to set the $\lambda$ value \citep[Figure~5]{faber2020neural}.
    \begin{align}
        \widehat{x_o} &= \lambda \cdot \widehat{x_o}
    \end{align}
    \item Calculates the sign bit ($\bm{\widehat{x^{\pm}}}$) and zero bit ($\widehat{\bm{x^{0}}}$) of the difference value using smooth continuous functions. 
    \begin{align}
        \widehat{x_o^{\pm}} &= \tanh(\widehat{x_o}) \\
        \widehat{x_o^{0}} &= 1 - 2\tanh(\widehat{x_o})^2
    \end{align}
    \item Learns a scale value (for each bit) and shared shift value. 
    \item Applies the scale and shift to the bit values, takes the sum of the results and passes the result through a $\sigmoid$. The resulting value represents the probability of the comparison being true/false.
    \begin{align}
        z_o &= \widehat{x_o^{\pm}} \cdot W^{\pm}_{i,o} + \widehat{x_o^{0}} \cdot W^{0}_{i,o} + b_o \\
        y_o &= \sigmoid(z_o)
    \end{align}
    \item Returns as output the comparison value and its negation value ($1-y_o$). 
\end{enumerate}
Given two inputs (relevant for the comparison), the NSR will compute the sign and zero bit of the difference of the two operands. 
The sign and zero bit definitions are continuous relaxations of the discrete definitions which rescale the bounds to avoid the gradients of partial derivatives becoming zero. That is:
$
\widehat{x_o^{\pm}} = 
    \begin{cases}
       1 & \textrm{if } \widehat{x_o} > 0 \\
       0 & \textrm{if } \widehat{x_o} = 0 \\
       -1 & \textrm{if } \widehat{x_o} < 0 
    \end{cases} \quad
$
and 
$
\quad
\widehat{x_o^{0}} = 
    \begin{cases}
       1 & \textrm{if } \widehat{x_o} = 0 \\
       -1 & \textrm{if } \widehat{x_o} \neq  0 
    \end{cases} .
$

To improve robustness to different initialisations, the NSR also implements redundancy which learns multiple independent operand pairs ($\widehat{x_o^{\textrm{OP1}}}$, $\widehat{x_o^{\textrm{OP2}}}$) in parallel for each output element. Each pair will have its own sign and zero bit, hence learning its own set of scale and shift values. These independent paths get aggregated together by summing the different $z_o$'s together. 

\section{NALU's Shortcomings and Existing Solutions} \label{sec:shortcomings-and-solutions}
We detail the weaknesses of NALU and explain existing solutions. We focus especially on the iNALU, NAU, NMU and NPU when looking at solutions, as these modules focus on overcoming the shortcomings of NALU. 
A summary of the discussed NALU issues and proposed solutions is given in Table~\ref{tab:nalu-negs}.

\begin{table*}[p]
\centering
\begin{threeparttable}
\begin{adjustbox}{width={\textwidth},totalheight={\textheight-0.75cm},keepaspectratio}
\begin{tabular}{p{3.5cm}p{2.5cm}p{2.5cm}p{2cm}p{2cm}}   \toprule 
\diagbox[height=3\line]{\textbf{Short-}\\\textbf{coming}}{\\\textbf{NALM}} & \textbf{NAU/NMU}  & \textbf{iNALU} & \textbf{NPU/Real NPU} & \textbf{CalcNet (G-NALU)} \\\midrule
\textbf{$\bm{NAC}_{\bullet}$ cannot have negative inputs/targets}          
    & NMU: Remove log-exponent transformation           & Sign correction (mixed sign vector)            & Sign retrieval              & Fixed rules and sign parsing                            \\ \hline
\textbf{Convergence of gate parameters}                                   
    & Stacking instead of gating                                     & Independent gating, separate weights per sub unit and regularisation loss         & -                            & -                                                     \\ \hline
\textbf{Fragile initialisation}                                           
    & Theoretically valid initialisation scheme    & Reinitialise model                             & -                            & -                                                     \\ \hline
\textbf{Weight inductive bias of \{-1,0,1\} not met (non-discrete solutions)} 
    & Regularisation loss term and clipping                      & Regularisation loss term   & (see below)\tnote{*}                            & -                                                     \\ \hline
\textbf{Gradient propagation}                                             
    & Linear weight matrix      & $NAC_{\bullet}$ clip and gradient clip                             & Relevance gating                            & Replace sigmoid and tanh exponent's with golden ratio \\ \hline
\textbf{Singularity (values close to 0)}                                  
    & NMU: Remove log-exponent transformation           & $NAC_{\bullet}$ clip                  & Complex space transformation and relevance gating & -                                                     \\ \hline
\textbf{Compositionality}                                  
    & -                                            & -                                              & -                             & Parsing algorithm                                     \\\bottomrule 
\end{tabular}
\end{adjustbox}

\begin{tablenotes}
  \item[*] The NPU and Real NPU supports fractional weights (e.g., 0.5 representing square-root) and therefore does not enforce discretisation.
\end{tablenotes}

\caption{Summarised NALU shortcomings and existing proposed solutions.}
\label{tab:nalu-negs}
\end{threeparttable}
\end{table*}

\subsection{Mixed Sign Inputs and Negative Outputs}
The $\mathrm{NAC}_{\bullet}$ cannot deal with mixed sign inputs/negative outputs. 
Equation~\ref{eq:NAC*} requires converting negative inputs into their positive counterparts because the log transformation cannot evaluate negative values. 
Therefore the sign of the input is lost, causing the $\mathrm{NAC}_{\bullet}$ to be unable to have negative target values. 
The use of an exponent also causes the inability to have negative outputs, as the range of an exponent is $\mathbb{R}_{>0}$. 
To allow for negative targets, a module can incorporate logic to deal with assigning the correct sign to the output such as the iNALU's sign correction mechanism~\citep{schlor2020inalu} or the NPU's inherent sign retrieval~\citep{heim2020neural}.

The sign correction mechanism creates a mixed sign vector ($\bm{msv}$) $\in \mathbb{R}^{O \times 1}$,
consisting of elements $\{-1,1\}$ (assuming $\bm{W}$ has converged to integers $\{-1,0,1$\}), where each element represents the correct sign for each output element.\footnote{Notice the similarity in calculation between the NMU (Equation~\ref{eq:nmu}) and iNALU's $\bm{msv}$ (Equation~\ref{eq:inalu-msv}).} 
The $\bm{msv}$ is element-wise multiplied to Equation~\ref{eq:NAC*} resulting in applying the relevant sign to the outputs of the multiplicative sub-unit. 
The $+1 - |W_{i,o}|$ means unselected inputs ($W_{i,o}=0$) will avoid affecting the final sign value, as they will only multiply the $msv_o$ value by 1. 
An alternate way to view the $\bm{msv}$ is as a gating mechanism, $sign(x_i) \textcolor{red}{\cdot |W_{i,o}|} + 1 \textcolor{blue}{ \cdot (1 - |W_{i,o}|)}$, where a \textcolor{red}{on gate} ($W_{i,o}= -1/1$) gives the sign and an \textcolor{blue}{off gate} ($W_{i,o}=0$) returns 1.

In the case of a RealNPU, the latter half of its definition (in matrix form), that is $\odot \, \cos(\bm{W^{\textrm{RE}}}\bm{k})$,
can be interpreted as a sign retrieval mechanism. $\bm{k}$ represents positive inputs as 0 and negative inputs as $\pi$ (assuming the gate value converged to select the input). 
Assuming convergence, $\bm{W^{\textrm{RE})}}$ values are $\{1,-1\}$ representing $\{\times,\div\}$. 
Two outcomes are possible from evaluating the expression: \textendash $\cos(\pm\pi)=-1$ or $\cos(0)=1$ where the output value represents the sign of the input value. 

Alternatively, it is possible to remove the need for transformations in the log/exponent space in Equation~\ref{eq:NAC*}, as \citet{madsen2020neural} does for defining the NMU (Equation~\ref{eq:nmu}). This means negative targets can be expressed because the sign is no longer removed from the input. 

\subsection{Gating Parameter Convergence}
The NALU gate, responsible for selection between the $\mathrm{NAC}_{+}$ and $\mathrm{NAC}_{\bullet}$ modules, is unable to converge reliably. 
This is due to the different convergence properties of the $\mathrm{NAC}_{+}$ and $\mathrm{NAC}_{\bullet}$~\citep[Appendix~C.5]{madsen2020neural}, which results in gate weights that converge to a discrete value but not the correct value. For example, converging to a 1 when the true gate value is 0 and visa-versa. 
In cases where the correct gate is selected, the NALU still fails to converge consistently~\citep[Appendix~C.5]{madsen2020neural} implying additional architectural issues exist for the unit. 
Alternatively, partial convergence of gate (and weight) parameters leads to a leaky gate effect \citep{schlor2020inalu}, where non-discretised parameters leads to non-optimal solutions. 
Such solutions may perform well on the interpolation data used during training but will not generalise to OOD data. 
This issue is amplified when additional NALU modules are stacked. 

Even when using the improved NAU and NMU modules, gating still leads to inferior results, therefore \citet{madsen2020neural} replace module gating with module stacking.
\citet{schlor2020inalu} suggests using separate weights for the iNALU sub-units (Equations~\ref{eq:inlau-Wa} and \ref{eq:inlau-Wm}) to improve convergence, and independent gating (removing $\bm{x}$ from Equation~\ref{eq:g}) so learning $\bm{g}$ is no longer influenced by input (see Equation~\ref{eq:inalu-g}). 
Removing the dependence of the input removes contradictory constraints on the gating that would lead to unoptimal solutions. 
Taking the example given by \citet[Section~3.3.6]{schlor2020inalu}, imagine two different input samples $\bm{x_1} = [2,2]$ and $\bm{x_2} = -\bm{x_1} = [-2,-2]$ and the task of adding, i.e., calculating $2+2=4$ for the first input and $-2-2=-4$ for the second. 
Assuming we use the NALU gating method, it implies that $\bm{g} = \sigmoid(\bm{x_1}\bm{G}) = \sigmoid(\bm{x_2}\bm{G})$ = $\sigmoid(-\bm{x_1}\bm{G})$. 
However, as $\bm{x_1} \neq \bm{x_2}$, the above statement is invalid. 

\subsection{Bias Considerations} 
\begin{figure*}
    \vskip 0.1in
    \centering
    \subfigure{\includegraphics[width=0.3\textwidth]{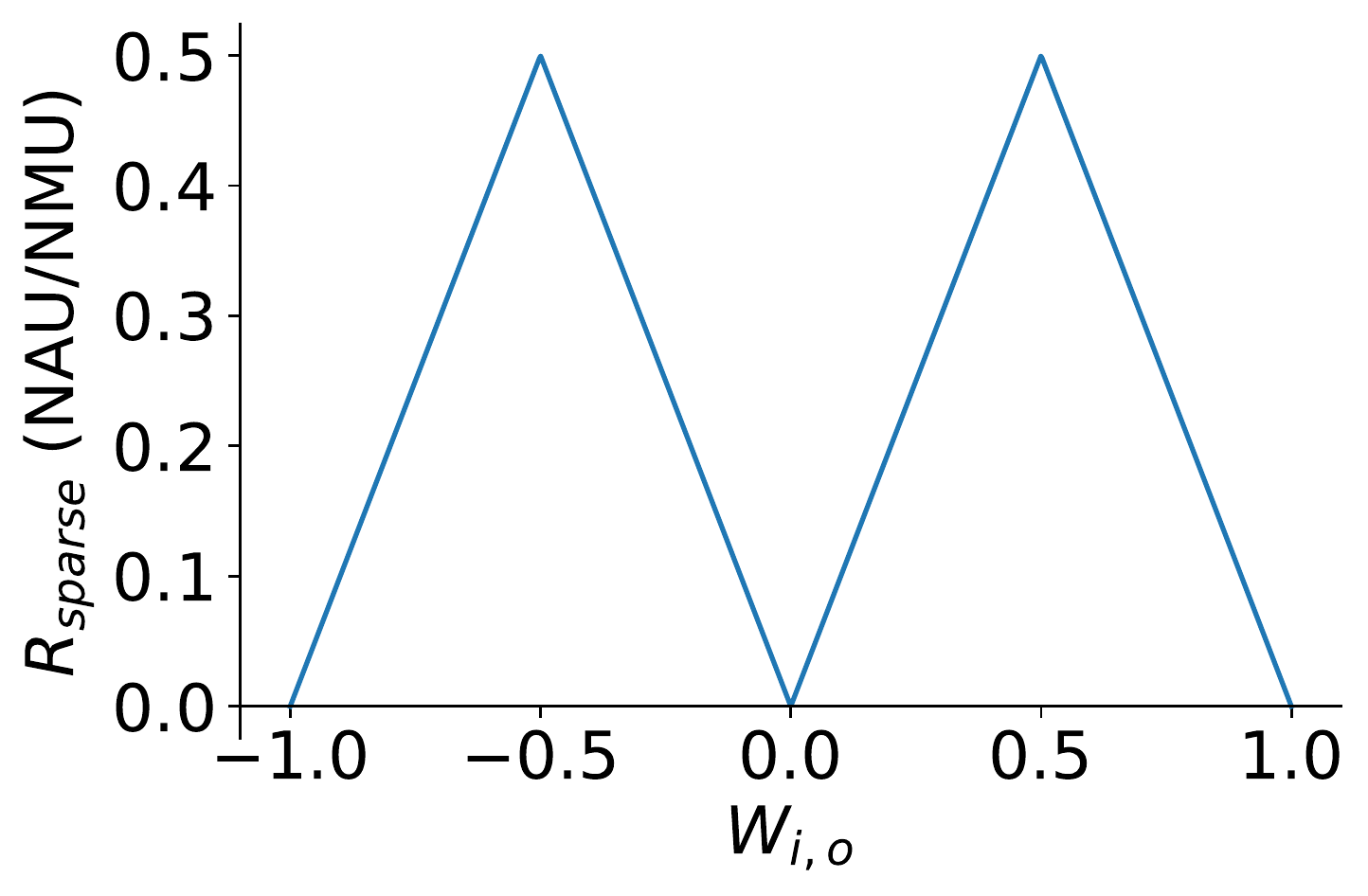}}
    \hspace{0.01\textwidth}
    \subfigure{\includegraphics[width=0.3\textwidth]{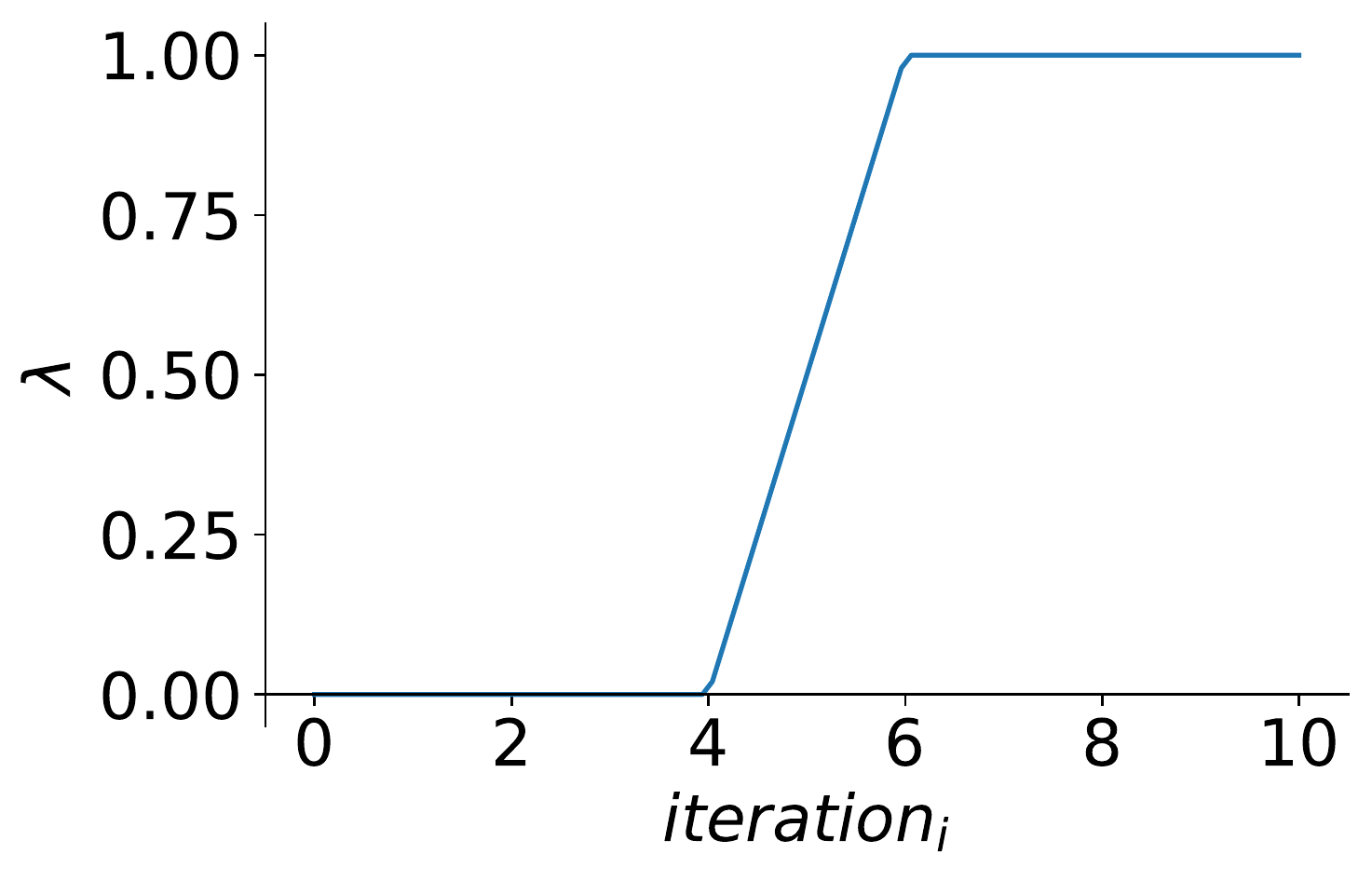}}
    \hspace{0.01\textwidth}
    \subfigure{\includegraphics[width=0.3\textwidth]{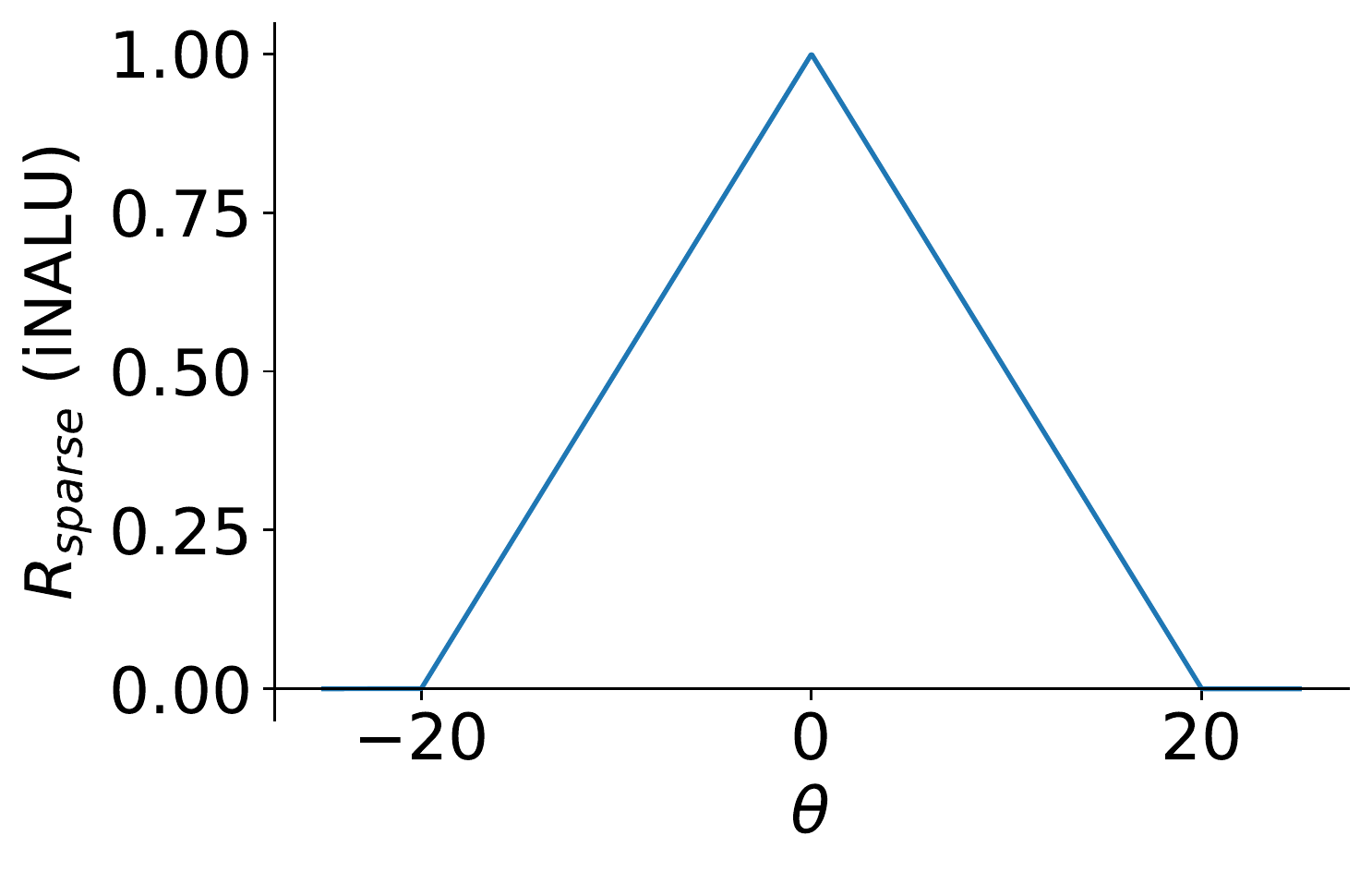}}
    \caption{
    Regularisation functions used to induce sparsity in learnable parameters. 
    Left: Sparsity regularisation used on the NAU and NMU (see Equation~\ref{eq:madsen-rsparse}), forcing values towards $\{-1,0,1\}$. 
    Middle: Scaling function (Equation~\ref{eq:regualizer-scaling}) to control the importance of the sparsity regularisation for the NAU and NMU. For this example, $\hat{\lambda}$ is set to 1 and the scale factor will grow between iterations 4 to 6. 
    Right: Sparsity regularisation for a single parameter used on the iNALU (see Equation~\ref{eq:inalu-rsparse}).}
    \label{fig:reg-schemes}
    \vskip -0.1in
\end{figure*}
The weight biases are achieved by adding a regularisation term for sparsity and using weight clamping~\citep{madsen2020neural, schlor2020inalu}. 
The regularisation penalty encourages weights to converge to the discrete values. 
An illustrative example of the \citet{madsen2020neural, schlor2020inalu} regularisation functions are found in Figure~\ref{fig:reg-schemes}. 
\citet{madsen2020neural} use sparsity regularisation (Equation~\ref{eq:madsen-rsparse}) to enforce the relevant biases for both NAU $\{-1,0,1\}$ and NMU $\{0,1\}$. 
Note that the absolute of $W_{i,o}$ is not necessary when using NMU. 
The regularisation activates and warms up over a predefined period of time to avoid overpowering the main mean squared error loss term (Equation~\ref{eq:regualizer-scaling}). 
Clamping is also applied to the weights beforehand to the ranges of the desired biases. 
iNALU uses a piece-wise function (Equation~\ref{eq:inalu-rsparse}) for regularisation on weight ($\widehat{\bm{W}^{\textrm{A}}}$, $\widehat{\bm{M^{\textrm{A}}}}$, $\widehat{\bm{W}^{\textrm{M}}}$, $\widehat{\bm{M^{\textrm{M}}}}$) and gate ($\bm{g}$) parameters 
to encourage discrete values that do not converge to near-zero values. 
Intuitively, this regularisation penalises the parameter to encourage it to move towards -t or t. Therefore, by having a large positive/negative value, when the parameter goes through a $\sigmoid$ or $\tanh$ activation (see Equations~\ref{eq:inlau-Wa}, \ref{eq:inlau-Wm} and \ref{eq:inalu-g}), the resulting value will be close to $\{-1,0,1\}$. 
Rather than a warmup period, regularisation occurs only once the loss is under a pre-defined threshold and stops once a discretisation threshold $t=20$ is met.

\subsection{Initialisation Considerations}
Good initialisations are crucial for convergence. 
Assuming the \citet{madsen2020neural} implementation of NALU is used for initialisation, then weight matrices are from a uniform distribution with the range calculated from the fan values,\footnote{\url{https://github.com/AndreasMadsen/stable-nalu/blob/2db888bf2dfcb1bba8d8065b94b7dab9dd178332/stable_nalu/layer/nac.py\#L22}} and the gate matrix from a Xavier uniform initialisation with a sigmoid gain.\footnote{\url{https://github.com/AndreasMadsen/stable-nalu/blob/2db888bf2dfcb1bba8d8065b94b7dab9dd178332/stable_nalu/layer/_abstract_nalu.py\#L90}} 
However, empirical results show difficulty for both optimisation and robustness with such initialisations. 
Fragility in optimisation results in converging to the expected parameter value difficult to achieve~\citep{madsen2020neural}, especially when redundant inputs that require sparse solutions exist. 
When redundancy exists, non-sparse solutions are not extrapolative, lacking transparency.

To ease optimisation, \citet{madsen2020neural} use a linear weight matrix construction (removing the need of non-linear transformations), while \citet{schlor2020inalu} use clipping on the $\mathrm{NAC}_{\bullet}$ calculation (see Equation~\ref{eq:inalu-mo-clip}). 
The minimum of the input is clipped to $\epsilon=10^{-7}$ and the result of the $\ln$ operation is clipped to be at most $\omega=20$. 
Using this clipping allows to avoid exploding intermediary results. 
Additionally, gradient clipping is used to avoid exploding gradients.  
\begin{figure*}
    \vskip 0.1in
    \centering
    \subfigure{\includegraphics[width=0.45\textwidth]{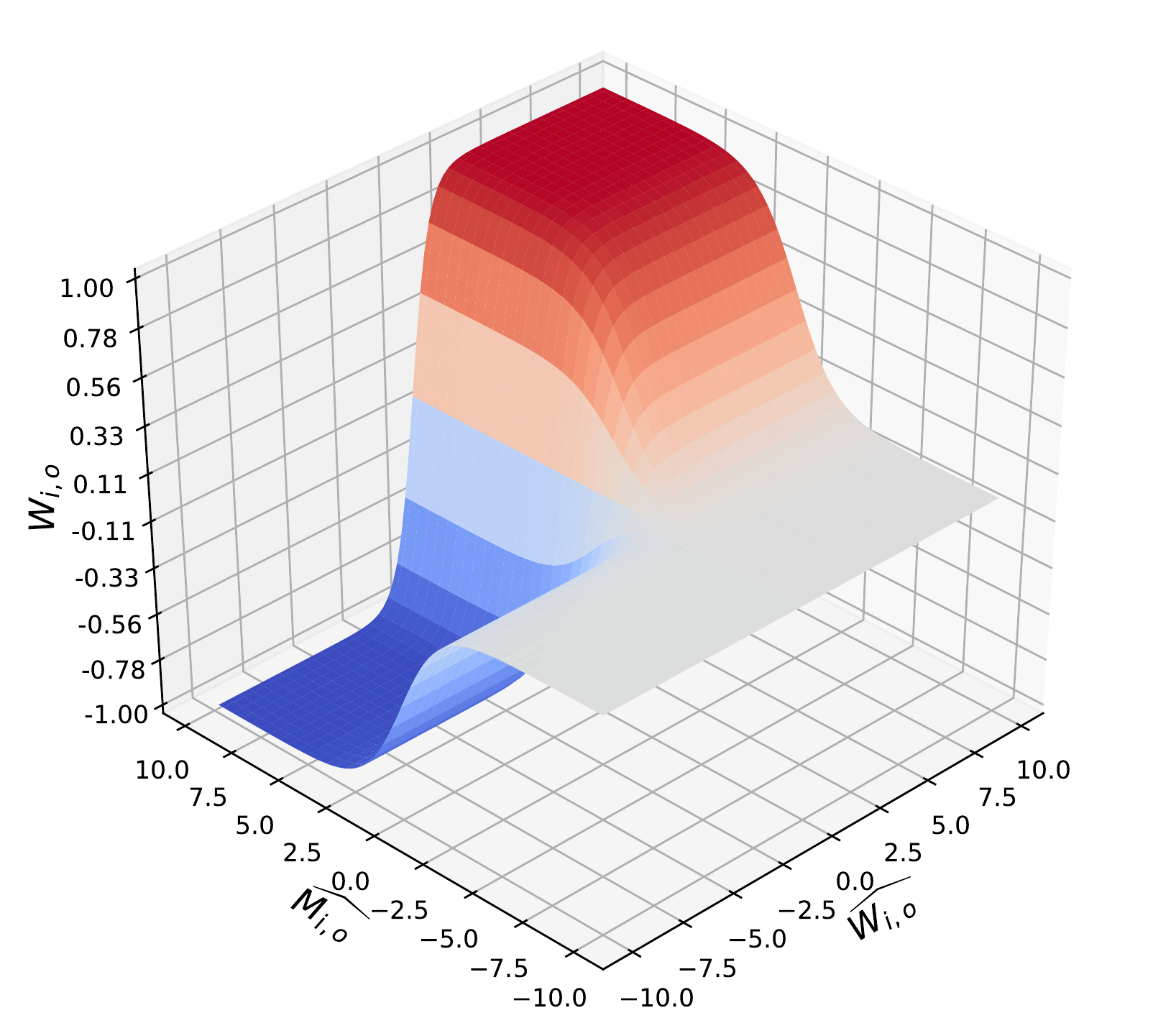}}
    \hspace{0.05\textwidth}
    \subfigure{\includegraphics[width=0.45\textwidth]{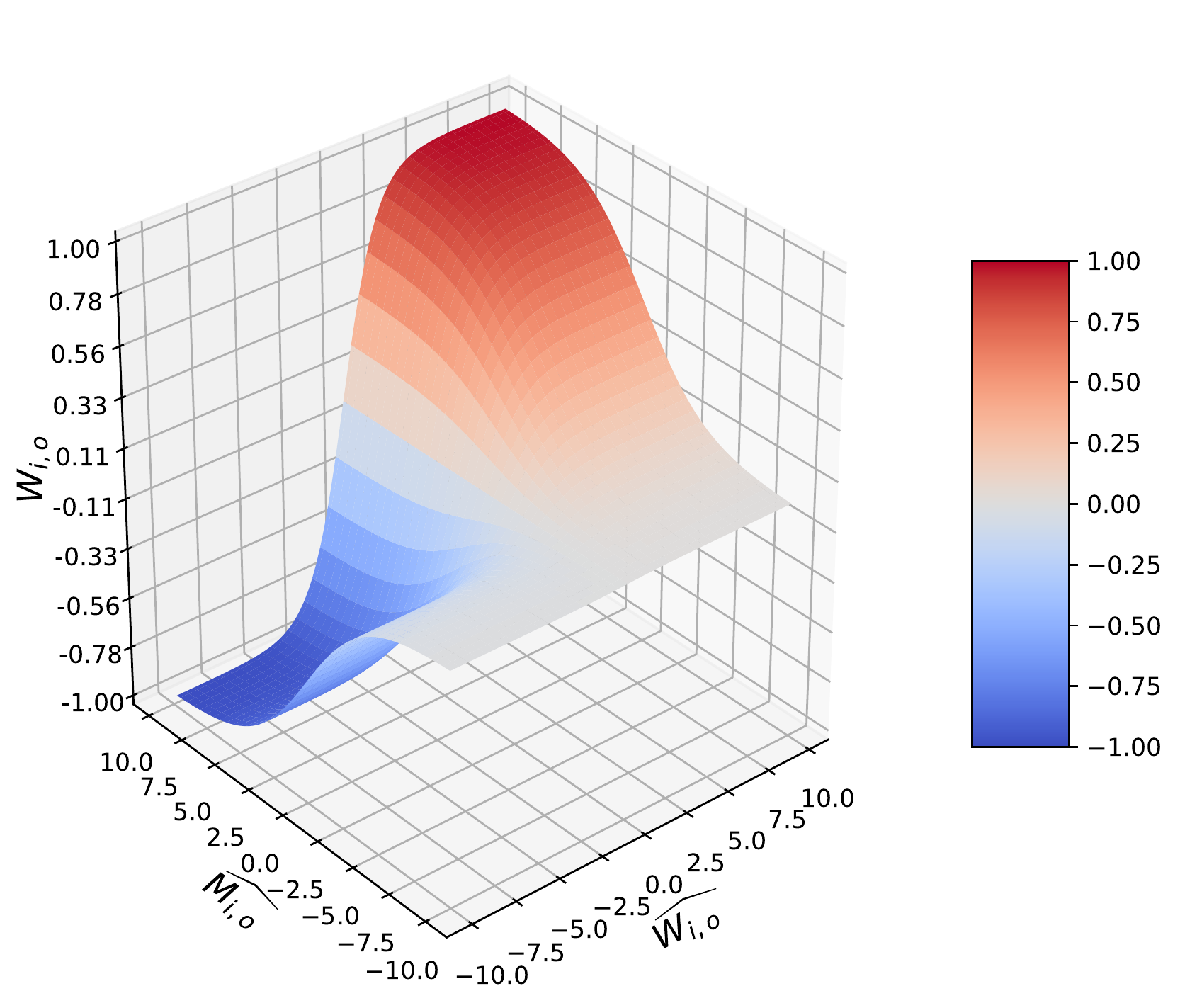}}
    \caption{Adapted from \citet[Figure~2 and Figure~4]{rana2019exploring} for showing the values for $\bm{W}$ used in NALU calculated over the domain of $\widehat{\bm{W}}$ and $\widehat{\bm{M}}$. 
    Left: Using NALU's calculation of $\bm{W}$ where $\tanh$ and $\sigmoid$ are calculated with base $\textrm{e}$. 
    Right: G-NALU's calculation of $\bm{W}$ where $\tanh$ and $\sigmoid$ are calculated with a golden ratio base resulting in smoother value transition.
    }
    \label{fig:gNALU-W}
    \vskip -0.1in
\end{figure*}

\citet{rana2019exploring} modify the non-linear activation's of the weight matrices in the NALU for smoother gradient propagation as shown by Figure~\ref{fig:gNALU-W}. 
In contrast, in attempts to avoid falling into a local optima, iNALU allows multiple reinitialisations of a model during training to counteract the non-optimal initialisation in NALU which contribute to vanishing gradients and convergence to local minimas. 
Reinitialisation occurs every $m^{th}$ epoch if the following two conditions are met: (1) the loss has not improved in the last n steps, (2) the loss is larger than a pre-defined threshold. 
The main disadvantage reinitialising multiple times during training is that it can require running more iterations which may be infeasible. For example, for a standalone NALM it is possible to keep reinitialising until a reasonable solution is found, however if the NALM is used as a subcomponent in a larger neural network then reinitialisation can be too costly. 
Through a grid search they find having the mean of the gate and NALU weight matrices $\widehat{\bm{M}}$, $\widehat{\bm{W}}$ initialised to be 0, -1 and 1 respectively, results in the most stable modules. 
However, even when using such initialisations, the stability problem remains for division \citep[Table~1]{schlor2020inalu}.

\subsection{Division}
\begin{figure}[t]
    \vskip 0.15in
    \begin{center}
    \centerline{\includegraphics[scale=0.5]{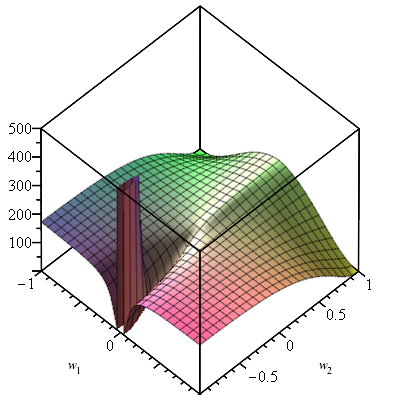}}
    \caption{Taken from \citet[Figure~2b]{madsen2020neural}. An example illustration of the unstable optimisation issue arising when using a stacked $\mathrm{NAC}+$  $\mathrm{NAC}_{\bullet}$ with $\epsilon=0.1$. The plot represents the root mean squared loss surface when modelling $(x_1+x_2)\cdot(x_1+x_2+x_3+x_4)$ for the input $[1,1.2,1.8,2]$. $w_1$ and $w_2$ represent the weight values to use for the $\mathrm{NAC}+$ and $\mathrm{NAC}_{\bullet}$ weight matrices such that
    $\mathbf{W}_1 = \left[\protect\begin{smallmatrix}
        w_1 & w_1 & 0 & 0 \\
        w_1 & w_1 & w_1 & w_1
    \protect\end{smallmatrix}\right]$ and 
    $\mathbf{W}_2 = \left[\protect\begin{smallmatrix}
        w_2 & w_2
    \protect\end{smallmatrix}\right]$
    .}
    \label{fig:madsen-singularity}
    \end{center}
    \vskip -0.15in
\end{figure}
Division is NALU's weakest operation~\citep{trask2018neural}. 
Having both division and multiplication in the same module causes optimisation difficulties. 
\citet{madsen2020neural} highlight the singularity issue (caused from division by 0 or values close to 0 bounded by an epsilon value) in the $\mathrm{NAC}_{\bullet}$ which causes exploding outputs (see Figure~\ref{fig:madsen-singularity}). 
This issue is amplified due to operations being applied in log space. 
The NMU removes the use of log, therefore is not epsilon bound. Furthermore, the NMU is only designed for multiplication.
The NPU takes \citet{madsen2020neural}'s interpretation of multiplication (using products of power functions) but applies it in a complex space enabling division and multiplication~\citep{heim2020neural}. 
Though the NPU cannot fully solve the singularity issue as a log transformation is still applied to the inputs, the relevance gating (see Equation~\ref{eq:npu-r}) aids in smoothing the loss surface to provide better convergence.
\citet{schlor2020inalu} observe that reinitialising modules numerous times during training can still lead to failure, implying that the issue lies in unit architecture as well as initialisation. 
Hence, division remains an open issue. 

\subsection{Compositionality}
A single NALU is unable to output expressions whose operations are from both $\{+, -\}$ and $\{\times, \div\}$, for example $x_1 + x_2 * x_3$.
\citet{bogin2020latent} hint at NALU's inflexibility to learn different expressions from same weights as once trained the learnt expression of a NALU is static meaning that expressions with a different ordering of operations will not work. 
\citet{rana2019exploring} develop CalcNet, a parsing algorithm, such that the expression to learn is decomposed into its intermediary sub-expressions which obey the rules of precedence (i.e., BIDMAS) and then is solved in a compositional manner. 
However, decomposition requires fixed rules and pre-trained sub-units which are undesirable because in order to decompose, the input must also contain the operations used, meaning that model is exposed to a priori relating to the underlying function. 

\section{Experiments and Findings of Modules for Arithmetic Tasks}\label{sec:exp-and-findings}
To better understand the existing evaluation of modules, we go through the experiments used in the papers for: NALU, iNALU, NAU, NMU, and NPU. 
We begin by indicating inconsistencies across papers for the two-layer arithmetic task setup, highlighting the different evaluation techniques used by each paper, encouraging the need of task standardisation. 
Inter-module comparison using existing findings is made to infer the best module per operation. 
We end this section by introducing a \textit{Single Module Arithmetic task} to act as a standardised benchmark for comparison against all existing arithmetic NALM modules. 

\subsection{Why are the Square and Square-Root Operations not included in this Discussion?}
Though mentioned in \citet{trask2018neural} that the NALU can learn to model square and square-rooting, we will purposefully avoid analysing the ability of the multiplicative modules to do square ($a^2$) and square-root ($\sqrt{a}$) operations. Rather, we focus only on the four core arithmetic operations: addition, subtraction, multiplication and division. 

Of the NALMs described previously in Sections~\ref{sec:nalu-arch}~and~\ref{sec:other-units} only the NALU, G-NALU, NPU and Real NPU are able to model squaring and square-rooting (without modifications to the original architecture/training methodology). 
The other multiplicative modules (NMU and iNALU) would have difficulty in modelling these operations which is explained below.

The squared operation can be solved when using a multiplication module in two ways, differing by the way the input is represented. 
The first way expects two inputs of the same value, which essentially results in a multiplication ($a \times a$), and the second way expects one input using it as the base and applying it with an exponent of two ($a^2$). 
The second way requires using a weight value of two (to correspond with the exponent), but this breaks the inductive bias on many of the modules that weights should have a magnitude up to 1 which is enforced using clipping. 
Therefore, we avoid analysing the square operation. 

As for the square-root operation an interpretable weight value corresponds to 0.5 resulting in square-rooting being modelled as $a^{\frac{1}{2}}$.  This contradicts the inductive bias of discrete weights of the NMU and iNALU which is enforced using regularisation penalties. Therefore, we avoid analysis square-root operation. 

\subsection{Two Layer Arithmetic Task}
\begin{figure}[t]
\centering
\includegraphics[width=\columnwidth*4/5]{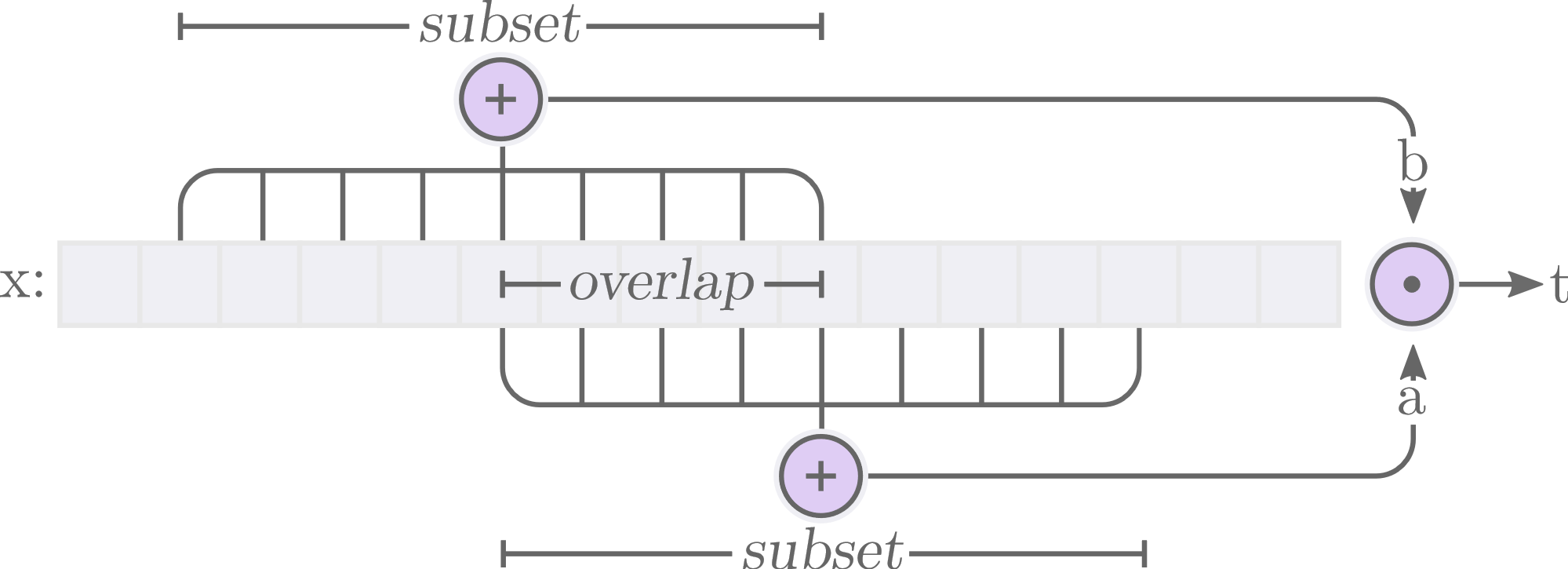}
\caption{Taken from \citet[Figure~6]{madsen2020neural}. Illustration on how to get from input vector to target scalar for the Arithmetic Dataset Task. This setup is solved using a stacked addition-multiplication module.}
\label{fig:DAT}
\end{figure}

A task consistently used in all papers is the `\textit{Static Simple Function Learning}' experiment~\citep{trask2018neural}, which evaluates the ability of a module to learn a trivial two-operation function. \citet{maep-madsen-johansen-2019} introduce their own experiment setup (including details for reproducibility) which they utilise in their later work \citep{madsen2020neural} under the name `\textit{Arithmetic Datasets}' task. 
Specifically, given an input vector $\mathbb{R}^{100}$ of floats, the first (addition) layer should learn to output two values (denoted a and b) which are the sums of two different partially overlapping slices (\ie, subsets) of the input, and the second layer should perform an operation on a and b. Figure~\ref{fig:DAT} illustrates such an example. \textbf{Due to the rigorous setup, evaluation metrics, and available code, we strongly suggest the \citet{maep-madsen-johansen-2019} experiment be used to test and compare new modules for the Two Layer Arithmetic task.} 

iNALU's experiments 4 (`\textit{Influence of Initalization}') and 5 (`\textit{Simple Function Learning Task}') is a copy of the task but is different to the \citet{madsen2020neural} setup. The experiments calculate a and b differently by not allowing for overlap between a and b, and allowing a and b to be made up of random (instead of consecutive) elements of the input. Also, the interpolation and extrapolation ranges are different. 
\citet{heim2020neural}'s claims that their `\textit{Large Scale Arithmetic}' task is equivalent to the \textit{Arithmetic Dataset} task. However, there are key distinctions between the two meaning the results from the two papers are not directly comparable. 
We highlight in Table~\ref{tab:madsen-vs-heim} differences between the three experiment setups.\footnote{We do not compare \citet{trask2018neural} as no details on the experiment setup is given. We do not compare \citet{rana2019exploring} as they do not include this experiment.} 

\begin{table*}[t]
\begin{center}
\vskip 0.15in
\begin{tabular}{p{3cm}p{3.5cm}p{3.9cm}p{3.5cm}} \toprule
\vspace{0.1cm}\textbf{Property}\vspace{0.1cm}   & \vspace{0.1cm}\textbf{\citet{madsen2020neural}}\vspace{0.1cm}              & \vspace{0.1cm}\textbf{\citet{heim2020neural}}\vspace{0.1cm}                &
\vspace{0.1cm}\textbf{\citet{schlor2020inalu}}\vspace{0.1cm}\\ \midrule
Hidden size                     & 2                 & 100       & 2\\ 
Iterations for one run          & 5,000,000         & 50,000    & 100,000\\ 
Number of seeds                 & 100               & 10        & 10\\ 
Learning rates                  & 1e-3              & 1e-2 for addition and 5e-3 for all other operations  & 1e-3\\ 
Subset and overlap ratios       & 0.25 and 0.5      & 0.5 and 0.25 (for addition, subtraction, and multiplication) & $0.3\dot{3}$ and 0\\ 
Division                        & a/b               & 1/a   & a/b\\ 
Interpolation and extrapolation ranges$^*$ 
    & Train: U{[}1,2) for all operations. \newline Test: U{[}2,6).
    & Train: S(-1,1) for addition, subtraction, and multiplication, S(0,0.5) for division. \newline Test: S(-4,4) for addition, subtraction and multiplication, S(-0.5,0.5) for division. 
    & Train: U{[}-3,3{]} and TN$_{(\mu=0,\sigma=1)}${[}-3,3{]}\newline Test: U{[}-5,-5{]} and TN$_{(\mu=3.5,\sigma=\frac{1}{6})}${[}3,4{]} respectively.\\ 
Programming framework                 
    & Pytorch (Python)                                
    & Flux (Julia)
    & Tensorflow (Python)
    \\ \bottomrule
\end{tabular}
\caption{Differences in the `\textit{Large Scale Arithmetic}' task used in the papers \citet{madsen2020neural} and \citet{heim2020neural}. `a' and `b' represent summed slices of the input, and are the expected output values for the addition module. $^*$U=Uniform, S=Sobol and TN=Truncated Normal.}
\label{tab:madsen-vs-heim}
\end{center}
\end{table*}

\subsubsection{Evaluation Metrics}\label{subsec:eval-metrics}
Currently, no de facto method exists for measuring arithmetic extrapolation performance on models.  
The purpose of evaluation metrics is to reflect whether a model solution is the true solution and be able to rank different model solutions against each other. 
We therefore explain the different metrics used in previous works. 
\textit{\cite{trask2018neural}} calculates a score for each model, where the score is the $\frac{\textrm{MSE loss of the model}} {\textrm{MSE loss of a randomly initialised model (with no training)}}$. 
A score of 0 reflects perfect accuracy while a score larger than 100 means the solution is worse than the baseline model. 
Though this method is good for relative rankings between different models, there is no indication to the relative performance against the gold solution~\cite{schlor2020inalu}. 
Furthermore, a randomly initialised model will most likely have poor performance, so the scaled errors of the other models seem better than they actually are. 
\textit{\citet{heim2020neural}} measures the median of the MSE with confidence intervals using the median absolute deviation. 
Compared to the mean, the median is less sensitive to outliers and skewed results, however as a result it discards information about individual errors which can be helpful when considering factors such as the extent of robustness against different initialisations.   
Both \textit{\citet{maep-madsen-johansen-2019, schlor2020inalu}} measure the MSE, but also compare if the MSE is within a threshold value representing the error of an ideal solution to a given precision. 
This threshold comparison produces a success metric in which each seed can be compared in a pass/fail situation which is averaged to a success rate.  

\textbf{Evaluation metrics used on the Arithmetic Dataset Task.} \textit{\cite{maep-madsen-johansen-2019}} extends the use of threshold based success by using: configuration-sensitive success thresholds, two additional metrics to measure speed of convergence and sparsity, and confidence intervals for each metric where each interval calculated using a different distribution family to best match the metric. 
Specifically, there are three evaluation metrics: (1) the success on the extrapolation dataset against a near optimal solution (\textit{success rate}), (2) the first iteration which the task is considered solved (\textit{speed of convergence}), and (3) the extent of discretisation towards the weights' inductive biases (\textit{sparsity error}). 
The sparsity error calculated by $\max\limits_{i,o}(\min(|W_{i,o}|, 1-|W_{i,o}|))$, calculates the NALM weight element which is the furthest away from the acceptable discrete weights for a NALM. For example, for the NMU if a weight was at 0.7 it would get a sparsity error of 0.3. 
A success means the MSE of the trained model is lower than a threshold value (i.e., the MSE of a near optimal solution). 
For the Arithmetic Dataset Task, the threshold is a simulated MSE on 1,000,000 data samples using a model where each weight of the addition is off an optimal weight value by $\epsilon=$1e-5. 
A near optimal solution is used over an optimal solution as it considers accumulated numerical precision errors (a limitation of hardware rather than module architecture). 
Each metric is calculated over different seeds where the total number of seeds should be enough to demonstrate issues on robustness, while keeping computation time reasonable.  
95\% confidence intervals are calculated for each metric. 
The success rate uses Binomial distribution because trials (i.e., run on a single seed) are either pass/fail situations. 
The convergence metric uses a Gamma distribution and sparsity error uses a Beta distribution. 
Both Beta and Gamma can easily approximate the normal distribution and support its corresponding metric. 

\subsection{Additional Experiments}
This section briefly summarises additional experiments given in the NALM papers. 
We do not cross-compare papers for each experiment as there is too little similarities between experiments. 

\citet{trask2018neural} carries out a recurrent version of their static task experiment to test the $\mathrm{NAC}_{+}$, where the subsets a and b are accumulated over multiple timesteps. 
The purpose of this task is to generate much larger output values to test NALU on. 
As well as pure arithmetic tasks, \citet{trask2018neural} tests NALU in other settings such as: translating numbers in text form into the numerical form (for example `two hundred and one' to 201), a block grid-world which requires travelling from point A to B in exactly n timesteps, and program evaluation for programs with arithmetic and control operations. 
However, the NALU is not utilised for its capabilities as a NALM in the text to number task as the NALU is applied to a LSTM's hidden state vector; therefore it is questionable if the arithmetic capabilities of NALU are being used, as the NALU may also have to decode the numerical values from the LSTM vector. 
MNIST is also used to evaluate NALU's abilities on being part of end-to-end applications. 
This includes exploring counting the occurrence of different digits, addition of a sequence of digits, and parity prediction. 

\citet{madsen2020neural} also use MNIST for testing the module's abilities to act as a recurrent module for adding/multiplying the digits. 
\citet{madsen2020neural} additionally provide experiments to express the validity of their modules. This includes modifying the number of redundant hidden units, different input training ranges, ablation on multiplication, stress testing the stacked NAU-NMU against difference input sizes, overlap ratios and subset ratios, showing the failure of gating in convergence, and parameter tuning regularisation parameters. 

\citet{schlor2020inalu} provide three additional experiments. 
Experiment 1 (`Minimal Arithmetic Task') uses a single-layer to do a single operation with no redundancy to see the effect of different input distributions. 
Experiment 2 (`Input Magnitude') sees the effect of training data by controlling the magnitude of the interpolation data. 
NALU fails on magnitudes greater than 1. 
iNALU remains unaffected for addition and subtraction. Multiplication performance is coupled to magnitude where extrapolation error increases with magnitude. 
Division is uncorrelated to the input magnitude. 
To increase problem difficulty, experiment 3 (`Simple Arithmetic Task') introduces redundancy where from 10 inputs only 2 are relevant. 
NALU improves on performance for exponentially distributed data when redundant inputs are introduced. 
iNALU show improvements for multiplication where the module is able to succeed on previously failed training ranges such as an exponential distribution with a scale parameter of 5 (meaning lambda is 0.2) but worsens for division.

\citet{heim2020neural} highlights the relevance gate's use via a toy experiment to select one of the two inputs. They show the relevance gate transforms regions away from the solution which contain no gradient information into regions with more instructive gradients \cite[Figure~3]{heim2020neural}. Additionally, they demonstrate an application of a stacked NAU-NPU module for equation discovery for an epidemiological model.

\subsection{Cross Module Comparison}
We compare existing findings across modules. 
NALU is no longer considered the state-of-the-art for neural arithmetic operation learning. 
For each operation the best module is as follows - \textbf{addition or subtraction}: NAU, \textbf{multiplication}: NMU, \textbf{division}: NPU (or RealNPU if the task is trivial).

iNALU generally outperforms NALU at the cost of additional parameters and complexities to the model. 
The magnitude of iNALU's improvement varies, as \citet{schlor2020inalu} claims vast improvements, while \citet{heim2020neural} claim minor. 
For division both the iNALU and NALU performances remain comparable. Success on multiplication is dependent on the input training range. 
\citet{heim2020neural} states the NMU outperforms iNALU on multiplication (as expected), but also addition and subtraction. 
The reason lies in the architecture used. 
The model is a stacked NAU-NMU meaning the addition/subtraction would be modelled by the NAU. Therefore, the NMU would only be required to act as a selector, selecting the output of the summation (that is, have a single weight at 1 and the rest at 0). 
Therefore, if two NMUs are stacked together we expect the failure in a pure addition/subtraction task as shown in \citet[Appendix~C.7]{madsen2020neural}. Surprisingly the two layer NMU was able to get 56\% success for subtraction, though 0\% success for addition~\citep[Table~6]{madsen2020neural}. 
\citet{heim2020neural} is the only work (at the time of writing this paper) to experimentally compare the main modules mentioned. 
Results show that the NPU outperforms the iNALU for multiplication and division. 
When stacked on top of a NAU, the NPU performs similar to the NMU for addition and subtraction. 
The NPU is outperformed by the NMU for multiplication, however it is more consistent in convergence against different runs.
For addition and subtraction, the NAU-NMU is the sparsest module (having the least number of non-zero weights). 
Arithmetic tasks using the basic arithmetic operation require the weight and gate values to be discrete. 
Regularisation penalties have been a popular approach~\citep{madsen2020neural, schlor2020inalu} to achieve this. 
The NPU uses L1 regularisation for arithmetic tasks, encouraging sparsity over discretisation due to its ability to express fractional powers. 
However, the main influence of causing sparsity in the NPU modules are from using the relevance gating. If this gating is removed (denoted by the NaiveNPU in the experiments), models are consistently less sparse for all operations \cite[Figure~7]{heim2020neural}.

\subsection{Single Module Arithmetic Task}

Having a standardised benchmark is essential for fair comparison of modules. 
As stated previously, so far, no such benchmark exists. 
Therefore, we provide results on a \textit{Single Module Arithmetic Task}, training modules on their respective operations over a range of different interpolation distributions and testing over a range of extrapolation distributions.\footnote{Code is available at: \url{https://github.com/bmistry4/nalm-benchmark}}

\textbf{Why not use the two-layered Arithmetic Dataset Task?} The Arithmetic Dataset Task requires modules to perform three sub-tasks: \textit{selection, operation, stacking}. 
Selection is the ability to deal with input redundancy for both modules (though more-so for the first layer addition module). 
Operation is the ability to carry out the correct operation/s (i.e., addition and multiplication). 
Stacking sees if training can propagate through two layers. 
Even with only two layers, there are already multiple components being assessed in a single task, making it difficult to analyse where issues lie. 
Therefore, to gain a better understanding of individual NALMs, we propose an experiment which evaluates if the operation/s the module specialises in can be learnt. 

\textbf{Setup.} A single module is used. The input size is two and output size is one, hence there is no input redundancy. Hence, the objective is to model: $y = x_1\;\circ \;x_2$ where $\circ \in \{+, -, \times, \div\}$.
We test the: NALU, iNALU, G-NALU, $\mathrm{NAC}_{+}$, $\mathrm{NAC}_{\bullet}$, NAU, NMU, NPU, and Real NPU. 
Each run trains for 50,000 iterations to allow for enough iterations until convergence. 
A MSE loss is used with an Adam optimiser. 
Interpolation (training/validation) and extrapolation (test) ranges are presented in Table~\ref{tab:SLTR-ranges}. 
Early stopping is applied using a validation dataset sampled from the interpolation range. 
Experiment/hyper-parameters set can be found in Appendix~\ref{app:parameters}. 

\begin{table}[t]
\vskip 0.1in
\begin{center}
\begin{small}
\begin{sc}
    \begin{tabular}{ll}
    \toprule
    \textbf{Interpolation} & \textbf{Extrapolation}     \\ \midrule
    {[}-20, -10)           & {[}-40, -20)               \\ 
    {[}-2, -1)             & {[}-6, -2)                 \\ 
    {[}-1.2, -1.1)         & {[}-6.1, -1.2)             \\ 
    {[}-0.2, -0.1)         & {[}-2, -0.2)               \\ 
    {[}-2, 2)              & {[}{[}-6, -2), {[}2, 6){]} \\ 
    {[}0.1, 0.2)           & {[}0.2, 2)                 \\ 
    {[}1, 2)               & {[}2, 6)                   \\ 
    {[}1.1, 1.2)           & {[}1.2, 6)                 \\ 
    {[}10, 20)             & {[}20, 40)                 \\ \bottomrule
    \end{tabular}
\end{sc}
\end{small}
\end{center}
\caption{Interpolation (train/validation) and extrapolation (test) ranges used for the Single Module Arithmetic Task. Data (as floats) is drawn from a Uniform distribution with the range values as the lower and upper bounds.}\label{tab:SLTR-ranges}
\vskip -0.1in
\end{table}

\textbf{Evaluation.} 
We adopt the \citet{maep-madsen-johansen-2019}'s evaluation scheme used for the Arithmetic Dataset Task (explained in Section~\ref{subsec:eval-metrics}) but adapt the expression used to generate the predictions of an $\epsilon$-perfect model ($y_o^\epsilon$). The expression used depends on the operation to model as shown below: 

\begin{align*}
\textbf{Addition:} \, y_o^\epsilon &= (x_1 + x_2) - \left(\sum_{i=1}^{I}|x_i|\right)\epsilon \\
\textbf{Subtraction:} \, y_o^\epsilon &=  (x_1 - x_2) - \left(\sum_{i=1}^{I}|x_i|\right)\epsilon \\
\textbf{Multiplication:} \, y_o^\epsilon &= (x_1x_2)(1-\epsilon)^2 \times \prod_{i\in X_{irr}}(1 - |x_i|\epsilon) \\
\textbf{Division:} \, y_o^\epsilon &= \frac{x_1(1 - \epsilon)}{x_2(1 + \epsilon)} \times \prod_{i\in X_{irr}}(1 - |x_i|\epsilon) 
\end{align*}

Assume $x_1$ and $x_2$ are the operands to apply the operation to and any remaining features ($x_3,...,x_n$) be irrelevant to the calculation and part of the set $X_{irr}$. We use $I$ to denote the total number of input features. 
In each case the $\epsilon$ for each feature will contribute some error towards the prediction. 
A simulated MSE is then generated with an $\epsilon=1e-5$ like in~\citet{maep-madsen-johansen-2019} and used as the threshold value to determine if a NALM converges successfully for a particular range by comparing the NALMs extrapolation error against the threshold value.\footnote{Other variations for generating the threshold exist which are further discussed in Appendix~\ref{app:eps-threshold}.} 

\subsubsection{Results}
We present the NALMs' performances on the four main arithmetic operations. 
Each figure consist of plots for each evaluation metric (success rate, speed of convergence and sparsity error) discussed in the evaluation paragraph above, with confidence intervals calculated over 25 seeds. 

\begin{figure*}[t]
\vskip 0.2in
\centering
\includegraphics[width=\textwidth]{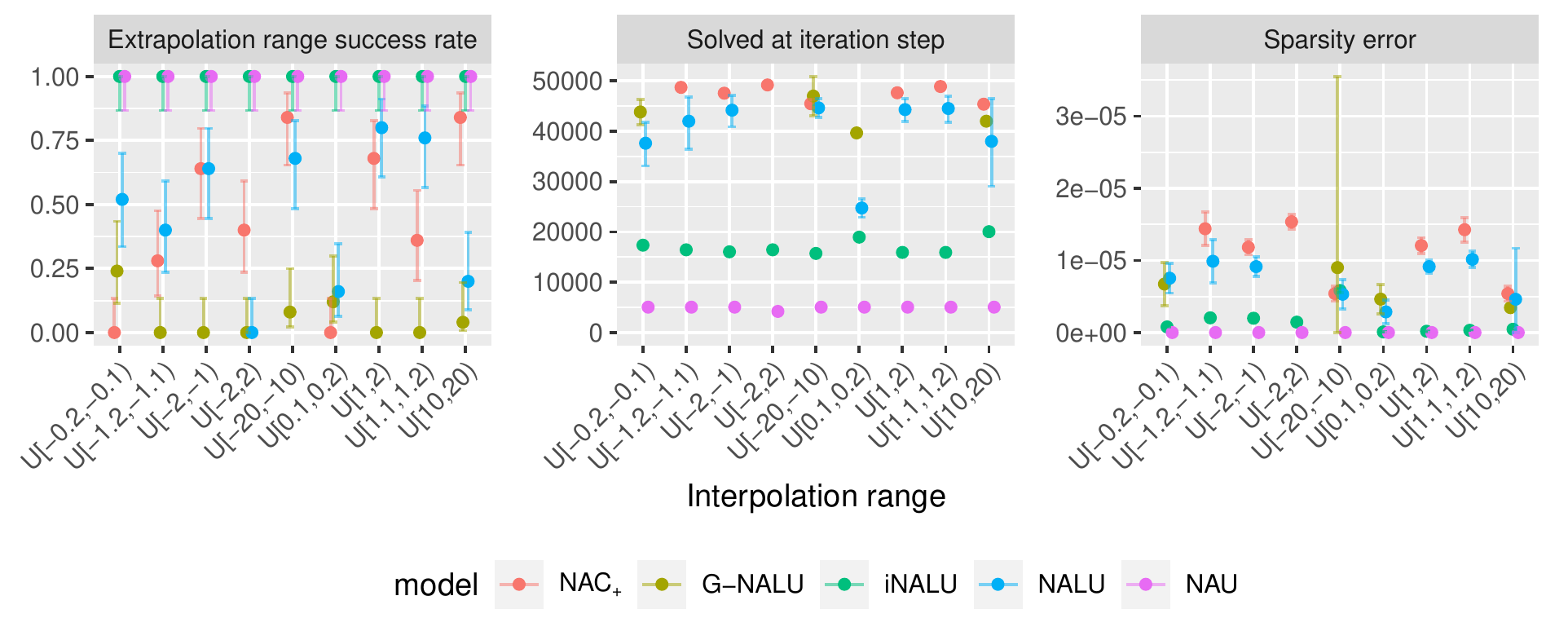}
\caption{Performance on Single Module Task for addition.}
\label{fig:slt-range-add}
\vskip -0.2in
\end{figure*}

\textbf{Addition (Figure~\ref{fig:slt-range-add}).}
The NAU has full success for all ranges correlating to the sparsity errors around 0 meaning that weights successfully converge to the expected value of 1. 
The iNALU also has full success but takes longer to solve and has a slightly larger sparsity error than the NAU. 
The NALU struggles with consistent performance especially for the small positive range ($\mathcal{U}$[0.1,0.2)), large positive range ($\mathcal{U}$[10,20)) and range with both positive and negative inputs ($\mathcal{U}$[-2,2)). The low sparsity error implies that discrete values are being converged to, though not to the correct ones. 
The $\mathrm{NAC}_{+}$ also struggles to obtain consistent results over different ranges like the NALU.  
The G-NALU performs the worst of all the modules obtaining non-zero success on only 4 of the 9 ranges. 

\begin{figure*}[t]
\vskip 0.2in
\centering
\includegraphics[width=\textwidth]{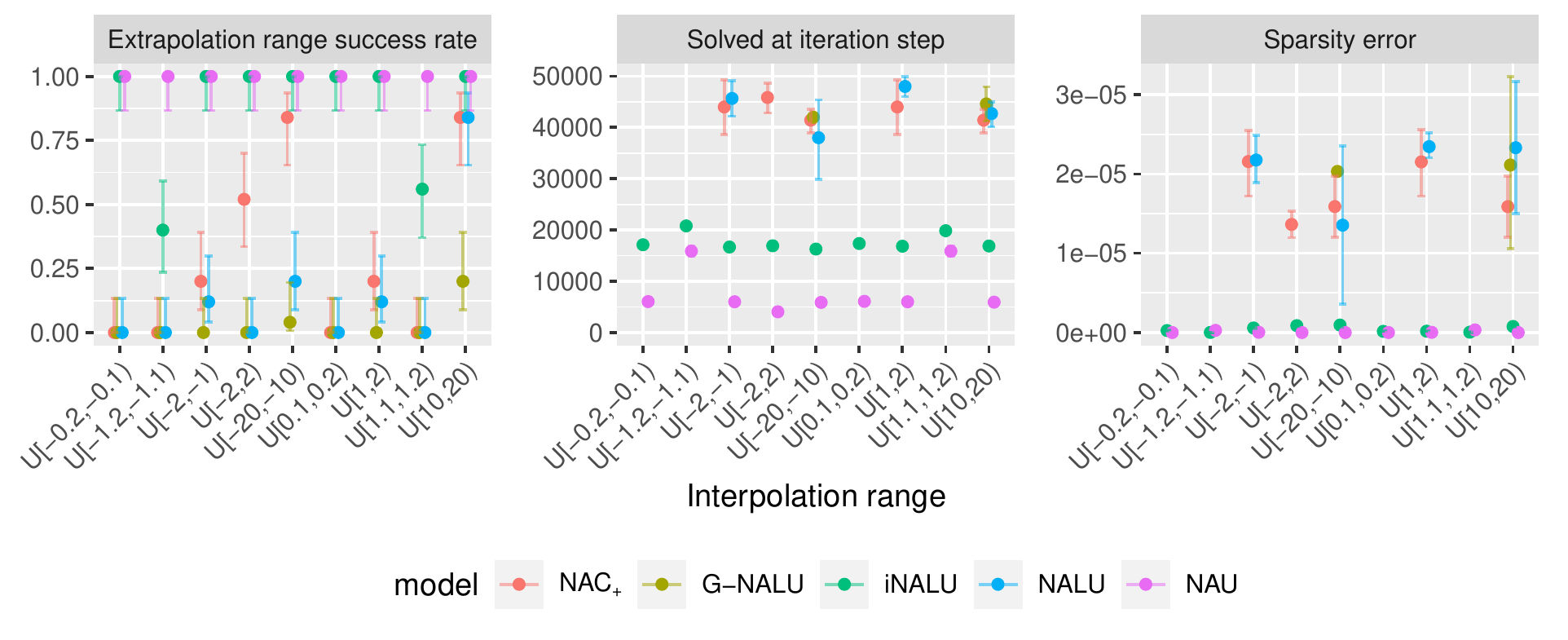}
\caption{Performance on Single Module Task for subtraction.}
\label{fig:slt-range-sub}
\vskip -0.2in
\end{figure*}

\textbf{Subtraction (Figure~\ref{fig:slt-range-sub}).}
The NAU has full success for all ranges. 
The solved at iterations does remain low, similar to addition, with perfect sparsity when converged. 
However, ranges $\mathcal{U}$[-1.2,-1.1) and $\mathcal{U}$[1.1,1.2) require over double the number of iterations to be solved compared to the rest of the ranges implying that small ranges with a mean of 1 can cause more challenging loss landscapes. 
The difficulty of these two ranges also holds for all other modules which have near 0 success (except the iNALU which has at least 40\% success). 
The iNALU has full success on all ranges excluding $\mathcal{U}$[-1.2,-1.1) and $\mathcal{U}$[1.1,1.2). 
Like addition, the solve speed and sparisty error of iNALU remain larger than the NAU. 
The NALU struggles much more with subtraction than with addition (except for $\mathcal{U}$[10,20)). 
The $\mathrm{NAC}_{+}$ outperforms NALU on 4 of the 9 ranges.  
The G-NALU does not outperform the NALU on any ranges. 

\begin{figure*}[t]
\vskip 0.2in
\centering
\includegraphics[width=\textwidth]{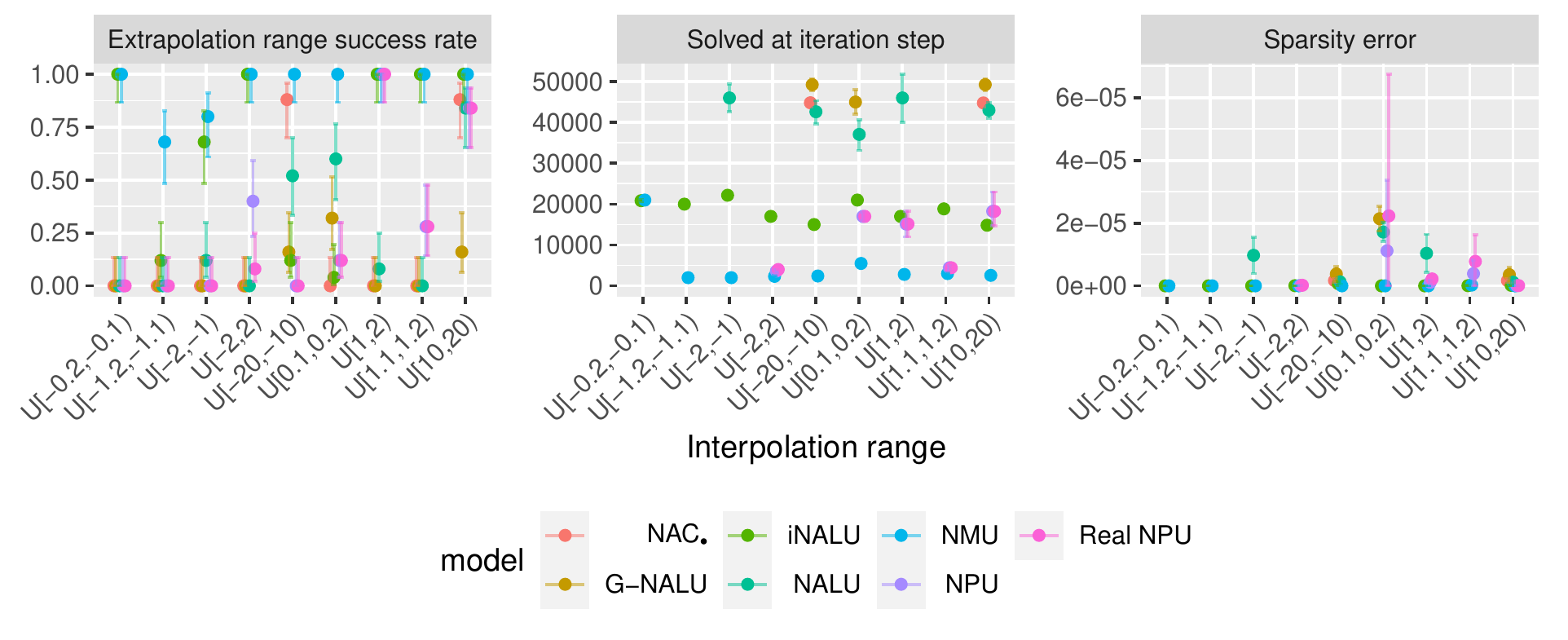}
\caption{Performance on Single Module Task for multiplication.}
\label{fig:slt-range-mul}
\vskip -0.2in
\end{figure*}

\textbf{Multiplication (Figure~\ref{fig:slt-range-mul}).} 
The NMU, iNALU, NPU and RealNPU have full success on range $\mathcal{U}$[1,2). 
The NMU struggles with some negative input ranges, i.e., $\mathcal{U}$[-1.2,-1.1) and $\mathcal{U}$[-2,~-1). 
Though NPUs in theory can learn with negative inputs, empirical results suggest the modules struggle. 
The NPU and Real NPU perform the same for all ranges except one, suggesting that the problem is not complex enough to require the use of the imaginary weight matrix. 
However, $\mathcal{U}$[-2,2) is an example in which $W_{im}$ is utilised (achieving 32\% more success than the RealNPU).  
Even though this range allows either of the input values to be positive or negative values, the learnt weights should be [1,1] for the real weights and [0,0] for the imaginary weights. 
The NALU can solve some ranges but no range with full success. 
The $\mathrm{NAC}_{\bullet}$ outperforms the NALU on the 2 ranges it has success but fails to achieve any success on the remaining 7 ranges. 
The iNALU outperforms the NALU on 7 ranges where it gains full success on 5 of those ranges.  
The G-NALU does not outperform the NALU on any ranges. 

\begin{figure*}[t]
\vskip 0.2in
\centering
\includegraphics[width=\textwidth]{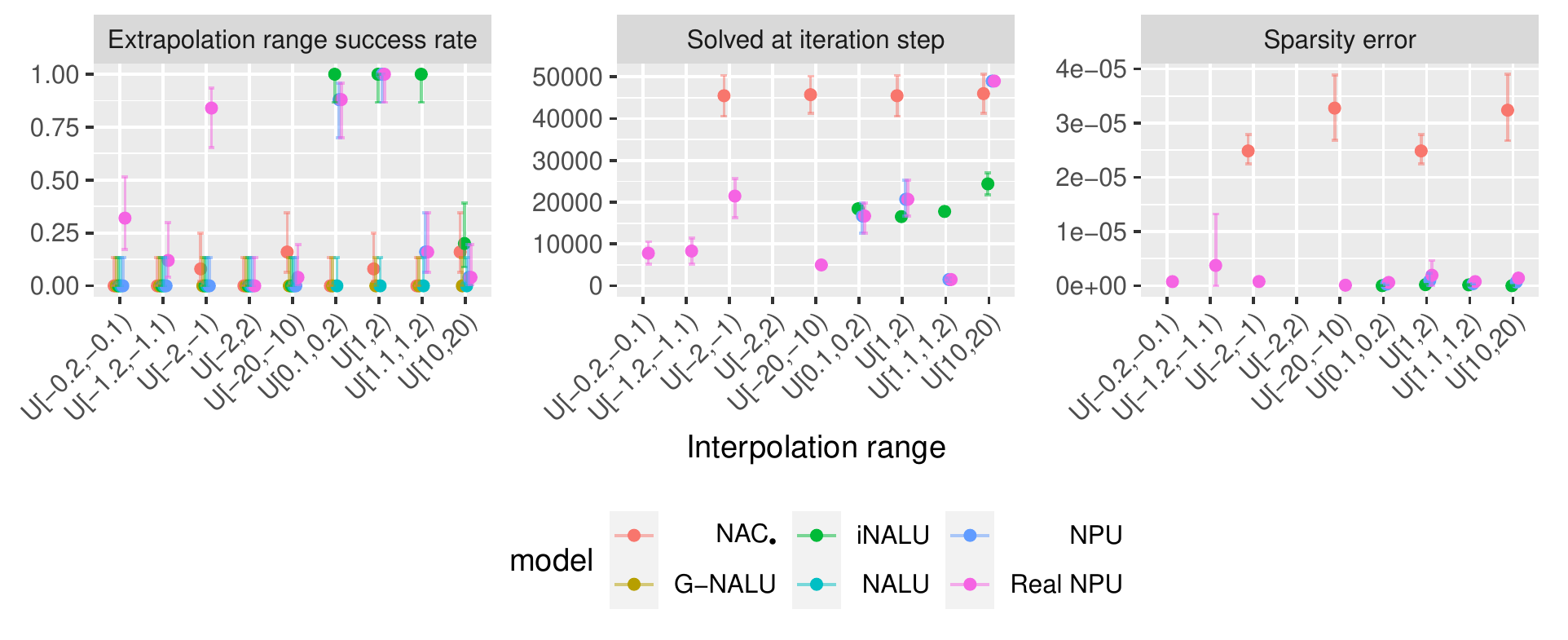}
\caption{Performance on Single Module Task for division.}
\label{fig:slt-range-div}
\vskip -0.2in
\end{figure*}

\textbf{Division (Figure~\ref{fig:slt-range-div}).}
No model solves division for all ranges. 
The iNALU is the only module to have a success rate of 1 on any range, fully solving 3 of the 9 ranges. 
This highlights the difficulty in modelling division even for the simplest case, aligning with prior claims~\cite{madsen2020neural}. 
The NPU and RealNPU performs perfectly for $\mathcal{U}$[1,2). 
The RealNPU has better performance over the NPU for negative input ranges. 
The $\mathrm{NAC}_{\bullet}$ is able to achieve some success on 4 ranges while the NALU and the G-NALU cannot achieve any success on all 9 ranges. 
The failure on $\mathcal{U}$[-2,2) for the NALU, iNALU and G-NALU is expected due to the inability to process mixed sign inputs caused by the modules' log-exponent transformation. 

\begin{figure*}[!t]
\vskip 0.2in
\centering
\includegraphics[width=\textwidth]{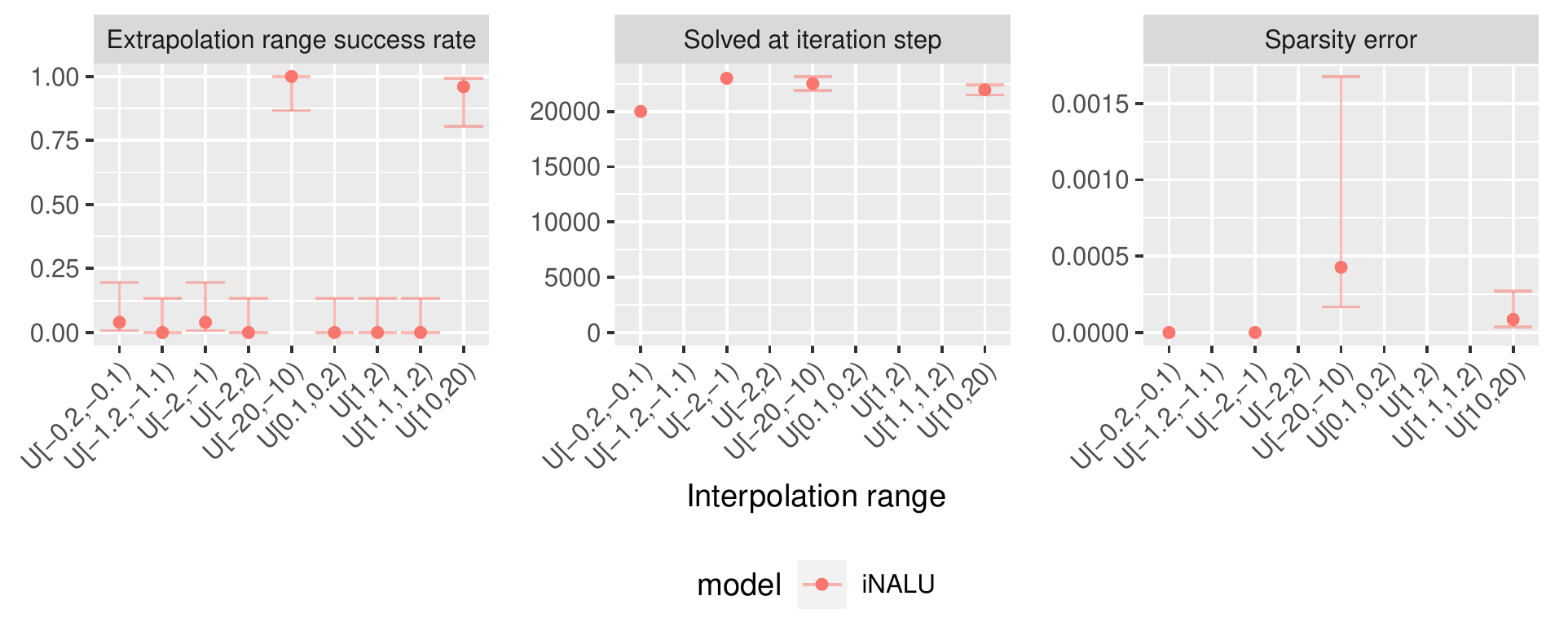}
\caption{iNALU failures on the Single Module Task for addition with 10 inputs (8 redundant inputs).}
\label{fig:in10-inalu-failures}
\vskip -0.2in
\end{figure*}

\begin{figure*}[!t]
\vskip 0.2in
\centering
\includegraphics[width=\textwidth]{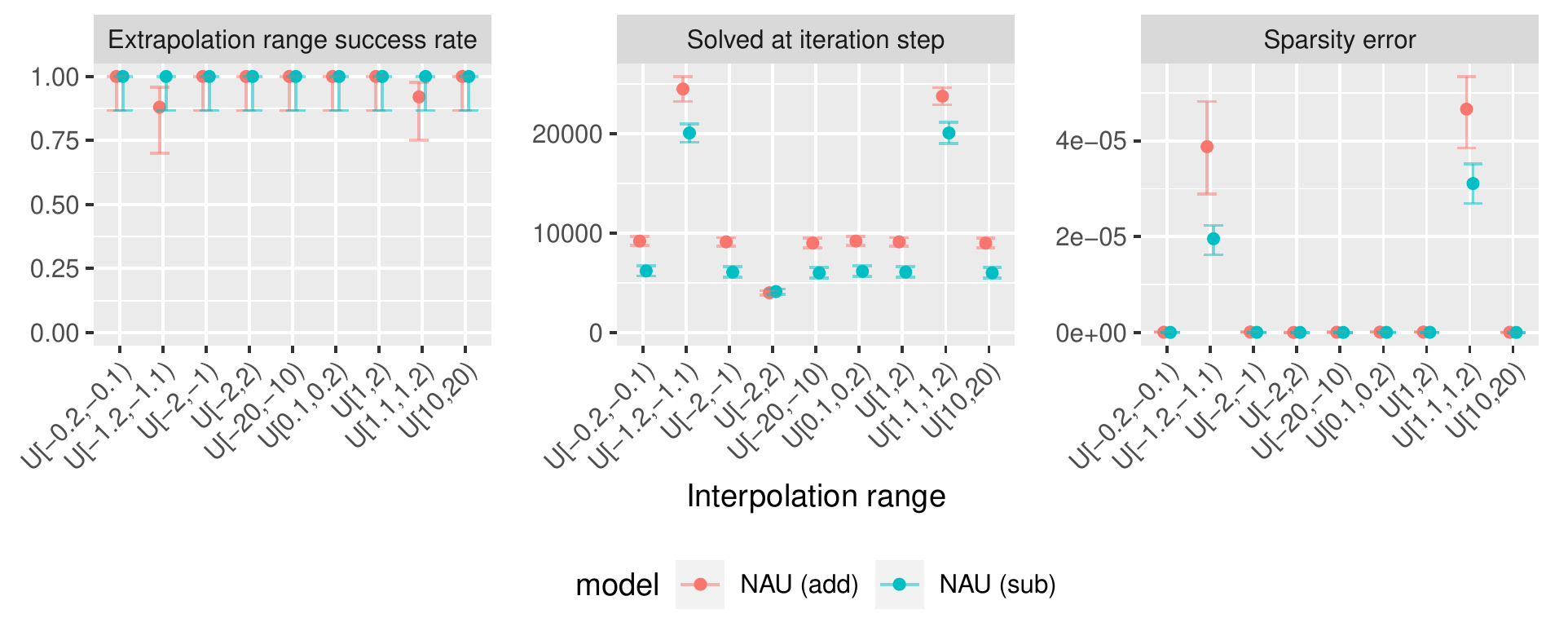}
\caption{NAU failures on the Single Module Task for addition with 10 inputs (8 redundant inputs).}
\label{fig:in10-nau-add-sub}
\vskip -0.2in
\end{figure*}

\begin{figure*}[!t]
\vskip 0.2in
\centering
\includegraphics[width=\textwidth]{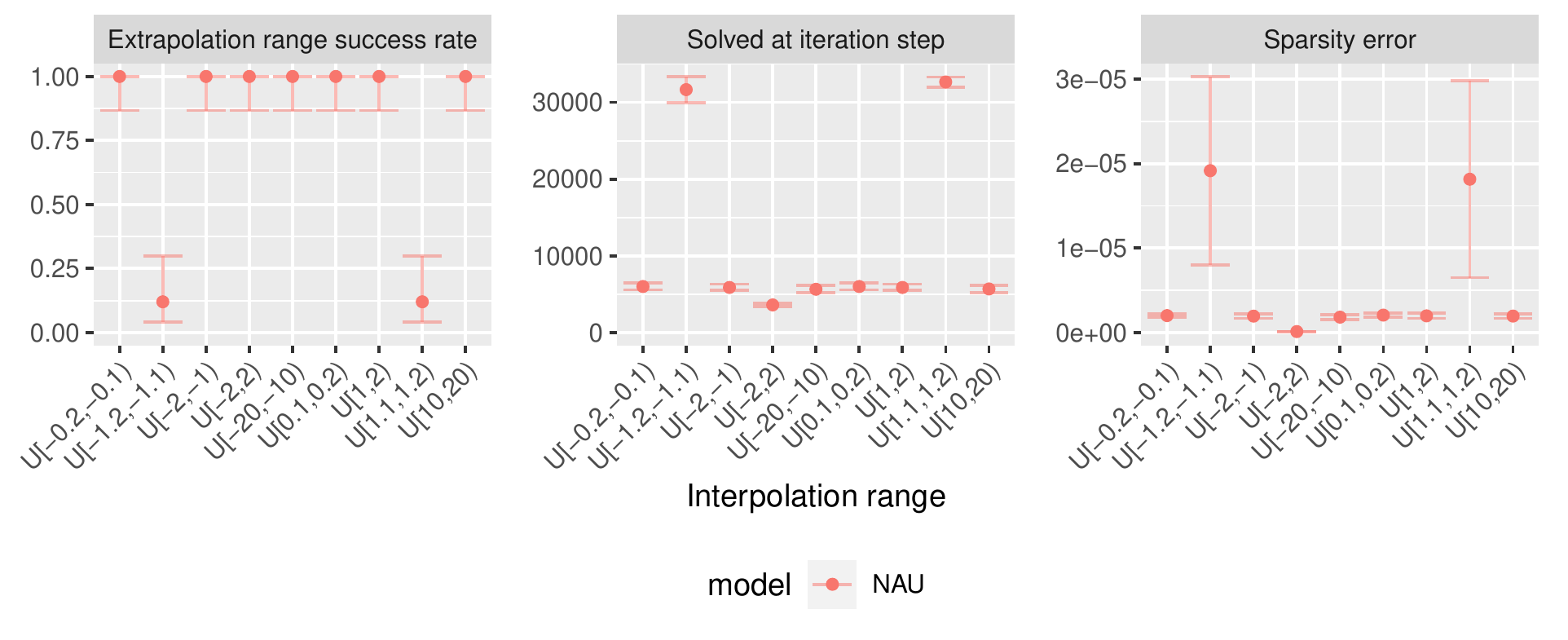}
\caption{NAU failures on the Single Module Task for subtraction with 100 inputs (98 redundant inputs).}
\label{fig:in100-nau-sub-failures}
\vskip -0.2in
\end{figure*}

\textbf{Testing limits of full success modules.} Achieving full success on all the ranges for a operation only occurred three times - the iNALU for addition and the NAU in addition and subtraction. To determine to what extent this holds we experiment with introducing redundant units to the input, resulting in an increase of the task difficulty. 
The iNALU (Figure~\ref{fig:in10-inalu-failures}) shows multiple failure ranges at 10 input units (8 redundant inputs), achieving reasonable success only on the larger ranges. 
The NAU fails at 10 inputs (8 redundant inputs) for both addition (Figure~\ref{fig:in10-nau-add-sub}) and at 100 inputs (98 redundant inputs) for subtraction  (Figure~\ref{fig:in100-nau-sub-failures}) on the same ranges U[-1.2,-.1.1) and U[1.1,1.2). 

\textbf{Summary.} 
The Single Module Task assesses the stability of individual NALMs. 
Having stability at this granularity is especially important, because if a NALM as a stand-alone unit is unstable how can we expect them to converge if applied larger networks where they are only a subcomponent? 
Overall, we find that a majority of NALMs are not robust to different training ranges. 
NALMs which achieved full success on all ranges for a operation on the two-input setup break when redundant inputs are introduced, however the extent varies depending on the module. 
Division is the most challenging operation followed by multiplication, subtraction and addition. 
NALMs which specialise in at most two operations are found to outperform the NALU in a majority of the cases. 
Of the NALMs which can model the four operations, i.e., the NALU, iNALU and G-NALU, the iNALU performs best on average over all operations, though the performance gain is less significant for multiplication and division.  

\section{Experiments and Findings of Modules for Logic Tasks}\label{sec:exp-and-findings-logic}
This section summarises the experiments provided in two existing logic based NALMs---the NLRL and the NSR. 

\subsection{NLRL} 
Preliminary results are given which tests basic logic and arithmetic operations: AND, OR, NOT and XOR, multiplication, addition with division, identity and constant selection. 
Each model consists of a stacked NLRL. 
Different numbers of intermediary units per layer are tested where the authors conclude increasing the units improves performance until a saturation point (found to be 8) is reached. 
A multi-operation based task requires stacking layers of NLRL which introduces redundancy as the stacked output and input layers both use negation gating. 
Therefore, if stacking is required, it is suggested to remove cases with consecutive negation gating layers and only have a single layer. 
Using a module with both AND and OR results in faster convergence (less iterations), compared to using a module only using AND operations, but has a longer computation time to train each iteration. 

\subsection{NSR}
\citet{faber2020neural} first check if the NSR can learn comparison operations on both an integer and floating point input setting. Results show that the NSR can learn the comparison functions with both input types and can extrapolate well. 
Modules struggle with learning the $=$ and $\neq$ operations, but performance can be improved by introducing redundancy through additional sets of weights during training \citep[Section~4.4]{faber2020neural}. 
The NSR can be attached with a NAU to learn piecewise functions. 
Findings suggest a simple continuous function (such as the absolute difference between two inputs) can be learnt with extrapolation capabilities, but a non-continuous function cannot. 
The NSR can be converted into a recurrent module to find the minimum of a list and to count the occurrence of a number in a sequence. The minimum task performs perfectly on all extrapolation settings however the counting task's performance reduces as sequence length increases. 
An additional task requires finding the shortest paths in a Graph Neural Network (GNN) where the network should learn to imitate the Bellman-Ford algorithm. 
The is used to show that the recurrent NSR can learn to aggregate numbers to a minimum. 
When extrapolating to larger graphs, performance improved with larger edge weights.  
Finally, a MNIST digit comparison task tested to see if the NSR can be used as a downstream module for a CNN in an end-to-end manner. 
Findings show that the NSR based network cannot outperform a vanilla CNN but is comparable to a MLP based network, where the underperformance was suggested to be a result of a weak learning signal. 

\section{Applications of NALU}\label{sec:applications}
This section describes uses of NALU as a sub-component in architectures to tackle practical problems outside the domain of solving arithmetic on numeric inputs. Success and failure cases are mentioned. 
We choose to focus on NALU applications on the basis that the improved modules discussed above can be applied in place of NALU to provide additional performance gains to the mentioned applications.

\subsection{Existing Applications}
\citet{xiao2020trajdata} insert a NALU layer between a two-layer Gated Recurrent Unit (GRU) and dense layer to predict vehicle trajectory of complex road sections (containing constantly changing directions). NALU improves extrapolation capabilities to deal with abnormal input cases outside the range of the GRU hidden states output. 

\citet{raj2020fast} combine $\mathrm{NAC}_{+}$ modules before LSTM cells for \textit{fast} training in the extraction of temporal features to classify videos for badminton strokes. They further experiment in using $\mathrm{NAC}_{+}$ modules with a dense layer to learn temporal transformations, finding better performance than the LSTM based module and the dense modules being quicker to train. They justify the use of the $\mathrm{NAC}_{+}$ as a way to produce sparse representations of frames, as non-relevant pixels would not be selected by the $\mathrm{NAC}_{+}$ resulting in 0 values, while relevant pixels accumulate.

\citet{Zhang2019EnhancingTC} use deep reinforcement learning to learn to schedule views on content-delivery-networks (CDNs) for crowdsourced-live-streaming (CLS). NALU's extrapolative ability alleviates the issue of data bias (which is the failure of models outside the training range) by using the NALU to build an offline simulator to train the agent when learning to choose actions. The simulator is composed of a two layer LSTM with a NALU layer attached to the end. 
\citet{Zhang2019LivesmartAQ} propose a novel framework (named Livesmart) for cost-efficient CLS scheduling on CDNs with a quality-of-service (QoS) guarantee. Two components required in Livesmart contain models using NALU. 
The first component (named new viewer predictor) uses a stacked LSTM-NALU to predict workloads from new viewers. 
The second component (named QoS characterizer) predicts the QoS of a CDN provider. This component uses a stack of Convolutional Neural Networks (CNNs), LSTM and NALU. 
Both components use the NALU's ability to capture OOD data to aid in dealing with rare events/unexpected data.

\citet{wu2020analogical} combines layers of $\mathrm{NAC}_{+}$ to learn to do addition and subtraction on vector embeddings to form novel compositions for creating analogies. Modules are applied to the output of an attention module (scoring candidate analogies) that is passed through a MLP. The output of the $\mathrm{NAC}_{+}$ modules is passed to a LSTM producing the final analogy encoding. 

The NALU has also been used with CNNs. 
\citet{rajaa2019convolutional} applies stacked NALUs to the end of convolution units to predict stock future stock prices. \citet{rana2020systematically} utilises the $\mathrm{NAC}_{+}$/NALU as residual connections modules to larger convolutional networks such as U-Net and a fully convolutional regression networks for cell counting in images. 
Such connections enable better generalisation when transitioning to data with higher cell counts to the training data. 
However, no observations are made to what the units learn which lead to an improvement on cell counting over the baseline models.

\citet{chennupati2020machine} uses the NALU as part of a larger architecture to predict the runtime of code on different hardware devices configured using hyperparameters. The NALU predicts the reuse profile of the program, keeping track of the count of memory references accessed in the execution trace. The NALU outperforms a genetic programming approach for doing such a prediction.

\citet{Teitelman2020StealingBF} explores the problem domain of cloning black-box functionality in a generalisable and interpretable way. A decision tree is trained to differentiate between different tasks of the black box. Each leaf of the tree is assigned a neural network comprising of stacked dense layers with a NALU layer between them. Each neural network is able to learn the black-box behaviour for a particular task. Like \citet{xiao2020trajdata}, results showed that NALU is required to learn the more complex tasks. 

Finally, \citet{sestili2018towards} suggests the NALU has potential use in networks which predict security defects in code. This is due to the module's ability to work with numerical inputs in a generalisable manner, instead of limiting the application to be bound to a fixed token vocabulary requiring lookups. 

\subsection{Applications Where NALU Is Inferior}
We discuss examples of situations in which NALU modules are a sub-optimal architecture choice for applicational settings. 
\citet{madsen2020neural} show that the NAU/NMU outperforms NALU in the MNIST sequence task for both addition and multiplication. 
\citet{dai2020abductive} show the arithmetic ability (named background knowledge) of the NALU is incapable in performing the MNIST task for addition or products when combined with a LSTM. 
Instead, they show a neural model for symbolic learning, which learns logic programs using pre-defined rules as background knowledge, can perform with over 95\% accuracy. 
However, we question whether the failure is a result of the NALU or due to the misuse of its abilities from combining it with a LSTM. For example, as the inputs are images, unless the LSTM converts each image into a numerical value which can be processed by the NALU in an arithmetic way it can be suggested that the LSTM is completing the task without the numerical capabilities of the NALM. 
\citet{jacovi2018neural} show that in black box cloning for the \citet{trask2018neural} MNIST addition task, their EstiNet model which captures non-differentiable models outperforms NALU. 
Though it can be argued that a more relevant comparison would test the $\mathrm{NAC}_{+}$ or the NAU which are solely designed for addition.
\citet{joseph2019momen} show that although the $\mathrm{NAC}_{\bullet}$ can learn the order for a polynomial transformation to a high accuracy, it is still outperformed by a pre-defined order two polynomial model. Results suggest that the $\mathrm{NAC}_{\bullet}$ may not have fully converged to express integer orders.
\citet{dobbels2020predicting} found the NALU was unable to extrapolate for the task of predicting far-infrared radiation fluxes from ultraviolet-mid-infrared fluxes. Though no clear reason was stated, the lack of extrapolation could be attributed to the co-dependence of features because of applying a fully connected layers prior to the module. 
\citet{jiancpu} considers the NALU as a hardware component concluding that the NALU has too high an area and power cost to be feasible for practical use. Implementing for addition costs 17 times the area of a digital adder, and the memory requirements for weight storage is energy inefficient for doing CPU operations. 

\section{Remaining Gaps}\label{sec:remaining-gaps}
This section discusses areas which remain to be fully addressed. We focus on: \textit{benchmarks, division, robustness, compositionality}, and \textit{interpretability  of more complex architectures.}

Having \textbf{benchmarks} is important in allowing for reliable comparison between modules. Such benchmarks should include a simple synthetic dataset which we detail in this paper, and a real-world data benchmark (which remains to be created) with a systematic evaluation. 

\textbf{Division} remains a challenge. To date no module has been able to reliably solve division. Currently the NPU by \citet{heim2020neural} is the best module to use, though it would struggle with input values close to zero. \citet{madsen2020neural} argues modelling division is not possible due to the singularity issue. 
One suggestion for dealing with the zero case is to take influence from \citet{reimann2019neural} which can have an option for showing an output which is invalid (or in their case all off values).

One goal of these modules is to be able to extrapolate. To achieve this, a module should be \textbf{robust} to being trained on any input range. \citet{madsen2020neural} show that modules are unable to achieve full success of all tested ranges (with the stacked NAU-NMU failing on a training range of [1.1,1.2], being unable to obtain a single success). Reinitialisation of weights~\citep{schlor2020inalu} during training could provide a solution, however this seems to be unlikely given \citet{madsen2020neural} tests against 100 model initialisations and using reinitalisation for a NALM that is part of a large end-to-end network may not be economical.

\textbf{Compositionality} is desirable. A model should be flexible, having the option to select different types of operations and model complex mathematical expressions. Currently the two popular approaches are gating and stacking. Gating has been found to not work as expected and give convergence issues. Stacking, though more reliable, has less options in operation selection than gating. Deep stacking of modules (in a non-recurrent fashion) remains untested.

It remains to be understood \textbf{how modules influence learning of other modules} (such as recurrent networks and CNNs) in their representations. For example, seeing if representations are more interpretable because of being trained with a module.

\section{Related Work}\label{sec:rel-work}
We now take a step back, overviewing related works, as part of the bigger picture in deep learning. In particular we focus on other arithmetic based architectures, inductive biases and specialist modules.

\subsection{Extrapolative Mathematics}
In contrast to specialists, generic neural architectures have also been investigated for learning mathematics. 
Two such examples include convolutional recurrent networks (used in the Neural GPU)~\cite{Kaiser2016} and transformers~\cite{lample2019deep}, of which both have been shown to be Turing complete.\footnote{That is, Turning complete under the assumption that arbitrary (rather than finite) precision is used.}~\citep{perez2018on}
Neural GPUs are constructed from convolutional gated recurrent units. The Neural GPUs can extrapolate to long sequence lengths (2000) from being trained on length 20 inputs, but use binary inputs rather than real numbers~\citep{Kaiser2016}. 
However, various training techniques require to be implemented such as curriculum learning, relaxed parameter sharing and dropout; such techniques are not required for training NALMs. 
Furthermore only a few Neural GPU models generalise to such a long sequence, but this has been improved on in \citet{freivalds2017improving} by simplifying the architecture/training and introducing diagonal gating and hard non-linearities with additional cost functions. 
Transformers, which can process numerical values, remain unsuccessful for extrapolation tasks which are simple such as arithmetic using multiplication~\citep{saxton2018analysing}. 
In contrast, once a NALM learns to apply an operation (by converging to the relevant interpretable weight) it will always calculate the operation correctly. 
For more complex maths such as integration, generalisation over different input/output sequence length data generators have also been identified as a weak point in transformers~\citep{lample2019deep}. 

Other approaches which can process raw numerical inputs include using reinforcement learning,  non-specialised MLP architectures and symbolic regression. 
\citet{chen2018neural} uses a multi-level hierarchical reinforcement learning approach allowing for operations to be decomposed into simpler operations and solved via specialised skill modules. A proximal policy optimisation strategy is used to train the modules responsible for decomposing and calling the specialised skill modules. Furthermore, a curriculum learning methodology is adopted by using a teacher-student continual learning strategy to control the task difficulty setting when learning.
In experiments for modelling the four arithmetic operations, the model is able to learn to full accuracy short sequence lengths (5) but cannot learn longer lengths (20). 
Similar to NALMs, the model especially struggles with division. 
A downside is that arithmetic operation/s require to be defined in the input unlike in NALMs where only the input values require to be given. 
~\citet{nollet2020learning} uses non-specialised MLP architectures to learn long multiplication and addition for up to 7 digits. By breaking the task into processing steps representing sub-operations allows for the input to act as external memory. Similar to~\citet{Kaiser2016}'s need of curriculum learning, active learning was required to control the difficulty of the dataset to learn long multi-digit multiplication. Though less interpretable than NALU weights, certain neurons in the MLP were found to encode digit operations for some operands. However, extrapolation performance to longer digits remained untested. 

In short, though various alternates to NALMs exist, each have their own shortcomings in regard to input format, extrapolation, and robustness.

\subsection{Inductive Biases}
Using an inductive bias, to give control over the learning space of the model, can be a critical factor in achieving generalisation \citep{Mitchell1980}. 
In NALMs, weights have inductive biases such that particular weights (e.g., discretised values) represent applying an arithmetic/logic operation. 
Others forms include utilising knowledge of the task to incorporate biases directly into the architecture. For example, using periodic activation functions (sin and cos) \citep{martius2017extrapolation} or simplifying expressions through symmetry/separability \citep{udrescu2020feynmanai} to model physics based expressions. 
Alternatively, regularisation can be used as a form of bias, incorporated with additional auxiliary loss terms \citep{relex2020lopedoto}. 
NALMs can use such losses to induce discretisation~\citep{schlor2020inalu, madsen2020neural}. 
Though auxiliary losses can only be minimised at training time and result in optimising an alternate objective to the original loss, a possible way to alleviate this through the use of unsupervised fine tuning and nested optimisation~\citep{alet2021tailoring}.

\subsection{Specialist Modules}
NALMs can be viewed as specialist modules for arithmetic/logic.
Using modules can provide better systematic generalisation than generic architectures \citep{bahdanau2018systematic}. 
Other works on specialist modules include \citet{zhang2018learningPermute} who introduces a module which learns permutation-invariant representation of a set. This is achieved by learning a pairwise ordering cost function with constraints to acquire desirable properties such as symmetry and the identity value. Importantly, this module does not require priori knowledge of the inputs which is analogous to how NALMs do not require to know the operation to learn. 
\citet{zhang2018learningCount} creates a module to learn to count objects in images from attention weights with the ability to reduce double counting of visual objects. The module takes in object proposals and converts them into a graph whose edges can be removed/scaled in a fully differentiable manner to recover the object count. In particular, they learn piecewise linear monotonically increasing functions, designed to handle overlapping object proposals while enforcing the constraint that the extreme cases of bounding boxes being either fully distinct/overlapping returning 0 and 1 respectively and having all other cases interpolate between the [0,1]. 
This is analogous to how many NALMs use the bounding parameter ranges to represent particular operations such as -1 for addition and 1 for subtraction, and interpolate between them when learning. Furthermore, this counting module is an example of a specialist which can be integrated into larger networks to improve performance for more complex tasks \citep{kim2018bilinear}. 

Utilising different specialists encourages factorisation of the task resulting in searching for reusable components. 
\citet{hu2017learning} produce specialist modules for compositional reasoning tasks in visual question answering. 
For example, the `find' and `relocate' modules output attention maps which can be applied to the visual input in order to carry out the sub-task. `Find' would be able to focus on an object/attribute (e.g., the red sphere) and `relocate' can infer spatial relations (e.g., focus on object A \textit{behind} object B). 
\citet{Jiang2019SelfAssemblingMN} use similar types of modules with attention maps as outputs but only use a text based modality. 
Modules with generic structures such as LSTMs can also be turned into specialist modules by controlling the update dynamics through inter-module competition and sparse communication \citep{goyal2021recurrent, goyal2020inductive}. 
However, due to the generic nature, such modules will not be interpretable at a parameter level like NALMs. 
\section{Conclusion}
Neural Arithmetic Logic Modules (NALMs) are a promising area of research for systematic generalisation. 
Focusing on the first Neural Arithmetic Unit, the NALU, we explained the unit's limitations along with existing solutions from other modules: iNALU, NAU, NMU, NPU, and CalcNet. 
We also detail the two logic NALMs: NLRL and NSR inspired by NALU. 
There exists a range of applications for the NALU, though some uses remain questionable. 
Cross-comparing modules suggest inconsistencies with experiment methodology and limitations existing in the current state-of-the-art modules. 
A new benchmark is provided for comparing arithmetic modules named the `Single Module Arithmetic Task'. 
Finally, we outline remaining research gaps regarding: solving division, robustness, compositionality and interpretability of complex architectures. 

\acks{We would like to thank Andreas Madsen for informative discussions and explanations regarding the Neural Arithmetic Units, and the anonymous reviewers who have help improve the manuscript. B.M. is supported by the EPSRC Doctoral Training Partnership (EP/R513325/1). J.H. received funding from the EPSRC Centre for Spatial Computational Learning (EP/S030069/1). The authors acknowledge the use of the IRIDIS High-Performance Computing Facility, the ECS Alpha Cluster, and associated support services at the University of Southampton in the completion of this work.}

\newpage
\appendix

\section{Architecture Illustration Key}\label{app:arch-key}
\begin{minipage}[c]{\textwidth}
\vskip -0.2in
\includegraphics[width=\textwidth]{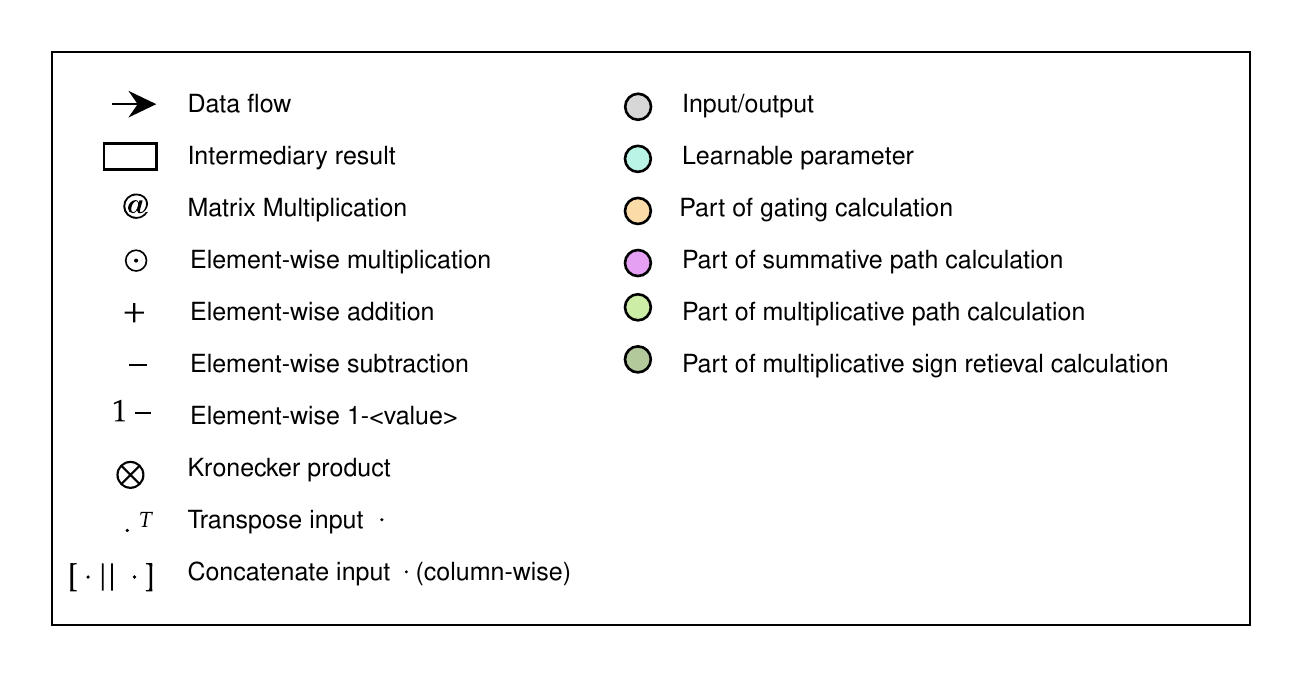}
\vskip -0.2in
\captionof{figure}{Key containing the symbols and colouring system used for architecture illustrations.}
\label{fig:arch-key}
\end{minipage}

\section{Module Illustrations}\label{app:unit-imgs}
Table~\ref{tab:unit-imgs} displays, in chronological order, the module architecture illustrations given in their respective papers. $^*$Note that we  modified the NALU architecture from~\protect\citet[Figure~2b]{trask2018neural} as the learned gate matrix ($\mathbb{R}^{3\times4}$) is mistakenly drawn as a vector ($\mathbb{R}^3$) in the original figure.)  
\begin{longtable}{p{4cm}p{10.5cm}}
    \toprule
    \textbf{Module} & \textbf{Architecture} \\\midrule
    \textbf{NALU$^*$}~\citep{trask2018neural} & \parbox[c]{1em}{\includegraphics[width=4in]{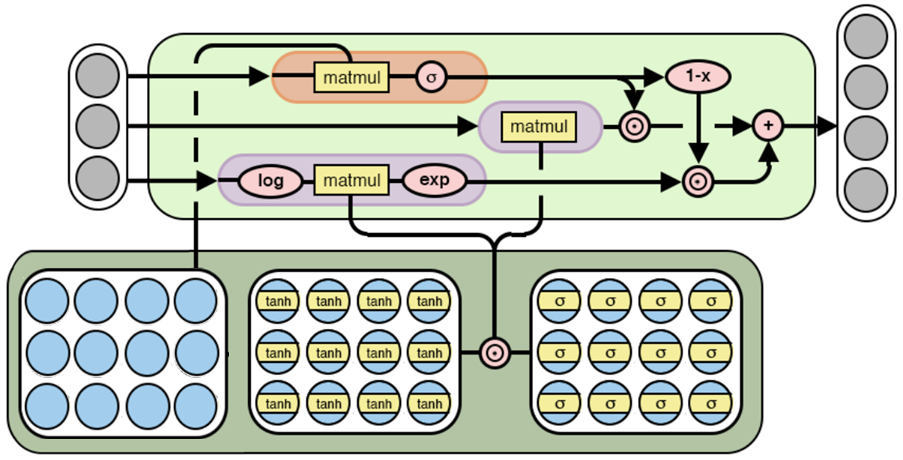}}\\\hline
    \textbf{NLRL}~\citep{reimann2019neural} & \parbox[c]{1em}{\includegraphics[width=4in]{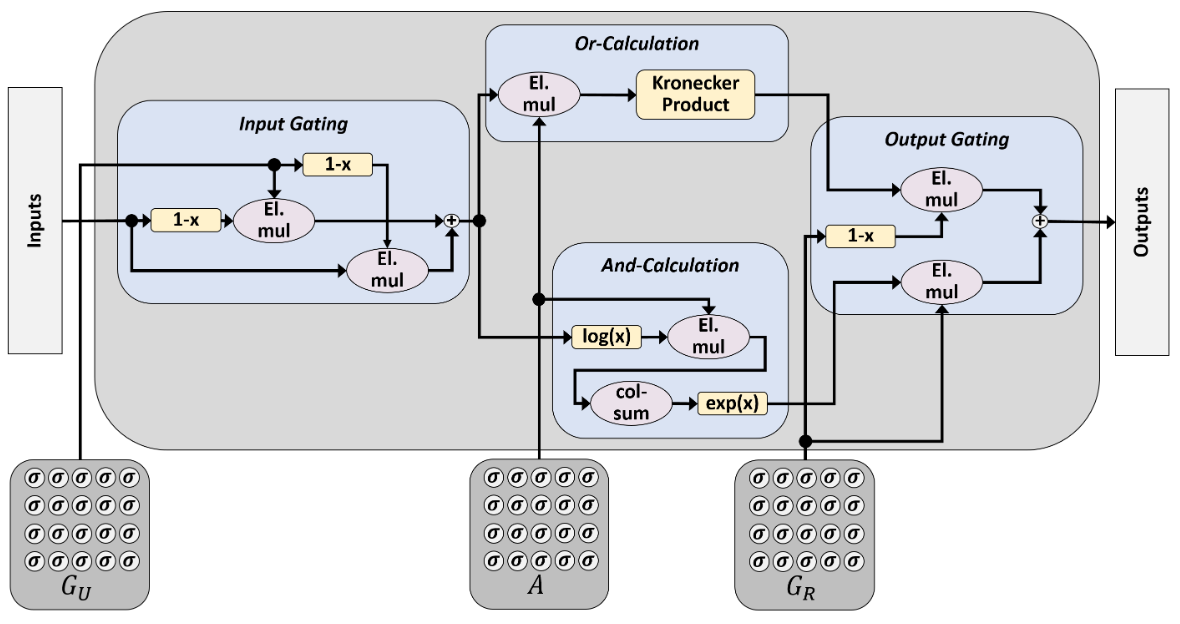}}\\\hline
    \textbf{G-NALU}~\citep{rajaa2019convolutional} & (No figure exists)\\\hline
    \textbf{NAU}~\citep{madsen2020neural} & (No figure exists)\\\hline
    \textbf{NMU}~\citep{madsen2020neural} & \parbox[c]{1em}{\includegraphics[width=4in]{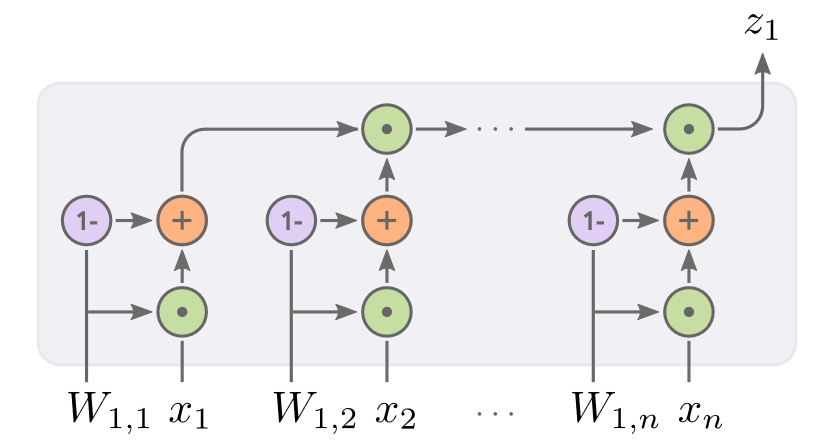}}\\\hline
    \textbf{NSR}~\citep{faber2020neural}& \parbox[c]{1em}{\includegraphics[width=4in]{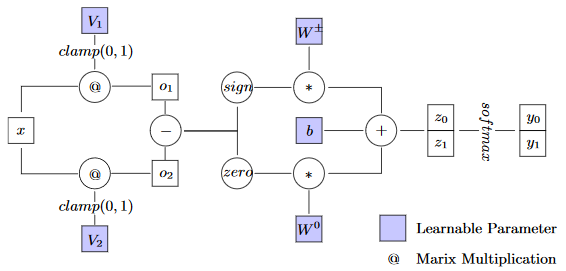}}\\\hline
    \textbf{iNALU}~\citep{schlor2020inalu}& \parbox[c]{1em}{\includegraphics[width=4in]{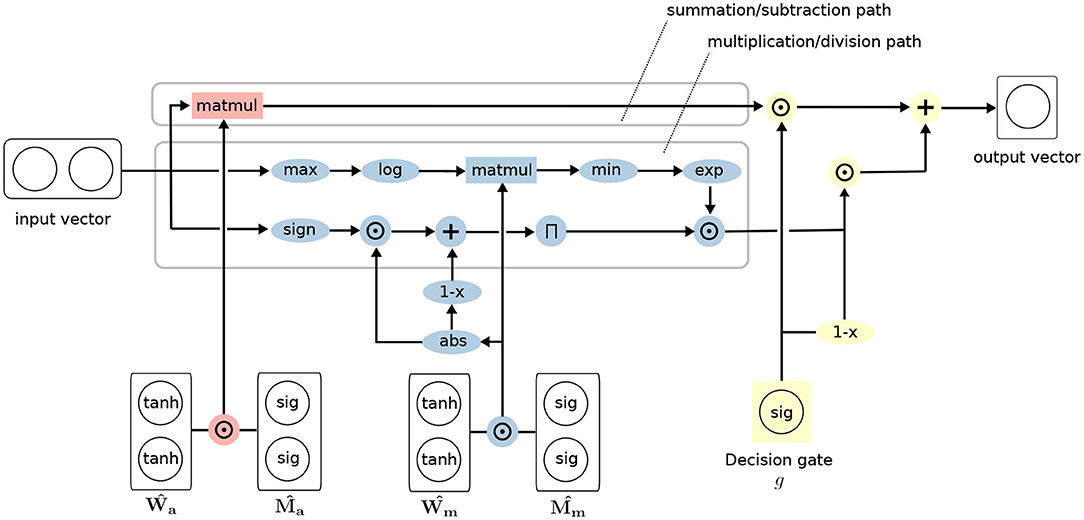}}\\\hline
    \textbf{NPU}~\citep{heim2020neural}& \parbox[c]{1em}{\includegraphics[width=4in]{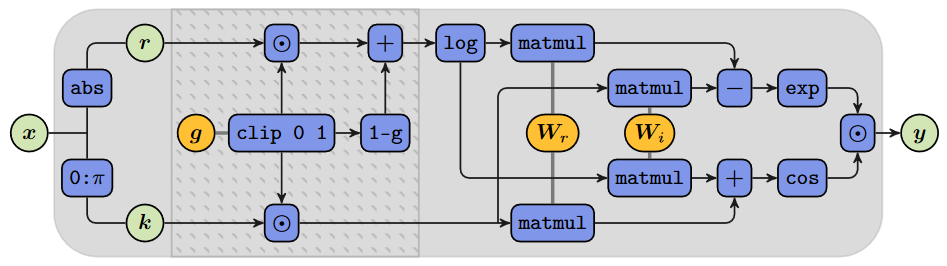}}\\\bottomrule
\caption{Module architecture illustrations taken from the original papers.}
\label{tab:unit-imgs}
\end{longtable}

\newpage
\section{Step-by-step Example using the NALU}\label{app:step-by-step-nalu}
To better understand how the internal process of a NALM, we provide a worked through example for addition using the NALU with parameters that can extrapolate. 

\textbf{Task:}
\textit{Subtract the the second input value from the first where the input is $\bm{x}= \left[\protect\begin{smallmatrix} 2 & 3 & 4 \protect\end{smallmatrix}\right]$. The output value should be $\left[\protect\begin{smallmatrix} -1 \protect\end{smallmatrix}\right]$.}

\textbf{Steps:}
\begin{enumerate}
    \item Calculate $\tanh(\bm{\widehat{W}})$. The operation is $+x_1 - x_2$ so
    $\tanh(\bm{\widehat{W}}) = \left[\protect\begin{smallmatrix}
        1 \\
        -1 \\
        0   
    \protect\end{smallmatrix}\right]$.
    \item  Calculate $\sigmoid(\bm{\widehat{M}})$. The first two input values are selected and the third is ignored, so 
    $\sigmoid(\bm{\widehat{M}}) = \left[\protect\begin{smallmatrix}
        1 \\
        1 \\
        0   
    \protect\end{smallmatrix}\right]$.
    
    \item Calculate $\tanh(\bm{\widehat{W}}) \odot \sigmoid(\bm{\widehat{M}})$ to obtain $\bm{W} = \left[\protect\begin{smallmatrix}
        1 \\
        -1 \\
        0   
    \protect\end{smallmatrix}\right]$. 
    \item Calculate the result of the summative path. 
    \begin{align*}
        \textrm{NAC}_+ &= \bm{xW} \\&= \left[\protect\begin{smallmatrix} 2 & 3 & 4 \protect\end{smallmatrix}\right] 
    \left[\protect\begin{smallmatrix}
        1 \\
        -1 \\
        0   
    \protect\end{smallmatrix}\right] 
    \\&= 
    \left[\protect\begin{smallmatrix} (2\times1) + (3\times-1) + (4\times0) \protect\end{smallmatrix}\right]
    \\&= \left[\protect\begin{smallmatrix} 2-3+0 \protect\end{smallmatrix}\right]
    \\&= \left[\protect\begin{smallmatrix} -1 \protect\end{smallmatrix}\right].
    \end{align*}
    
    \item Calculate the result of the multiplicative path. (For simplicity, let us assume $\epsilon = 0$.)
    \begin{align*}
    \textrm{NAC}_\bullet &= \exp(\bm{W}\ln(|\bm{x}| + \epsilon)
    \\&= 
    \exp(
    \left[\protect\begin{smallmatrix}1\\-1 \\0\protect\end{smallmatrix}\right]
    \ln(|\left[\protect\begin{smallmatrix} 2 & 3 & 4 \protect\end{smallmatrix}\right]|)
    \\&=\exp(
    \left[\protect\begin{smallmatrix}1\\-1 \\0\protect\end{smallmatrix}\right]
    \left[\protect\begin{smallmatrix} \ln(2) & \ln(3) & \ln(4) \protect\end{smallmatrix}\right])
    \\&=\exp(
    \left[\protect\begin{smallmatrix} \ln(2^1) + \ln(3^{-1}) + \ln(4^0) \protect\end{smallmatrix}\right])
    \\&= 
    \exp(\ln([2^1 \times 3^{-1} \times 4^0])
    \\&= 
    \exp(\ln(\left[\protect\begin{smallmatrix}2 \times \frac{1}{3} \times 1\protect\end{smallmatrix}\right])) 
    \\&= \exp(\ln(\left[\protect\begin{smallmatrix}\frac{2}{3}\protect\end{smallmatrix}\right])) 
    \\&= \left[\protect\begin{smallmatrix} \frac{2}{3} \protect\end{smallmatrix}\right].
    \end{align*}
    \item The target expression requires the summative path ($\textrm{NAC}_+$) and ignores the multiplicative path ($\textrm{NAC}_\bullet$), therefore gate is 
    $\sigmoid(\bm{x}\bm{G}) = \left[\protect\begin{smallmatrix} 1 \protect\end{smallmatrix}\right] $. 
    
    \item Combine all the pieces to get the output. \\
    \begin{align*}
        \bm{\hat{y}} &= \bm{g} \odot \bm{a} + (\bm{1} - \bm{g}) \odot \bm{m}
        \\&= 
        \left[\protect\begin{smallmatrix} 1 \protect\end{smallmatrix}\right] \odot \left[\protect\begin{smallmatrix} -1 \protect\end{smallmatrix}\right] + (\left[\protect\begin{smallmatrix} 1 \protect\end{smallmatrix}\right] - \left[\protect\begin{smallmatrix} 1 \protect\end{smallmatrix}\right]) \odot \left[\protect\begin{smallmatrix}\frac{2}{3}\protect\end{smallmatrix}\right]
        \\&= \left[\protect\begin{smallmatrix} -1 \protect\end{smallmatrix}\right] + \left[\protect\begin{smallmatrix} 0 \protect\end{smallmatrix}\right] 
        \\&= \left[\protect\begin{smallmatrix} -1 \protect\end{smallmatrix}\right].
    \end{align*}
\end{enumerate}
\newpage
\section{Single Module Task Module Parameters}\label{app:parameters}
Refer to Tables~\ref{tab:exp-params}, \ref{tab:sltr-npu-params}, and \ref{tab:additional-params} for the breakdown of parameters used in the Single Module Task for experiments with input size 2.
\begin{table}[t!]
\vskip 0.1in
\begin{center}
\begin{tabular}{p{3cm}l}
\toprule
\textbf{Parameter}          & \textbf{Single Module Task}     \\ \midrule
\textbf{Layers}             & 1                              \\ 
\textbf{Input size}         & 2                             \\ 
\textbf{Subset ratio}       & 0.5                            \\ 
\textbf{Overlap ratio}      & 0                              \\ 
\textbf{Total iterations}  & 50000                          \\ 
\textbf{Train samples}      & 128 per batch                  \\ 
\textbf{Validation samples$^*$} & 10000                          \\ 
\textbf{Test samples$^*$}       & 10000                          \\ 
\textbf{Seeds}              & 25                             \\ 
\textbf{Optimiser}          & Adam (with default parameters) \\ 
\textbf{Learning rate$^{\dagger}$}      & 1.00E-03                   \\ \bottomrule
\end{tabular}
\end{center}
\caption{Parameters which are applied to all modules. $^*$Validation and test datasets generate one batch of samples at the start which gets used for evaluation for all iterations. $^\dagger$Will hold unless specified otherwise.}
\label{tab:exp-params}
\vskip -0.1in
\end{table}

\begin{table}[h!]
\centering
\vskip 0.1in
\begin{tabular}{p{1.5cm}lll}
\toprule
\textbf{Module}                          & \textbf{Operation}                 & \textbf{Parameter}              & \textbf{Value} \\ \midrule
\multirow{5}{*}{\parbox{1.5cm}{\textbf{NPU, RealNPU}}} 
    & \textbf{Mul}                       & ($\beta_{start}$,$\beta_{end}$) & (1e-7,1e-5)    \\ \cline{2-4} 
      & \textbf{Div}                       & ($\beta_{start}$,$\beta_{end}$) & (1e-9,1e-7)    \\ \cline{2-4} 
      & \multirow{3}{*}{\textbf{Mul, Div}} & $\beta_{growth}$                & 10             \\ 
      &                                    & $\beta_{step}$                  & 10000          \\ 
      &                                    & Learning rate                   & 5.00E-03       \\ \bottomrule
\end{tabular}
\caption{Parameters specific to the NPU and RealNPU modules for the Single Module Task.}
\label{tab:sltr-npu-params}
\vskip -0.1in
\end{table}

\begin{table}[h!]
\vskip 0.1in
\begin{center}
\begin{tabular}{ll}
\toprule
\textbf{Parameter}  & \textbf{Single Module Task} \\ \midrule
$\hat{\lambda}$     & 0.01 (NAU), 10 (NMU)      \\ 
$\lambda_{start}$   & 20000                      \\ 
$\lambda_{end}$     & 35000                     \\ \bottomrule              
\end{tabular}
\end{center}
\caption{Parameters specific to the NAU and NMU modules for the Single Module Task.}
\label{tab:additional-params}
\vskip -0.1in
\end{table}

\begin{table}[h!]
\vskip 0.1in
\begin{center}
\begin{tabular}{ll}
\toprule
\textbf{Parameter}  & \textbf{Single Module Task} \\ \midrule
$\omega$     & 20\\ 
$t$          & 20\\ 
Gradient clip range     & [-0.1,0.1]\\     
Max stored losses (for reinitialisation check) & 5000\\
Minimum number of epochs before regularisation starts & 10000\\\bottomrule
\end{tabular}
\end{center}
\caption{Parameters specific to the iNALU for the Single Module Task.}
\label{tab:sltr-inalu-params}
\vskip -0.1in
\end{table}

\section{Single Module Task: Alternative Options for Generating a Success Threshold}\label{app:eps-threshold}
Other methods can be used to generate the $\epsilon$-threshold. 
The factors which can be changed include: 
\begin{itemize}
    \item the $\epsilon$-perfect model, e.g., we could use a $\epsilon$-perfect NALU expression which uses log space. 
    \item the comparison metric against the perfect model. A MSE is used but other metrics such as PCC or MAPE are also valid, with each metric having it's own biases.
    \item the value of $\epsilon$ to control the tolerance of the threshold. Larger values would be more tolerant while smaller values are harsher. 
\end{itemize}

All these can be modified and should be considered if creating a new threshold evaluation scheme. 
However, the three points to be consistent on no matter the chosen evaluation method is to: 
(1) be task and range dependant, 
(2) use the same threshold when comparing models on the same task, 
and (3) not make the generation of the threshold dependant on the benchmarked model. 

A suggestion for an additional metric would be to generate a $\epsilon$-threshold used to measure if the model weights have converged enough prior to applying any (discretisation) regularisation. 
Using the NMU as an example, a pre-regularisation threshold would set the $epsilon$ to be $<0.5$. 
The resultant threshold will therefore determine if the weights have converged towards the direction of the expected discrete value before the regularisation begins to get applied. 
Having this threshold helps determine where the module would be having difficulties i.e., if the problem is the finding the global-minima or if the problem lies in regularisation of the weights to the final value since the regularisation (if given enough priority) would force values to round to the nearest integer. 

\newpage
\section{Results for the Single Module Task}\label{app:sltr-tables}
\subsection{Addition}
\begin{longtable}[t]{rrlll}
\caption{\label{tab:benchmark-sltr-op-add}Results for addition. Comparison of the success-rate, model convergence iteration, and the sparsity error, with 95\% confidence interval on the ``single layer'' task. Each value is a summary of 25 different seeds. 
  Bold values refers to the best result for a evaluation metric for a single module across the different ranges.}\\
\toprule
\multicolumn{1}{c}{Model} & \multicolumn{1}{c}{Range} & \multicolumn{1}{c}{Success} & \multicolumn{1}{c}{Solved at} & \multicolumn{1}{c}{Sparsity error} \\
\cmidrule(l{3pt}r{3pt}){1-1} \cmidrule(l{3pt}r{3pt}){2-2} \cmidrule(l{3pt}r{3pt}){3-3} \cmidrule(l{3pt}r{3pt}){4-4} \cmidrule(l{3pt}r{3pt}){5-5}
 &  & Rate & Mean & Mean\\
\midrule
\endfirsthead
\caption[]{Results for addition. Comparison of the success-rate, model convergence iteration, and the sparsity error, with 95\% confidence interval on the ``single layer'' task. Each value is a summary of 25 different seed \textit{(continued)}}\\
\toprule
\multicolumn{1}{c}{Model} & \multicolumn{1}{c}{Range} & \multicolumn{1}{c}{Success} & \multicolumn{1}{c}{Solved at} & \multicolumn{1}{c}{Sparsity error} \\
\cmidrule(l{3pt}r{3pt}){1-1} \cmidrule(l{3pt}r{3pt}){2-2} \cmidrule(l{3pt}r{3pt}){3-3} \cmidrule(l{3pt}r{3pt}){4-4} \cmidrule(l{3pt}r{3pt}){5-5}
 &  & Rate & Mean & Mean\\
\midrule
\endhead
\
\endfoot
\bottomrule
\endlastfoot
 & U[-0.2,-0.1) & $0\% {~}^{+13\%}_{-0\%}$ & --- & ---\\

\nopagebreak
 & U[-1.2,-1.1) & $28\% {~}^{+20\%}_{-14\%}$ & $4.9 \cdot 10^{4} {~}^{+8.2 \cdot 10^{2}}_{-8.3 \cdot 10^{2}}$ & $1.4 \cdot 10^{-5} {~}^{+2.3 \cdot 10^{-6}}_{-2.3 \cdot 10^{-6}}$\\

\nopagebreak
 & U[-2,-1) & $64\% {~}^{+16\%}_{-19\%}$ & $4.8 \cdot 10^{4} {~}^{+7.3 \cdot 10^{2}}_{-7.4 \cdot 10^{2}}$ & $1.2 \cdot 10^{-5} {~}^{+1.0 \cdot 10^{-6}}_{-1.0 \cdot 10^{-6}}$\\

\nopagebreak
 & U[-2,2) & $40\% {~}^{+19\%}_{-17\%}$ & $4.9 \cdot 10^{4} {~}^{+6.4 \cdot 10^{2}}_{-6.4 \cdot 10^{2}}$ & $1.5 \cdot 10^{-5} {~}^{+1.0 \cdot 10^{-6}}_{-1.0 \cdot 10^{-6}}$\\

\nopagebreak
 & U[-20,-10) & $\mathbf{84\%} {~}^{+10\%}_{-19\%}$ & $4.5 \cdot 10^{4} {~}^{+8.2 \cdot 10^{2}}_{-8.3 \cdot 10^{2}}$ & $\mathbf{5.4 \cdot 10^{-6}} {~}^{+1.1 \cdot 10^{-6}}_{-1.1 \cdot 10^{-6}}$\\

\nopagebreak
 & U[0.1,0.2) & $0\% {~}^{+13\%}_{-0\%}$ & --- & ---\\

\nopagebreak
 & U[1,2) & $68\% {~}^{+15\%}_{-20\%}$ & $4.8 \cdot 10^{4} {~}^{+7.5 \cdot 10^{2}}_{-7.5 \cdot 10^{2}}$ & $1.2 \cdot 10^{-5} {~}^{+1.1 \cdot 10^{-6}}_{-1.1 \cdot 10^{-6}}$\\

\nopagebreak
 & U[1.1,1.2) & $36\% {~}^{+19\%}_{-16\%}$ & $4.9 \cdot 10^{4} {~}^{+8.3 \cdot 10^{2}}_{-8.3 \cdot 10^{2}}$ & $1.4 \cdot 10^{-5} {~}^{+1.7 \cdot 10^{-6}}_{-1.7 \cdot 10^{-6}}$\\

\nopagebreak
\multirow{-9}{*}{\raggedleft\arraybackslash $\mathrm{NAC}_{+}$} & U[10,20) & $\mathbf{84\%} {~}^{+10\%}_{-19\%}$ & $\mathbf{4.5 \cdot 10^{4}} {~}^{+7.8 \cdot 10^{2}}_{-7.8 \cdot 10^{2}}$ & $\mathbf{5.4 \cdot 10^{-6}} {~}^{+1.1 \cdot 10^{-6}}_{-1.1 \cdot 10^{-6}}$\\
\cmidrule{1-5}
\nopagebreak
 & U[-0.2,-0.1) & $\mathbf{24\%} {~}^{+19\%}_{-13\%}$ & $4.4 \cdot 10^{4} {~}^{+2.5 \cdot 10^{3}}_{-2.5 \cdot 10^{3}}$ & $6.7 \cdot 10^{-6} {~}^{+3.0 \cdot 10^{-6}}_{-3.0 \cdot 10^{-6}}$\\

\nopagebreak
 & U[-1.2,-1.1) & $0\% {~}^{+13\%}_{-0\%}$ & --- & ---\\

\nopagebreak
 & U[-2,-1) & $0\% {~}^{+13\%}_{-0\%}$ & --- & ---\\

\nopagebreak
 & U[-2,2) & $0\% {~}^{+13\%}_{-0\%}$ & --- & ---\\

\nopagebreak
\nopagebreak
 & U[-20,-10) & $8\% {~}^{+17\%}_{-6\%}$ & $4.7 \cdot 10^{4} {~}^{+3.9 \cdot 10^{3}}_{-3.9 \cdot 10^{3}}$ & $9.0 \cdot 10^{-6} {~}^{+2.7 \cdot 10^{-5}}_{-9.0 \cdot 10^{-6}}$\\

\nopagebreak
 & U[0.1,0.2) & $12\% {~}^{+18\%}_{-8\%}$ & $\mathbf{4.0 \cdot 10^{4}} {~}^{+6.5 \cdot 10^{2}}_{-6.5 \cdot 10^{2}}$ & $4.6 \cdot 10^{-6} {~}^{+2.0 \cdot 10^{-6}}_{-2.0 \cdot 10^{-6}}$\\

\nopagebreak
 & U[1,2) & $0\% {~}^{+13\%}_{-0\%}$ & --- & ---\\

\nopagebreak
 & U[1.1,1.2) & $0\% {~}^{+13\%}_{-0\%}$ & --- & ---\\

\nopagebreak
\multirow{-9}{*}{\raggedleft\arraybackslash G-NALU} & U[10,20) & $4\% {~}^{+16\%}_{-3\%}$ & $4.2 \cdot 10^{4}$ & $\mathbf{3.5 \cdot 10^{-6}}$\\
\cmidrule{1-5}
\nopagebreak
 & U[-0.2,-0.1) & $\mathbf{100\%} {~}^{+0\%}_{-13\%}$ & $1.7 \cdot 10^{4} {~}^{+1.9 \cdot 10^{2}}_{-1.9 \cdot 10^{2}}$ & $7.8 \cdot 10^{-7} {~}^{+4.1 \cdot 10^{-8}}_{-4.1 \cdot 10^{-8}}$\\

\nopagebreak
 & U[-1.2,-1.1) & $\mathbf{100\%} {~}^{+0\%}_{-13\%}$ & $1.6 \cdot 10^{4} {~}^{+2.0 \cdot 10^{2}}_{-2.0 \cdot 10^{2}}$ & $2.0 \cdot 10^{-6} {~}^{+1.5 \cdot 10^{-7}}_{-1.5 \cdot 10^{-7}}$\\

\nopagebreak
 & U[-2,-1) & $\mathbf{100\%} {~}^{+0\%}_{-13\%}$ & $1.6 \cdot 10^{4} {~}^{+7.8 \cdot 10^{1}}_{-7.8 \cdot 10^{1}}$ & $2.0 \cdot 10^{-6} {~}^{+1.5 \cdot 10^{-7}}_{-1.5 \cdot 10^{-7}}$\\

\nopagebreak
 & U[-2,2) & $\mathbf{100\%} {~}^{+0\%}_{-13\%}$ & $1.6 \cdot 10^{4} {~}^{+2.0 \cdot 10^{2}}_{-2.0 \cdot 10^{2}}$ & $1.4 \cdot 10^{-6} {~}^{+2.9 \cdot 10^{-7}}_{-2.9 \cdot 10^{-7}}$\\

\nopagebreak
 & U[-20,-10) & $\mathbf{100\%} {~}^{+0\%}_{-13\%}$ & $\mathbf{1.6 \cdot 10^{4}} {~}^{+4.3 \cdot 10^{2}}_{-4.3 \cdot 10^{2}}$ & $5.8 \cdot 10^{-6} {~}^{+3.1 \cdot 10^{-7}}_{-3.1 \cdot 10^{-7}}$\\

\nopagebreak
 & U[0.1,0.2) & $\mathbf{100\%} {~}^{+0\%}_{-13\%}$ & $1.9 \cdot 10^{4} {~}^{+7.8 \cdot 10^{1}}_{-7.8 \cdot 10^{1}}$ & $\mathbf{8.8 \cdot 10^{-8}} {~}^{+1.2 \cdot 10^{-8}}_{-1.2 \cdot 10^{-8}}$\\

\nopagebreak
 & U[1,2) & $\mathbf{100\%} {~}^{+0\%}_{-13\%}$ & $1.6 \cdot 10^{4} {~}^{+1.1 \cdot 10^{2}}_{-1.1 \cdot 10^{2}}$ & $1.8 \cdot 10^{-7} {~}^{+4.5 \cdot 10^{-8}}_{-4.5 \cdot 10^{-8}}$\\

\nopagebreak
 & U[1.1,1.2) & $\mathbf{100\%} {~}^{+0\%}_{-13\%}$ & $1.6 \cdot 10^{4} {~}^{+1.1 \cdot 10^{2}}_{-1.1 \cdot 10^{2}}$ & $3.1 \cdot 10^{-7} {~}^{+1.9 \cdot 10^{-8}}_{-1.9 \cdot 10^{-8}}$\\

\nopagebreak
\multirow{-9}{*}{\raggedleft\arraybackslash iNALU} & U[10,20) & $\mathbf{100\%} {~}^{+0\%}_{-13\%}$ & $2.0 \cdot 10^{4} {~}^{+7.8 \cdot 10^{1}}_{-7.8 \cdot 10^{1}}$ & $4.6 \cdot 10^{-7} {~}^{+8.3 \cdot 10^{-8}}_{-8.3 \cdot 10^{-8}}$\\
\cmidrule{1-5}
\nopagebreak
 & U[-0.2,-0.1) & $52\% {~}^{+18\%}_{-19\%}$ & $3.8 \cdot 10^{4} {~}^{+4.2 \cdot 10^{3}}_{-4.5 \cdot 10^{3}}$ & $7.5 \cdot 10^{-6} {~}^{+2.1 \cdot 10^{-6}}_{-2.1 \cdot 10^{-6}}$\\

\nopagebreak
 & U[-1.2,-1.1) & $40\% {~}^{+19\%}_{-17\%}$ & $4.2 \cdot 10^{4} {~}^{+4.8 \cdot 10^{3}}_{-5.5 \cdot 10^{3}}$ & $9.9 \cdot 10^{-6} {~}^{+3.0 \cdot 10^{-6}}_{-3.0 \cdot 10^{-6}}$\\

\nopagebreak
 & U[-2,-1) & $64\% {~}^{+16\%}_{-19\%}$ & $4.4 \cdot 10^{4} {~}^{+3.0 \cdot 10^{3}}_{-3.3 \cdot 10^{3}}$ & $9.1 \cdot 10^{-6} {~}^{+1.4 \cdot 10^{-6}}_{-1.4 \cdot 10^{-6}}$\\

\nopagebreak
 & U[-2,2) & $0\% {~}^{+13\%}_{-0\%}$ & --- & ---\\

 & U[-20,-10) & $68\% {~}^{+15\%}_{-20\%}$ & $4.5 \cdot 10^{4} {~}^{+1.8 \cdot 10^{3}}_{-1.9 \cdot 10^{3}}$ & $5.3 \cdot 10^{-6} {~}^{+2.0 \cdot 10^{-6}}_{-2.0 \cdot 10^{-6}}$\\

\nopagebreak
 & U[0.1,0.2) & $16\% {~}^{+19\%}_{-10\%}$ & $\mathbf{2.5 \cdot 10^{4}} {~}^{+1.8 \cdot 10^{3}}_{-1.9 \cdot 10^{3}}$ & $\mathbf{2.9 \cdot 10^{-6}} {~}^{+1.7 \cdot 10^{-6}}_{-1.7 \cdot 10^{-6}}$\\

\nopagebreak
 & U[1,2) & $\mathbf{80\%} {~}^{+11\%}_{-19\%}$ & $4.4 \cdot 10^{4} {~}^{+2.2 \cdot 10^{3}}_{-2.4 \cdot 10^{3}}$ & $9.1 \cdot 10^{-6} {~}^{+8.8 \cdot 10^{-7}}_{-8.8 \cdot 10^{-7}}$\\

\nopagebreak
 & U[1.1,1.2) & $76\% {~}^{+13\%}_{-19\%}$ & $4.5 \cdot 10^{4} {~}^{+2.5 \cdot 10^{3}}_{-2.8 \cdot 10^{3}}$ & $1.0 \cdot 10^{-5} {~}^{+1.1 \cdot 10^{-6}}_{-1.1 \cdot 10^{-6}}$\\

\nopagebreak
\multirow{-9}{*}{\raggedleft\arraybackslash NALU} & U[10,20) & $20\% {~}^{+19\%}_{-11\%}$ & $3.8 \cdot 10^{4} {~}^{+8.4 \cdot 10^{3}}_{-8.9 \cdot 10^{3}}$ & $4.6 \cdot 10^{-6} {~}^{+7.1 \cdot 10^{-6}}_{-4.6 \cdot 10^{-6}}$\\
\cmidrule{1-5}
\nopagebreak
 & U[-0.2,-0.1) & $\mathbf{100\%} {~}^{+0\%}_{-13\%}$ & $5.0 \cdot 10^{3} {~}^{+2.6 \cdot 10^{2}}_{-2.6 \cdot 10^{2}}$ & $\mathbf{1.0 \cdot 10^{-16}} {~}^{+NaN \cdot 10^{-Inf}}_{-NaN \cdot 10^{-Inf}}$\\

\nopagebreak
 & U[-1.2,-1.1) & $\mathbf{100\%} {~}^{+0\%}_{-13\%}$ & $5.0 \cdot 10^{3} {~}^{+2.6 \cdot 10^{2}}_{-2.6 \cdot 10^{2}}$ & $\mathbf{1.0 \cdot 10^{-16}} {~}^{+NaN \cdot 10^{-Inf}}_{-NaN \cdot 10^{-Inf}}$\\

\nopagebreak
 & U[-2,-1) & $\mathbf{100\%} {~}^{+0\%}_{-13\%}$ & $5.0 \cdot 10^{3} {~}^{+2.6 \cdot 10^{2}}_{-2.6 \cdot 10^{2}}$ & $\mathbf{1.0 \cdot 10^{-16}} {~}^{+NaN \cdot 10^{-Inf}}_{-NaN \cdot 10^{-Inf}}$\\

\nopagebreak
 & U[-2,2) & $\mathbf{100\%} {~}^{+0\%}_{-13\%}$ & $\mathbf{4.2 \cdot 10^{3}} {~}^{+2.6 \cdot 10^{2}}_{-2.8 \cdot 10^{2}}$ & $\mathbf{1.0 \cdot 10^{-16}} {~}^{+NaN \cdot 10^{-Inf}}_{-NaN \cdot 10^{-Inf}}$\\

\nopagebreak
 & U[-20,-10) & $\mathbf{100\%} {~}^{+0\%}_{-13\%}$ & $5.0 \cdot 10^{3} {~}^{+2.6 \cdot 10^{2}}_{-2.6 \cdot 10^{2}}$ & $\mathbf{1.0 \cdot 10^{-16}} {~}^{+NaN \cdot 10^{-Inf}}_{-NaN \cdot 10^{-Inf}}$\\

\nopagebreak
 & U[0.1,0.2) & $\mathbf{100\%} {~}^{+0\%}_{-13\%}$ & $5.0 \cdot 10^{3} {~}^{+2.6 \cdot 10^{2}}_{-2.6 \cdot 10^{2}}$ & $\mathbf{1.0 \cdot 10^{-16}} {~}^{+NaN \cdot 10^{-Inf}}_{-NaN \cdot 10^{-Inf}}$\\

\nopagebreak
 & U[1,2) & $\mathbf{100\%} {~}^{+0\%}_{-13\%}$ & $5.0 \cdot 10^{3} {~}^{+2.6 \cdot 10^{2}}_{-2.6 \cdot 10^{2}}$ & $\mathbf{1.0 \cdot 10^{-16}} {~}^{+NaN \cdot 10^{-Inf}}_{-NaN \cdot 10^{-Inf}}$\\

\nopagebreak
 & U[1.1,1.2) & $\mathbf{100\%} {~}^{+0\%}_{-13\%}$ & $5.0 \cdot 10^{3} {~}^{+2.6 \cdot 10^{2}}_{-2.6 \cdot 10^{2}}$ & $\mathbf{1.0 \cdot 10^{-16}} {~}^{+NaN \cdot 10^{-Inf}}_{-NaN \cdot 10^{-Inf}}$\\

\nopagebreak
\multirow{-9}{*}{\raggedleft\arraybackslash NAU} & U[10,20) & $\mathbf{100\%} {~}^{+0\%}_{-13\%}$ & $5.0 \cdot 10^{3} {~}^{+2.6 \cdot 10^{2}}_{-2.6 \cdot 10^{2}}$ & $\mathbf{1.0 \cdot 10^{-16}} {~}^{+NaN \cdot 10^{-Inf}}_{-NaN \cdot 10^{-Inf}}$\\*
\end{longtable}

\pagebreak
\subsection{Subtraction}
\begin{longtable}[t]{rrlll}
\caption{\label{tab:benchmark-sltr-op-sub}Results for subtraction. Comparison of the success-rate, model convergence iteration, and the sparsity error, with 95\% confidence interval on the ``single layer'' task. Each value is a summary of 25 different seeds. 
  Bold values refers to the best result for a evaluation metric for a single module across the different ranges.}\\
\toprule
\multicolumn{1}{c}{Model} & \multicolumn{1}{c}{Range} & \multicolumn{1}{c}{Success} & \multicolumn{1}{c}{Solved at} & \multicolumn{1}{c}{Sparsity error} \\
\cmidrule(l{3pt}r{3pt}){1-1} \cmidrule(l{3pt}r{3pt}){2-2} \cmidrule(l{3pt}r{3pt}){3-3} \cmidrule(l{3pt}r{3pt}){4-4} \cmidrule(l{3pt}r{3pt}){5-5}
 &  & Rate & Mean & Mean\\
\midrule
\endfirsthead
\caption[]{Results for subtraction. Comparison of the success-rate, model convergence iteration, and the sparsity error, with 95\% confidence interval on the ``single layer'' task. Each value is a summary of 25 different seed \textit{(continued)}}\\
\toprule
\multicolumn{1}{c}{Model} & \multicolumn{1}{c}{Range} & \multicolumn{1}{c}{Success} & \multicolumn{1}{c}{Solved at} & \multicolumn{1}{c}{Sparsity error} \\
\cmidrule(l{3pt}r{3pt}){1-1} \cmidrule(l{3pt}r{3pt}){2-2} \cmidrule(l{3pt}r{3pt}){3-3} \cmidrule(l{3pt}r{3pt}){4-4} \cmidrule(l{3pt}r{3pt}){5-5}
 &  & Rate & Mean & Mean\\
\midrule
\endhead
\
\endfoot
\bottomrule
\endlastfoot
 & U[-0.2,-0.1) & $0\% {~}^{+13\%}_{-0\%}$ & --- & ---\\

\nopagebreak
 & U[-1.2,-1.1) & $0\% {~}^{+13\%}_{-0\%}$ & --- & ---\\

\nopagebreak
 & U[-2,-1) & $20\% {~}^{+19\%}_{-11\%}$ & $4.4 \cdot 10^{4} {~}^{+5.2 \cdot 10^{3}}_{-5.4 \cdot 10^{3}}$ & $2.2 \cdot 10^{-5} {~}^{+4.0 \cdot 10^{-6}}_{-4.4 \cdot 10^{-6}}$\\

\nopagebreak
 & U[-2,2) & $52\% {~}^{+18\%}_{-19\%}$ & $4.6 \cdot 10^{4} {~}^{+2.8 \cdot 10^{3}}_{-3.0 \cdot 10^{3}}$ & $\mathbf{1.4 \cdot 10^{-5}} {~}^{+1.7 \cdot 10^{-6}}_{-1.7 \cdot 10^{-6}}$\\

\nopagebreak
 & U[-20,-10) & $\mathbf{84\%} {~}^{+10\%}_{-19\%}$ & $\mathbf{4.1 \cdot 10^{4}} {~}^{+2.2 \cdot 10^{3}}_{-2.4 \cdot 10^{3}}$ & $1.6 \cdot 10^{-5} {~}^{+3.9 \cdot 10^{-6}}_{-3.9 \cdot 10^{-6}}$\\

\nopagebreak
 & U[0.1,0.2) & $0\% {~}^{+13\%}_{-0\%}$ & --- & ---\\

\nopagebreak
 & U[1,2) & $20\% {~}^{+19\%}_{-11\%}$ & $4.4 \cdot 10^{4} {~}^{+5.2 \cdot 10^{3}}_{-5.4 \cdot 10^{3}}$ & $2.2 \cdot 10^{-5} {~}^{+4.0 \cdot 10^{-6}}_{-4.3 \cdot 10^{-6}}$\\

\nopagebreak
 & U[1.1,1.2) & $0\% {~}^{+13\%}_{-0\%}$ & --- & ---\\

\nopagebreak
\multirow{-9}{*}{\raggedleft\arraybackslash $\mathrm{NAC}_{+}$} & U[10,20) & $\mathbf{84\%} {~}^{+10\%}_{-19\%}$ & $\mathbf{4.1 \cdot 10^{4}} {~}^{+2.2 \cdot 10^{3}}_{-2.4 \cdot 10^{3}}$ & $1.6 \cdot 10^{-5} {~}^{+3.9 \cdot 10^{-6}}_{-3.9 \cdot 10^{-6}}$\\
\cmidrule{1-5}
\nopagebreak
 & U[-0.2,-0.1) & $0\% {~}^{+13\%}_{-0\%}$ & --- & ---\\

\nopagebreak
\nopagebreak
 & U[-1.2,-1.1) & $0\% {~}^{+13\%}_{-0\%}$ & --- & ---\\

\nopagebreak
\nopagebreak
 & U[-2,-1) & $0\% {~}^{+13\%}_{-0\%}$ & --- & ---\\

\nopagebreak
 & U[-2,2) & $0\% {~}^{+13\%}_{-0\%}$ & --- & ---\\

\nopagebreak
\nopagebreak
 & U[-20,-10) & $4\% {~}^{+16\%}_{-3\%}$ & $\mathbf{4.2 \cdot 10^{4}}$ & $\mathbf{2.0 \cdot 10^{-5}}$\\

\nopagebreak
 & U[0.1,0.2) & $0\% {~}^{+13\%}_{-0\%}$ & --- & ---\\

\nopagebreak
\nopagebreak
 & U[1,2) & $0\% {~}^{+13\%}_{-0\%}$ & --- & ---\\

\nopagebreak
 & U[1.1,1.2) & $0\% {~}^{+13\%}_{-0\%}$ & --- & ---\\

\nopagebreak
\nopagebreak
\multirow{-9}{*}{\raggedleft\arraybackslash G-NALU} & U[10,20) & $\mathbf{20\%} {~}^{+19\%}_{-11\%}$ & $4.5 \cdot 10^{4} {~}^{+3.3 \cdot 10^{3}}_{-3.4 \cdot 10^{3}}$ & $2.1 \cdot 10^{-5} {~}^{+1.1 \cdot 10^{-5}}_{-1.1 \cdot 10^{-5}}$\\
\cmidrule{1-5}
\nopagebreak
 & U[-0.2,-0.1) & $\mathbf{100\%} {~}^{+0\%}_{-13\%}$ & $1.7 \cdot 10^{4} {~}^{+1.3 \cdot 10^{2}}_{-1.3 \cdot 10^{2}}$ & $2.5 \cdot 10^{-7} {~}^{+2.7 \cdot 10^{-8}}_{-2.7 \cdot 10^{-8}}$\\

\nopagebreak
 & U[-1.2,-1.1) & $40\% {~}^{+19\%}_{-17\%}$ & $2.1 \cdot 10^{4} {~}^{+3.9 \cdot 10^{2}}_{-3.9 \cdot 10^{2}}$ & $\mathbf{1.3 \cdot 10^{-8}} {~}^{+8.0 \cdot 10^{-9}}_{-8.0 \cdot 10^{-9}}$\\

\nopagebreak
 & U[-2,-1) & $\mathbf{100\%} {~}^{+0\%}_{-13\%}$ & $1.7 \cdot 10^{4} {~}^{+2.7 \cdot 10^{2}}_{-2.7 \cdot 10^{2}}$ & $5.8 \cdot 10^{-7} {~}^{+1.2 \cdot 10^{-7}}_{-1.2 \cdot 10^{-7}}$\\

\nopagebreak
 & U[-2,2) & $\mathbf{100\%} {~}^{+0\%}_{-13\%}$ & $1.7 \cdot 10^{4} {~}^{+1.6 \cdot 10^{2}}_{-1.6 \cdot 10^{2}}$ & $8.7 \cdot 10^{-7} {~}^{+8.2 \cdot 10^{-8}}_{-8.2 \cdot 10^{-8}}$\\

\nopagebreak
 & U[-20,-10) & $\mathbf{100\%} {~}^{+0\%}_{-13\%}$ & $\mathbf{1.6 \cdot 10^{4}} {~}^{+5.7 \cdot 10^{2}}_{-5.6 \cdot 10^{2}}$ & $9.4 \cdot 10^{-7} {~}^{+2.1 \cdot 10^{-7}}_{-2.1 \cdot 10^{-7}}$\\

\nopagebreak
 & U[0.1,0.2) & $\mathbf{100\%} {~}^{+0\%}_{-13\%}$ & $1.7 \cdot 10^{4} {~}^{+1.9 \cdot 10^{2}}_{-1.9 \cdot 10^{2}}$ & $1.5 \cdot 10^{-7} {~}^{+2.0 \cdot 10^{-8}}_{-2.0 \cdot 10^{-8}}$\\

\nopagebreak
 & U[1,2) & $\mathbf{100\%} {~}^{+0\%}_{-13\%}$ & $1.7 \cdot 10^{4} {~}^{+1.5 \cdot 10^{2}}_{-1.5 \cdot 10^{2}}$ & $1.7 \cdot 10^{-7} {~}^{+2.7 \cdot 10^{-8}}_{-2.7 \cdot 10^{-8}}$\\

\nopagebreak
 & U[1.1,1.2) & $56\% {~}^{+17\%}_{-19\%}$ & $2.0 \cdot 10^{4} {~}^{+1.9 \cdot 10^{2}}_{-1.9 \cdot 10^{2}}$ & $4.4 \cdot 10^{-8} {~}^{+4.4 \cdot 10^{-8}}_{-4.4 \cdot 10^{-8}}$\\

\nopagebreak
\multirow{-9}{*}{\raggedleft\arraybackslash iNALU} & U[10,20) & $\mathbf{100\%} {~}^{+0\%}_{-13\%}$ & $1.7 \cdot 10^{4} {~}^{+7.7 \cdot 10^{2}}_{-7.8 \cdot 10^{2}}$ & $7.6 \cdot 10^{-7} {~}^{+6.7 \cdot 10^{-8}}_{-6.7 \cdot 10^{-8}}$\\
\cmidrule{1-5}
\nopagebreak
 & U[-0.2,-0.1) & $0\% {~}^{+13\%}_{-0\%}$ & --- & ---\\

 & U[-1.2,-1.1) & $0\% {~}^{+13\%}_{-0\%}$ & --- & ---\\

 & U[-2,-1) & $12\% {~}^{+18\%}_{-8\%}$ & $4.6 \cdot 10^{4} {~}^{+3.4 \cdot 10^{3}}_{-3.5 \cdot 10^{3}}$ & $2.2 \cdot 10^{-5} {~}^{+3.1 \cdot 10^{-6}}_{-2.9 \cdot 10^{-6}}$\\

\nopagebreak
 & U[-2,2) & $0\% {~}^{+13\%}_{-0\%}$ & --- & ---\\

 & U[-20,-10) & $20\% {~}^{+19\%}_{-11\%}$ & $\mathbf{3.8 \cdot 10^{4}} {~}^{+7.4 \cdot 10^{3}}_{-8.1 \cdot 10^{3}}$ & $\mathbf{1.4 \cdot 10^{-5}} {~}^{+10.0 \cdot 10^{-6}}_{-10.0 \cdot 10^{-6}}$\\

\nopagebreak
 & U[0.1,0.2) & $0\% {~}^{+13\%}_{-0\%}$ & --- & ---\\

 & U[1,2) & $12\% {~}^{+18\%}_{-8\%}$ & $4.8 \cdot 10^{4} {~}^{+2.0 \cdot 10^{3}}_{-2.0 \cdot 10^{3}}$ & $2.3 \cdot 10^{-5} {~}^{+1.7 \cdot 10^{-6}}_{-1.4 \cdot 10^{-6}}$\\

\nopagebreak
 & U[1.1,1.2) & $0\% {~}^{+13\%}_{-0\%}$ & --- & ---\\

\multirow{-9}{*}{\raggedleft\arraybackslash NALU} & U[10,20) & $\mathbf{84\%} {~}^{+10\%}_{-19\%}$ & $4.3 \cdot 10^{4} {~}^{+2.3 \cdot 10^{3}}_{-2.6 \cdot 10^{3}}$ & $2.3 \cdot 10^{-5} {~}^{+8.3 \cdot 10^{-6}}_{-8.3 \cdot 10^{-6}}$\\
\cmidrule{1-5}
\nopagebreak
 & U[-0.2,-0.1) & $\mathbf{100\%} {~}^{+0\%}_{-13\%}$ & $6.0 \cdot 10^{3} {~}^{+6.4 \cdot 10^{2}}_{-6.6 \cdot 10^{2}}$ & $\mathbf{1.0 \cdot 10^{-16}} {~}^{+NaN \cdot 10^{-Inf}}_{-NaN \cdot 10^{-Inf}}$\\

\nopagebreak
 & U[-1.2,-1.1) & $\mathbf{100\%} {~}^{+0\%}_{-13\%}$ & $1.6 \cdot 10^{4} {~}^{+1.0 \cdot 10^{3}}_{-1.1 \cdot 10^{3}}$ & $2.7 \cdot 10^{-7} {~}^{+1.2 \cdot 10^{-7}}_{-1.2 \cdot 10^{-7}}$\\

\nopagebreak
 & U[-2,-1) & $\mathbf{100\%} {~}^{+0\%}_{-13\%}$ & $6.0 \cdot 10^{3} {~}^{+6.2 \cdot 10^{2}}_{-6.4 \cdot 10^{2}}$ & $3.1 \cdot 10^{-8} {~}^{+1.3 \cdot 10^{-8}}_{-1.3 \cdot 10^{-8}}$\\

\nopagebreak
 & U[-2,2) & $\mathbf{100\%} {~}^{+0\%}_{-13\%}$ & $\mathbf{4.0 \cdot 10^{3}} {~}^{+2.7 \cdot 10^{2}}_{-2.8 \cdot 10^{2}}$ & $\mathbf{1.0 \cdot 10^{-16}} {~}^{+NaN \cdot 10^{-Inf}}_{-NaN \cdot 10^{-Inf}}$\\

\nopagebreak
 & U[-20,-10) & $\mathbf{100\%} {~}^{+0\%}_{-13\%}$ & $5.9 \cdot 10^{3} {~}^{+6.3 \cdot 10^{2}}_{-6.5 \cdot 10^{2}}$ & $\mathbf{1.0 \cdot 10^{-16}} {~}^{+NaN \cdot 10^{-Inf}}_{-NaN \cdot 10^{-Inf}}$\\

\nopagebreak
 & U[0.1,0.2) & $\mathbf{100\%} {~}^{+0\%}_{-13\%}$ & $6.1 \cdot 10^{3} {~}^{+6.3 \cdot 10^{2}}_{-6.5 \cdot 10^{2}}$ & $\mathbf{1.0 \cdot 10^{-16}} {~}^{+NaN \cdot 10^{-Inf}}_{-NaN \cdot 10^{-Inf}}$\\

\nopagebreak
 & U[1,2) & $\mathbf{100\%} {~}^{+0\%}_{-13\%}$ & $6.0 \cdot 10^{3} {~}^{+6.2 \cdot 10^{2}}_{-6.4 \cdot 10^{2}}$ & $2.4 \cdot 10^{-8} {~}^{+1.2 \cdot 10^{-8}}_{-1.2 \cdot 10^{-8}}$\\

\nopagebreak
 & U[1.1,1.2) & $\mathbf{100\%} {~}^{+0\%}_{-13\%}$ & $1.6 \cdot 10^{4} {~}^{+1.0 \cdot 10^{3}}_{-1.1 \cdot 10^{3}}$ & $3.2 \cdot 10^{-7} {~}^{+1.5 \cdot 10^{-7}}_{-1.5 \cdot 10^{-7}}$\\

\nopagebreak
\multirow{-9}{*}{\raggedleft\arraybackslash NAU} & U[10,20) & $\mathbf{100\%} {~}^{+0\%}_{-13\%}$ & $5.9 \cdot 10^{3} {~}^{+6.2 \cdot 10^{2}}_{-6.4 \cdot 10^{2}}$ & $\mathbf{1.0 \cdot 10^{-16}} {~}^{+NaN \cdot 10^{-Inf}}_{-NaN \cdot 10^{-Inf}}$\\*
\end{longtable}

\pagebreak
\subsection{Multiplication}

\begin{longtable}[t]{rrlll}
\caption{\label{tab:benchmark-sltr-op-mul}Results for multiplication. Comparison of the success-rate, model convergence iteration, and the sparsity error, with 95\% confidence interval on the ``single layer'' task. Each value is a summary of 25 different seeds. 
  Bold values refers to the best result for a evaluation metric for a single module across the different ranges.}\\
\toprule
\multicolumn{1}{c}{Model} & \multicolumn{1}{c}{Range} & \multicolumn{1}{c}{Success} & \multicolumn{1}{c}{Solved at} & \multicolumn{1}{c}{Sparsity error} \\
\cmidrule(l{3pt}r{3pt}){1-1} \cmidrule(l{3pt}r{3pt}){2-2} \cmidrule(l{3pt}r{3pt}){3-3} \cmidrule(l{3pt}r{3pt}){4-4} \cmidrule(l{3pt}r{3pt}){5-5}
 &  & Rate & Mean & Mean\\
\midrule
\endfirsthead
\caption[]{Results for multiplication. Comparison of the success-rate, model convergence iteration, and the sparsity error, with 95\% confidence interval on the ``single layer'' task. Each value is a summary of 25 different seed \textit{(continued)}}\\
\toprule
\multicolumn{1}{c}{Model} & \multicolumn{1}{c}{Range} & \multicolumn{1}{c}{Success} & \multicolumn{1}{c}{Solved at} & \multicolumn{1}{c}{Sparsity error} \\
\cmidrule(l{3pt}r{3pt}){1-1} \cmidrule(l{3pt}r{3pt}){2-2} \cmidrule(l{3pt}r{3pt}){3-3} \cmidrule(l{3pt}r{3pt}){4-4} \cmidrule(l{3pt}r{3pt}){5-5}
 &  & Rate & Mean & Mean\\
\midrule
\endhead
\
\endfoot
\bottomrule
\endlastfoot
 & U[-0.2,-0.1) & $0\% {~}^{+13\%}_{-0\%}$ & --- & ---\\

\nopagebreak
 & U[-1.2,-1.1) & $0\% {~}^{+13\%}_{-0\%}$ & --- & ---\\

\nopagebreak
 & U[-2,-1) & $0\% {~}^{+13\%}_{-0\%}$ & --- & ---\\

\nopagebreak
 & U[-2,2) & $0\% {~}^{+13\%}_{-0\%}$ & --- & ---\\

\nopagebreak
 & U[-20,-10) & $\mathbf{88\%} {~}^{+8\%}_{-18\%}$ & $4.5 \cdot 10^{4} {~}^{+9.9 \cdot 10^{2}}_{-9.8 \cdot 10^{2}}$ & $\mathbf{1.7 \cdot 10^{-6}} {~}^{+5.0 \cdot 10^{-7}}_{-5.0 \cdot 10^{-7}}$\\

\nopagebreak
 & U[0.1,0.2) & $0\% {~}^{+13\%}_{-0\%}$ & --- & ---\\

\nopagebreak
 & U[1,2) & $0\% {~}^{+13\%}_{-0\%}$ & --- & ---\\

\nopagebreak
 & U[1.1,1.2) & $0\% {~}^{+13\%}_{-0\%}$ & --- & ---\\

\nopagebreak
\multirow{-9}{*}{\raggedleft\arraybackslash $\mathrm{NAC}_{\bullet}$} & U[10,20) & $\mathbf{88\%} {~}^{+8\%}_{-18\%}$ & $\mathbf{4.5 \cdot 10^{4}} {~}^{+9.9 \cdot 10^{2}}_{-9.9 \cdot 10^{2}}$ & $\mathbf{1.7 \cdot 10^{-6}} {~}^{+5.0 \cdot 10^{-7}}_{-5.0 \cdot 10^{-7}}$\\
\cmidrule{1-5}
\nopagebreak
 & U[-0.2,-0.1) & $0\% {~}^{+13\%}_{-0\%}$ & --- & ---\\

\nopagebreak
\nopagebreak
 & U[-1.2,-1.1) & $0\% {~}^{+13\%}_{-0\%}$ & --- & ---\\

\nopagebreak
\nopagebreak
 & U[-2,-1) & $0\% {~}^{+13\%}_{-0\%}$ & --- & ---\\

\nopagebreak
 & U[-2,2) & $0\% {~}^{+13\%}_{-0\%}$ & --- & ---\\

\nopagebreak
\nopagebreak
 & U[-20,-10) & $16\% {~}^{+19\%}_{-10\%}$ & $4.9 \cdot 10^{4} {~}^{+1.5 \cdot 10^{3}}_{-1.5 \cdot 10^{3}}$ & $3.8 \cdot 10^{-6} {~}^{+2.4 \cdot 10^{-6}}_{-2.4 \cdot 10^{-6}}$\\

\nopagebreak
 & U[0.1,0.2) & $\mathbf{32\%} {~}^{+20\%}_{-15\%}$ & $\mathbf{4.5 \cdot 10^{4}} {~}^{+3.0 \cdot 10^{3}}_{-3.0 \cdot 10^{3}}$ & $2.1 \cdot 10^{-5} {~}^{+4.0 \cdot 10^{-6}}_{-4.0 \cdot 10^{-6}}$\\

\nopagebreak
 & U[1,2) & $0\% {~}^{+13\%}_{-0\%}$ & --- & ---\\

\nopagebreak
 & U[1.1,1.2) & $0\% {~}^{+13\%}_{-0\%}$ & --- & ---\\

\nopagebreak
\nopagebreak
\multirow{-9}{*}{\raggedleft\arraybackslash G-NALU} & U[10,20) & $16\% {~}^{+19\%}_{-10\%}$ & $4.9 \cdot 10^{4} {~}^{+1.5 \cdot 10^{3}}_{-1.5 \cdot 10^{3}}$ & $\mathbf{3.5 \cdot 10^{-6}} {~}^{+2.3 \cdot 10^{-6}}_{-2.3 \cdot 10^{-6}}$\\
\cmidrule{1-5}
\nopagebreak
 & U[-0.2,-0.1) & $\mathbf{100\%} {~}^{+0\%}_{-13\%}$ & $2.1 \cdot 10^{4} {~}^{+1.8 \cdot 10^{2}}_{-1.9 \cdot 10^{2}}$ & $1.9 \cdot 10^{-9} {~}^{+1.7 \cdot 10^{-9}}_{-1.7 \cdot 10^{-9}}$\\

\nopagebreak
 & U[-1.2,-1.1) & $12\% {~}^{+18\%}_{-8\%}$ & $2.0 \cdot 10^{4} {~}^{+NaN \cdot 10^{-Inf}}_{-NaN \cdot 10^{-Inf}}$ & $\mathbf{1.0 \cdot 10^{-16}} {~}^{+NaN \cdot 10^{-Inf}}_{-NaN \cdot 10^{-Inf}}$\\

\nopagebreak
 & U[-2,-1) & $68\% {~}^{+15\%}_{-20\%}$ & $2.2 \cdot 10^{4} {~}^{+9.3 \cdot 10^{2}}_{-9.7 \cdot 10^{2}}$ & $3.5 \cdot 10^{-9} {~}^{+7.4 \cdot 10^{-9}}_{-3.5 \cdot 10^{-9}}$\\

\nopagebreak
 & U[-2,2) & $\mathbf{100\%} {~}^{+0\%}_{-13\%}$ & $1.7 \cdot 10^{4} {~}^{+NaN \cdot 10^{-Inf}}_{-NaN \cdot 10^{-Inf}}$ & $4.3 \cdot 10^{-8} {~}^{+1.4 \cdot 10^{-8}}_{-1.4 \cdot 10^{-8}}$\\

\nopagebreak
 & U[-20,-10) & $12\% {~}^{+18\%}_{-8\%}$ & $1.5 \cdot 10^{4} {~}^{+NaN \cdot 10^{-Inf}}_{-NaN \cdot 10^{-Inf}}$ & $7.2 \cdot 10^{-7} {~}^{+6.5 \cdot 10^{-7}}_{-6.5 \cdot 10^{-7}}$\\

\nopagebreak
 & U[0.1,0.2) & $4\% {~}^{+16\%}_{-3\%}$ & $2.1 \cdot 10^{4}$ & $1.0 \cdot 10^{-9}$\\

\nopagebreak
 & U[1,2) & $\mathbf{100\%} {~}^{+0\%}_{-13\%}$ & $1.7 \cdot 10^{4} {~}^{+NaN \cdot 10^{-Inf}}_{-NaN \cdot 10^{-Inf}}$ & $\mathbf{1.0 \cdot 10^{-16}} {~}^{+NaN \cdot 10^{-Inf}}_{-NaN \cdot 10^{-Inf}}$\\

\nopagebreak
 & U[1.1,1.2) & $\mathbf{100\%} {~}^{+0\%}_{-13\%}$ & $1.9 \cdot 10^{4} {~}^{+1.5 \cdot 10^{2}}_{-1.5 \cdot 10^{2}}$ & $5.7 \cdot 10^{-8} {~}^{+4.9 \cdot 10^{-9}}_{-4.9 \cdot 10^{-9}}$\\

\nopagebreak
\multirow{-9}{*}{\raggedleft\arraybackslash iNALU} & U[10,20) & $\mathbf{100\%} {~}^{+0\%}_{-13\%}$ & $\mathbf{1.5 \cdot 10^{4}} {~}^{+1.5 \cdot 10^{2}}_{-1.5 \cdot 10^{2}}$ & $1.5 \cdot 10^{-7} {~}^{+2.3 \cdot 10^{-8}}_{-2.3 \cdot 10^{-8}}$\\
\cmidrule{1-5}
\nopagebreak
 & U[-0.2,-0.1) & $0\% {~}^{+13\%}_{-0\%}$ & --- & ---\\

 & U[-1.2,-1.1) & $0\% {~}^{+13\%}_{-0\%}$ & --- & ---\\

 & U[-2,-1) & $12\% {~}^{+18\%}_{-8\%}$ & $4.6 \cdot 10^{4} {~}^{+3.4 \cdot 10^{3}}_{-3.4 \cdot 10^{3}}$ & $9.7 \cdot 10^{-6} {~}^{+5.8 \cdot 10^{-6}}_{-5.8 \cdot 10^{-6}}$\\

\nopagebreak
 & U[-2,2) & $0\% {~}^{+13\%}_{-0\%}$ & --- & ---\\

 & U[-20,-10) & $52\% {~}^{+18\%}_{-19\%}$ & $4.3 \cdot 10^{4} {~}^{+2.7 \cdot 10^{3}}_{-2.9 \cdot 10^{3}}$ & $1.3 \cdot 10^{-6} {~}^{+7.9 \cdot 10^{-7}}_{-7.9 \cdot 10^{-7}}$\\

\nopagebreak
 & U[0.1,0.2) & $60\% {~}^{+17\%}_{-19\%}$ & $\mathbf{3.7 \cdot 10^{4}} {~}^{+3.6 \cdot 10^{3}}_{-3.9 \cdot 10^{3}}$ & $1.7 \cdot 10^{-5} {~}^{+3.0 \cdot 10^{-6}}_{-3.0 \cdot 10^{-6}}$\\

\nopagebreak
 & U[1,2) & $8\% {~}^{+17\%}_{-6\%}$ & $4.6 \cdot 10^{4} {~}^{+5.8 \cdot 10^{3}}_{-5.9 \cdot 10^{3}}$ & $1.0 \cdot 10^{-5} {~}^{+6.1 \cdot 10^{-6}}_{-6.1 \cdot 10^{-6}}$\\

\nopagebreak
 & U[1.1,1.2) & $0\% {~}^{+13\%}_{-0\%}$ & --- & ---\\

\multirow{-9}{*}{\raggedleft\arraybackslash NALU} & U[10,20) & $\mathbf{84\%} {~}^{+10\%}_{-19\%}$ & $4.3 \cdot 10^{4} {~}^{+1.8 \cdot 10^{3}}_{-1.9 \cdot 10^{3}}$ & $\mathbf{1.2 \cdot 10^{-6}} {~}^{+5.0 \cdot 10^{-7}}_{-5.0 \cdot 10^{-7}}$\\
\cmidrule{1-5}
\nopagebreak
 & U[-0.2,-0.1) & $\mathbf{100\%} {~}^{+0\%}_{-13\%}$ & $2.1 \cdot 10^{4} {~}^{+NaN \cdot 10^{-Inf}}_{-NaN \cdot 10^{-Inf}}$ & $\mathbf{1.0 \cdot 10^{-16}} {~}^{+NaN \cdot 10^{-Inf}}_{-NaN \cdot 10^{-Inf}}$\\

\nopagebreak
 & U[-1.2,-1.1) & $68\% {~}^{+15\%}_{-20\%}$ & $\mathbf{2.0 \cdot 10^{3}} {~}^{+NaN \cdot 10^{-Inf}}_{-NaN \cdot 10^{-Inf}}$ & $\mathbf{1.0 \cdot 10^{-16}} {~}^{+NaN \cdot 10^{-Inf}}_{-NaN \cdot 10^{-Inf}}$\\

\nopagebreak
 & U[-2,-1) & $80\% {~}^{+11\%}_{-19\%}$ & $\mathbf{2.0 \cdot 10^{3}} {~}^{+NaN \cdot 10^{-Inf}}_{-NaN \cdot 10^{-Inf}}$ & $\mathbf{1.0 \cdot 10^{-16}} {~}^{+NaN \cdot 10^{-Inf}}_{-NaN \cdot 10^{-Inf}}$\\

\nopagebreak
 & U[-2,2) & $\mathbf{100\%} {~}^{+0\%}_{-13\%}$ & $2.3 \cdot 10^{3} {~}^{+1.7 \cdot 10^{2}}_{-1.7 \cdot 10^{2}}$ & $\mathbf{1.0 \cdot 10^{-16}} {~}^{+NaN \cdot 10^{-Inf}}_{-NaN \cdot 10^{-Inf}}$\\

\nopagebreak
 & U[-20,-10) & $\mathbf{100\%} {~}^{+0\%}_{-13\%}$ & $2.4 \cdot 10^{3} {~}^{+1.8 \cdot 10^{2}}_{-1.9 \cdot 10^{2}}$ & $\mathbf{1.0 \cdot 10^{-16}} {~}^{+NaN \cdot 10^{-Inf}}_{-NaN \cdot 10^{-Inf}}$\\

\nopagebreak
 & U[0.1,0.2) & $\mathbf{100\%} {~}^{+0\%}_{-13\%}$ & $5.5 \cdot 10^{3} {~}^{+4.4 \cdot 10^{2}}_{-4.3 \cdot 10^{2}}$ & $\mathbf{1.0 \cdot 10^{-16}} {~}^{+NaN \cdot 10^{-Inf}}_{-NaN \cdot 10^{-Inf}}$\\

\nopagebreak
 & U[1,2) & $\mathbf{100\%} {~}^{+0\%}_{-13\%}$ & $2.8 \cdot 10^{3} {~}^{+1.6 \cdot 10^{2}}_{-1.7 \cdot 10^{2}}$ & $\mathbf{1.0 \cdot 10^{-16}} {~}^{+NaN \cdot 10^{-Inf}}_{-NaN \cdot 10^{-Inf}}$\\

\nopagebreak
 & U[1.1,1.2) & $\mathbf{100\%} {~}^{+0\%}_{-13\%}$ & $3.0 \cdot 10^{3} {~}^{+2.4 \cdot 10^{2}}_{-2.5 \cdot 10^{2}}$ & $2.3 \cdot 10^{-7} {~}^{+8.4 \cdot 10^{-8}}_{-8.4 \cdot 10^{-8}}$\\

\nopagebreak
\multirow{-9}{*}{\raggedleft\arraybackslash NMU} & U[10,20) & $\mathbf{100\%} {~}^{+0\%}_{-13\%}$ & $2.6 \cdot 10^{3} {~}^{+1.9 \cdot 10^{2}}_{-2.0 \cdot 10^{2}}$ & $\mathbf{1.0 \cdot 10^{-16}} {~}^{+NaN \cdot 10^{-Inf}}_{-NaN \cdot 10^{-Inf}}$\\
\cmidrule{1-5}
\nopagebreak
 & U[-0.2,-0.1) & $0\% {~}^{+13\%}_{-0\%}$ & --- & ---\\

\nopagebreak
 & U[-1.2,-1.1) & $0\% {~}^{+13\%}_{-0\%}$ & --- & ---\\

\nopagebreak
 & U[-2,-1) & $0\% {~}^{+13\%}_{-0\%}$ & --- & ---\\

\nopagebreak
 & U[-2,2) & $40\% {~}^{+19\%}_{-17\%}$ & $\mathbf{3.6 \cdot 10^{3}} {~}^{+3.9 \cdot 10^{2}}_{-4.5 \cdot 10^{2}}$ & $1.2 \cdot 10^{-7} {~}^{+2.3 \cdot 10^{-8}}_{-2.3 \cdot 10^{-8}}$\\

\nopagebreak
 & U[-20,-10) & $0\% {~}^{+13\%}_{-0\%}$ & --- & ---\\

\nopagebreak
 & U[0.1,0.2) & $12\% {~}^{+18\%}_{-8\%}$ & $1.7 \cdot 10^{4} {~}^{+1.1 \cdot 10^{3}}_{-1.1 \cdot 10^{3}}$ & $1.1 \cdot 10^{-5} {~}^{+2.3 \cdot 10^{-5}}_{-1.1 \cdot 10^{-5}}$\\

\nopagebreak
 & U[1,2) & $\mathbf{100\%} {~}^{+0\%}_{-13\%}$ & $1.5 \cdot 10^{4} {~}^{+3.2 \cdot 10^{3}}_{-3.1 \cdot 10^{3}}$ & $1.1 \cdot 10^{-6} {~}^{+5.4 \cdot 10^{-7}}_{-5.4 \cdot 10^{-7}}$\\

\nopagebreak
 & U[1.1,1.2) & $28\% {~}^{+20\%}_{-14\%}$ & $4.4 \cdot 10^{3} {~}^{+5.5 \cdot 10^{2}}_{-5.9 \cdot 10^{2}}$ & $3.9 \cdot 10^{-6} {~}^{+4.2 \cdot 10^{-6}}_{-3.9 \cdot 10^{-6}}$\\

\nopagebreak
\multirow{-9}{*}{\raggedleft\arraybackslash NPU} & U[10,20) & $84\% {~}^{+10\%}_{-19\%}$ & $1.8 \cdot 10^{4} {~}^{+4.7 \cdot 10^{3}}_{-3.6 \cdot 10^{3}}$ & $\mathbf{4.5 \cdot 10^{-8}} {~}^{+2.4 \cdot 10^{-8}}_{-2.4 \cdot 10^{-8}}$\\
\cmidrule{1-5}
\nopagebreak
 & U[-0.2,-0.1) & $0\% {~}^{+13\%}_{-0\%}$ & --- & ---\\

\nopagebreak
 & U[-1.2,-1.1) & $0\% {~}^{+13\%}_{-0\%}$ & --- & ---\\

\nopagebreak
 & U[-2,-1) & $0\% {~}^{+13\%}_{-0\%}$ & --- & ---\\

\nopagebreak
 & U[-2,2) & $8\% {~}^{+17\%}_{-6\%}$ & $\mathbf{4.0 \cdot 10^{3}} {~}^{+NaN \cdot 10^{-Inf}}_{-NaN \cdot 10^{-Inf}}$ & $1.8 \cdot 10^{-7} {~}^{+NaN \cdot 10^{-Inf}}_{-NaN \cdot 10^{-Inf}}$\\

\nopagebreak
 & U[-20,-10) & $0\% {~}^{+13\%}_{-0\%}$ & --- & ---\\

\nopagebreak
 & U[0.1,0.2) & $12\% {~}^{+18\%}_{-8\%}$ & $1.7 \cdot 10^{4} {~}^{+1.1 \cdot 10^{3}}_{-1.1 \cdot 10^{3}}$ & $2.2 \cdot 10^{-5} {~}^{+4.5 \cdot 10^{-5}}_{-2.2 \cdot 10^{-5}}$\\

\nopagebreak
 & U[1,2) & $\mathbf{100\%} {~}^{+0\%}_{-13\%}$ & $1.5 \cdot 10^{4} {~}^{+3.2 \cdot 10^{3}}_{-3.1 \cdot 10^{3}}$ & $2.2 \cdot 10^{-6} {~}^{+1.1 \cdot 10^{-6}}_{-1.1 \cdot 10^{-6}}$\\

\nopagebreak
 & U[1.1,1.2) & $28\% {~}^{+20\%}_{-14\%}$ & $4.4 \cdot 10^{3} {~}^{+5.5 \cdot 10^{2}}_{-5.9 \cdot 10^{2}}$ & $7.8 \cdot 10^{-6} {~}^{+8.5 \cdot 10^{-6}}_{-7.8 \cdot 10^{-6}}$\\

\nopagebreak
\multirow{-9}{*}{\raggedleft\arraybackslash Real NPU} & U[10,20) & $84\% {~}^{+10\%}_{-19\%}$ & $1.8 \cdot 10^{4} {~}^{+4.7 \cdot 10^{3}}_{-3.6 \cdot 10^{3}}$ & $\mathbf{9.1 \cdot 10^{-8}} {~}^{+4.7 \cdot 10^{-8}}_{-4.7 \cdot 10^{-8}}$\\*
\end{longtable}

\pagebreak
\subsection{Division}
\begin{longtable}[t]{rrlll}
\caption{\label{tab:benchmark-sltr-op-div}Results for division. Comparison of the success-rate, model convergence iteration, and the sparsity error, with 95\% confidence interval on the ``single layer'' task. Each value is a summary of 25 different seeds. 
  Bold values refers to the best result for a evaluation metric for a single module across the different ranges.}\\
\toprule
\multicolumn{1}{c}{Model} & \multicolumn{1}{c}{Range} & \multicolumn{1}{c}{Success} & \multicolumn{1}{c}{Solved at} & \multicolumn{1}{c}{Sparsity error} \\
\cmidrule(l{3pt}r{3pt}){1-1} \cmidrule(l{3pt}r{3pt}){2-2} \cmidrule(l{3pt}r{3pt}){3-3} \cmidrule(l{3pt}r{3pt}){4-4} \cmidrule(l{3pt}r{3pt}){5-5}
 &  & Rate & Mean & Mean\\
\midrule
\endfirsthead
\caption[]{Results for division. Comparison of the success-rate, model convergence iteration, and the sparsity error, with 95\% confidence interval on the ``single layer'' task. Each value is a summary of 25 different seed \textit{(continued)}}\\
\toprule
\multicolumn{1}{c}{Model} & \multicolumn{1}{c}{Range} & \multicolumn{1}{c}{Success} & \multicolumn{1}{c}{Solved at} & \multicolumn{1}{c}{Sparsity error} \\
\cmidrule(l{3pt}r{3pt}){1-1} \cmidrule(l{3pt}r{3pt}){2-2} \cmidrule(l{3pt}r{3pt}){3-3} \cmidrule(l{3pt}r{3pt}){4-4} \cmidrule(l{3pt}r{3pt}){5-5}
 &  & Rate & Mean & Mean\\
\midrule
\endhead
\
\endfoot
\bottomrule
\endlastfoot
 & U[-0.2,-0.1) & $0\% {~}^{+13\%}_{-0\%}$ & --- & ---\\

\nopagebreak
 & U[-1.2,-1.1) & $0\% {~}^{+13\%}_{-0\%}$ & --- & ---\\

\nopagebreak
 & U[-2,-1) & $8\% {~}^{+17\%}_{-6\%}$ & $\mathbf{4.6 \cdot 10^{4}} {~}^{+4.8 \cdot 10^{3}}_{-4.9 \cdot 10^{3}}$ & $\mathbf{2.5 \cdot 10^{-5}} {~}^{+3.0 \cdot 10^{-6}}_{-2.4 \cdot 10^{-6}}$\\

\nopagebreak
 & U[-2,2) & $0\% {~}^{+13\%}_{-0\%}$ & --- & ---\\

\nopagebreak
 & U[-20,-10) & $\mathbf{16\%} {~}^{+19\%}_{-10\%}$ & $4.6 \cdot 10^{4} {~}^{+4.4 \cdot 10^{3}}_{-4.5 \cdot 10^{3}}$ & $3.3 \cdot 10^{-5} {~}^{+6.1 \cdot 10^{-6}}_{-5.9 \cdot 10^{-6}}$\\

\nopagebreak
 & U[0.1,0.2) & $0\% {~}^{+13\%}_{-0\%}$ & --- & ---\\

\nopagebreak
 & U[1,2) & $8\% {~}^{+17\%}_{-6\%}$ & $\mathbf{4.6 \cdot 10^{4}} {~}^{+4.8 \cdot 10^{3}}_{-4.9 \cdot 10^{3}}$ & $\mathbf{2.5 \cdot 10^{-5}} {~}^{+3.0 \cdot 10^{-6}}_{-2.4 \cdot 10^{-6}}$\\

\nopagebreak
 & U[1.1,1.2) & $0\% {~}^{+13\%}_{-0\%}$ & --- & ---\\

\nopagebreak
\multirow{-9}{*}{\raggedleft\arraybackslash $\mathrm{NAC}_{\bullet}$} & U[10,20) & $\mathbf{16\%} {~}^{+19\%}_{-10\%}$ & $4.6 \cdot 10^{4} {~}^{+4.6 \cdot 10^{3}}_{-4.8 \cdot 10^{3}}$ & $3.2 \cdot 10^{-5} {~}^{+6.6 \cdot 10^{-6}}_{-5.6 \cdot 10^{-6}}$\\
\cmidrule{1-5}
\nopagebreak
 & U[-0.2,-0.1) & $\mathbf{0\%} {~}^{+13\%}_{-0\%}$ & --- & ---\\

\nopagebreak
\nopagebreak
 & U[-1.2,-1.1) & $\mathbf{0\%} {~}^{+13\%}_{-0\%}$ & --- & ---\\

\nopagebreak
\nopagebreak
 & U[-2,-1) & $\mathbf{0\%} {~}^{+13\%}_{-0\%}$ & --- & ---\\

\nopagebreak
\nopagebreak
 & U[-2,2) & $\mathbf{0\%} {~}^{+13\%}_{-0\%}$ & --- & ---\\

\nopagebreak
\nopagebreak
 & U[-20,-10) & $\mathbf{0\%} {~}^{+13\%}_{-0\%}$ & --- & ---\\

\nopagebreak
\nopagebreak
 & U[0.1,0.2) & $\mathbf{0\%} {~}^{+13\%}_{-0\%}$ & --- & ---\\

\nopagebreak
\nopagebreak
 & U[1,2) & $\mathbf{0\%} {~}^{+13\%}_{-0\%}$ & --- & ---\\

\nopagebreak
\nopagebreak
 & U[1.1,1.2) & $\mathbf{0\%} {~}^{+13\%}_{-0\%}$ & --- & ---\\

\nopagebreak
\nopagebreak
\multirow{-9}{*}{\raggedleft\arraybackslash G-NALU} & U[10,20) & $\mathbf{0\%} {~}^{+13\%}_{-0\%}$ & --- & ---\\
\cmidrule{1-5}
\nopagebreak
\nopagebreak
 & U[-0.2,-0.1) & $0\% {~}^{+13\%}_{-0\%}$ & --- & ---\\

\nopagebreak
 & U[-1.2,-1.1) & $0\% {~}^{+13\%}_{-0\%}$ & --- & ---\\

\nopagebreak
 & U[-2,-1) & $0\% {~}^{+13\%}_{-0\%}$ & --- & ---\\

\nopagebreak
 & U[-2,2) & $0\% {~}^{+13\%}_{-0\%}$ & --- & ---\\

\nopagebreak
 & U[-20,-10) & $0\% {~}^{+13\%}_{-0\%}$ & --- & ---\\

\nopagebreak
 & U[0.1,0.2) & $\mathbf{100\%} {~}^{+0\%}_{-13\%}$ & $1.8 \cdot 10^{4} {~}^{+2.0 \cdot 10^{2}}_{-2.0 \cdot 10^{2}}$ & $3.1 \cdot 10^{-8} {~}^{+1.3 \cdot 10^{-8}}_{-1.3 \cdot 10^{-8}}$\\

\nopagebreak
 & U[1,2) & $\mathbf{100\%} {~}^{+0\%}_{-13\%}$ & $\mathbf{1.7 \cdot 10^{4}} {~}^{+2.0 \cdot 10^{2}}_{-2.0 \cdot 10^{2}}$ & $2.1 \cdot 10^{-7} {~}^{+2.8 \cdot 10^{-8}}_{-2.8 \cdot 10^{-8}}$\\

\nopagebreak
 & U[1.1,1.2) & $\mathbf{100\%} {~}^{+0\%}_{-13\%}$ & $1.8 \cdot 10^{4} {~}^{+2.3 \cdot 10^{2}}_{-2.3 \cdot 10^{2}}$ & $1.7 \cdot 10^{-7} {~}^{+4.0 \cdot 10^{-8}}_{-4.0 \cdot 10^{-8}}$\\

\nopagebreak
\multirow{-9}{*}{\raggedleft\arraybackslash iNALU} & U[10,20) & $20\% {~}^{+19\%}_{-11\%}$ & $2.4 \cdot 10^{4} {~}^{+2.6 \cdot 10^{3}}_{-2.7 \cdot 10^{3}}$ & $\mathbf{2.3 \cdot 10^{-8}} {~}^{+5.9 \cdot 10^{-8}}_{-2.3 \cdot 10^{-8}}$\\
\cmidrule{1-5}
\nopagebreak
 & U[-0.2,-0.1) & $\mathbf{0\%} {~}^{+13\%}_{-0\%}$ & --- & ---\\

 & U[-1.2,-1.1) & $\mathbf{0\%} {~}^{+13\%}_{-0\%}$ & --- & ---\\

 & U[-2,-1) & $\mathbf{0\%} {~}^{+13\%}_{-0\%}$ & --- & ---\\

 & U[-2,2) & $\mathbf{0\%} {~}^{+13\%}_{-0\%}$ & --- & ---\\

 & U[-20,-10) & $\mathbf{0\%} {~}^{+13\%}_{-0\%}$ & --- & ---\\

 & U[0.1,0.2) & $\mathbf{0\%} {~}^{+13\%}_{-0\%}$ & --- & ---\\

 & U[1,2) & $\mathbf{0\%} {~}^{+13\%}_{-0\%}$ & --- & ---\\

 & U[1.1,1.2) & $\mathbf{0\%} {~}^{+13\%}_{-0\%}$ & --- & ---\\

\multirow{-9}{*}{\raggedleft\arraybackslash NALU} & U[10,20) & $\mathbf{0\%} {~}^{+13\%}_{-0\%}$ & --- & ---\\
\cmidrule{1-5}
 & U[-0.2,-0.1) & $0\% {~}^{+13\%}_{-0\%}$ & --- & ---\\

\nopagebreak
 & U[-1.2,-1.1) & $0\% {~}^{+13\%}_{-0\%}$ & --- & ---\\

\nopagebreak
 & U[-2,-1) & $0\% {~}^{+13\%}_{-0\%}$ & --- & ---\\

\nopagebreak
 & U[-2,2) & $0\% {~}^{+13\%}_{-0\%}$ & --- & ---\\

\nopagebreak
 & U[-20,-10) & $0\% {~}^{+13\%}_{-0\%}$ & --- & ---\\

\nopagebreak
 & U[0.1,0.2) & $88\% {~}^{+8\%}_{-18\%}$ & $1.7 \cdot 10^{4} {~}^{+3.1 \cdot 10^{3}}_{-4.2 \cdot 10^{3}}$ & $\mathbf{3.1 \cdot 10^{-7}} {~}^{+5.2 \cdot 10^{-8}}_{-5.2 \cdot 10^{-8}}$\\

\nopagebreak
 & U[1,2) & $\mathbf{100\%} {~}^{+0\%}_{-13\%}$ & $2.1 \cdot 10^{4} {~}^{+4.6 \cdot 10^{3}}_{-4.1 \cdot 10^{3}}$ & $9.8 \cdot 10^{-7} {~}^{+1.4 \cdot 10^{-6}}_{-9.8 \cdot 10^{-7}}$\\

\nopagebreak
 & U[1.1,1.2) & $16\% {~}^{+19\%}_{-10\%}$ & $\mathbf{1.5 \cdot 10^{3}} {~}^{+4.2 \cdot 10^{2}}_{-4.9 \cdot 10^{2}}$ & $3.9 \cdot 10^{-7} {~}^{+3.9 \cdot 10^{-8}}_{-3.9 \cdot 10^{-8}}$\\

\nopagebreak
\multirow{-9}{*}{\raggedleft\arraybackslash NPU} & U[10,20) & $4\% {~}^{+16\%}_{-3\%}$ & $4.9 \cdot 10^{4}$ & $7.2 \cdot 10^{-7}$\\
\cmidrule{1-5}
\nopagebreak
 & U[-0.2,-0.1) & $32\% {~}^{+20\%}_{-15\%}$ & $7.8 \cdot 10^{3} {~}^{+2.7 \cdot 10^{3}}_{-2.6 \cdot 10^{3}}$ & $7.7 \cdot 10^{-7} {~}^{+1.4 \cdot 10^{-7}}_{-1.4 \cdot 10^{-7}}$\\

\nopagebreak
 & U[-1.2,-1.1) & $12\% {~}^{+18\%}_{-8\%}$ & $8.3 \cdot 10^{3} {~}^{+3.1 \cdot 10^{3}}_{-3.2 \cdot 10^{3}}$ & $3.8 \cdot 10^{-6} {~}^{+9.5 \cdot 10^{-6}}_{-3.8 \cdot 10^{-6}}$\\

\nopagebreak
 & U[-2,-1) & $84\% {~}^{+10\%}_{-19\%}$ & $2.1 \cdot 10^{4} {~}^{+4.2 \cdot 10^{3}}_{-5.2 \cdot 10^{3}}$ & $7.9 \cdot 10^{-7} {~}^{+4.5 \cdot 10^{-7}}_{-4.5 \cdot 10^{-7}}$\\

\nopagebreak
 & U[-2,2) & $0\% {~}^{+13\%}_{-0\%}$ & --- & ---\\

\nopagebreak
 & U[-20,-10) & $4\% {~}^{+16\%}_{-3\%}$ & $5.0 \cdot 10^{3}$ & $\mathbf{1.2 \cdot 10^{-7}}$\\

\nopagebreak
 & U[0.1,0.2) & $88\% {~}^{+8\%}_{-18\%}$ & $1.7 \cdot 10^{4} {~}^{+3.1 \cdot 10^{3}}_{-4.2 \cdot 10^{3}}$ & $6.3 \cdot 10^{-7} {~}^{+1.0 \cdot 10^{-7}}_{-1.0 \cdot 10^{-7}}$\\

\nopagebreak
 & U[1,2) & $\mathbf{100\%} {~}^{+0\%}_{-13\%}$ & $2.1 \cdot 10^{4} {~}^{+4.6 \cdot 10^{3}}_{-4.1 \cdot 10^{3}}$ & $2.0 \cdot 10^{-6} {~}^{+2.7 \cdot 10^{-6}}_{-2.0 \cdot 10^{-6}}$\\

\nopagebreak
 & U[1.1,1.2) & $16\% {~}^{+19\%}_{-10\%}$ & $\mathbf{1.5 \cdot 10^{3}} {~}^{+4.2 \cdot 10^{2}}_{-4.9 \cdot 10^{2}}$ & $7.7 \cdot 10^{-7} {~}^{+7.7 \cdot 10^{-8}}_{-7.7 \cdot 10^{-8}}$\\

\nopagebreak
\multirow{-9}{*}{\raggedleft\arraybackslash Real NPU} & U[10,20) & $4\% {~}^{+16\%}_{-3\%}$ & $4.9 \cdot 10^{4}$ & $1.4 \cdot 10^{-6}$\\*
\end{longtable}


\vskip 0.2in
\bibliography{references}

\end{document}